\documentclass[twoside,11pt]{article}

\usepackage[accepted]{melba}

\usepackage{amsmath,amsfonts}

\usepackage{gensymb}
\RequirePackage{algorithmic}
\usepackage[utf8]{inputenc} % allow utf-8 input
\usepackage[T1]{fontenc}    % use 8-bit T1 fonts
\usepackage{hyperref}       % hyperlinks
\usepackage{url}            % simple URL typesetting
\usepackage{booktabs}       % professional-quality tables
\usepackage{amsfonts}       % blackboard math symbols
\usepackage{nicefrac}       % compact symbols for 1/2, etc.
\usepackage{microtype}      % microtypography
\usepackage{lipsum}
\usepackage{graphicx}
\graphicspath{ {./images/} }
\usepackage{xcolor}
\usepackage{microtype}
\usepackage{graphicx}
\usepackage{subfigure}
\usepackage{booktabs} % for professional tables
\usepackage{breqn}
\newcommand\inv[1]{#1\raisebox{1.15ex}{$\scriptscriptstyle-\!1$}}
\newcommand{\RNum}[1]{\uppercase\expandafter{\romannumeral #1\relax}}

\usepackage[labelfont=bf]{caption}
\usepackage{tabularx}
\usepackage{booktabs}
\usepackage{multirow}
\usepackage{makecell}
\newcommand{\R}{\mathbb{R}}
\usepackage{}
\usepackage{float}
\usepackage{graphicx}
\usepackage{lipsum}
\usepackage[utf8]{inputenc} % allow utf-8 input
\usepackage[T1]{fontenc}    % use 8-bit T1 fonts
\usepackage{hyperref}       % hyperlinks
\usepackage{url}            % simple URL typesetting
\usepackage{booktabs}       % professional-quality tables
\usepackage{amsfonts}       % blackboard math symbols
\usepackage{nicefrac}       % compact symbols for 1/2, etc.
\usepackage{microtype}      % microtypography
\newcommand{\minus}{\scalebox{0.75}[1.0]{$-$}}
\usepackage{amsmath}
\usepackage[ruled,vlined]{algorithm2e}

\melbaheading{2022:009}{https://www.melba-journal.org/papers/2022:009.html}{2022}{1-29}{09/2021}{04/2022}{Popescu et al.}{Information Processing in Medical Imaging (IPMI) 2021}{Aasa Feragen, Stefan Sommer, Julia Schnabel, Mads Nielsen}

\ShortHeadings{Distributional Gaussian Process Layers}{Popescu et al.}
\firstpageno{1}

\title{Distributional Gaussian Processes Layers for Out-of-Distribution Detection}

\author{\name Sebastian G. Popescu \email s.popescu16@imperial.ac.uk \\  % start right after \author{, or there will be an extra space
	\addr Biomedical Image Analysis Group, Imperial College London, London, U.K.
	\AND
	\name David J. Sharp \email david.sharp@imperial.ac.uk \\
	\addr Computational, Cognitive \& Clinical Neuroimaging Laboratory, Imperial College London, London, U.K.
	\AND 
	\name James H. Cole \email  james.cole@ucl.ac.uk \\
	\addr Centre for Medical Image Computing, University College London, London, U.K.	
	\AND 
	\name Konstantinos Kamnitsas \email  konstantinos.kamnitsas@eng.ox.ac.uk \\
	\addr Department of Engineering Science, University of Oxford, Oxford, U.K. \\
	Biomedical Image Analysis Group, Imperial College London, London, U.K.
	\AND 
	\name Ben Glocker \email  b.glocker@imperial.ac.uk \\
	\addr Biomedical Image Analysis Group, Imperial College London, London, U.K.	
}

\begin{document}

% top matterhttps://www.overleaf.com/project/5ee5bc8f2d7ae30001ebf64e
\maketitle

% abstract
\begin{abstract}%   <- trailing '%' for backward compatibility of .sty file

Machine learning models deployed on medical imaging tasks must be equipped with
out-of-distribution detection capabilities in order to avoid erroneous predictions. It is unsure whether out-of-distribution detection models reliant on deep neural networks are suitable for detecting domain shifts in medical imaging. Gaussian Processes can reliably separate in-distribution data points from out-of-distribution data points via their mathematical construction. Hence, we propose a parameter efficient Bayesian layer for hierarchical convolutional Gaussian Processes that incorporates Gaussian Processes operating in Wasserstein-2 space to reliably propagate uncertainty. This directly replaces convolving Gaussian Processes with a distance-preserving affine operator on distributions. Our experiments on brain tissue-segmentation show that the resulting architecture approaches the performance of well-established deterministic segmentation algorithms (U-Net), which has not been achieved with previous hierarchical Gaussian Processes. Moreover, by applying the same segmentation model to out-of-distribution data (i.e., images with pathology such as brain tumors), we show that our uncertainty estimates result in out-of-distribution detection that outperforms the capabilities of previous Bayesian networks and reconstruction-based approaches that learn normative distributions. To facilitate future work our code is publicly available\footnote{\url{https://github.com/SebastianPopescu/DistGP_Layers}}.
\end{abstract}

% keywords
\begin{keywords}
  Gaussian Processes, Image Segmentation, Out-of-distribution Detection
\end{keywords}

\section{Introduction}

Deep learning methods have achieved state-of-the-art results on a plethora of medical image segmentation tasks to clinical risk assessment \citep{zhou2021review, tang2019role, imai2020validation}. However, their application in clinical settings remains challenging due to issues pertaining to lack of reliability and miscalibration of confidence estimates. Reliably estimating uncertainty in predictions is also of vital interest in adjacent machine learning fields such as reinforcement learning, to guide exploration, or in active learning, to guide the selection of data points for the next iteration of labelling. Most research into incorporating uncertainty into medical image segmentation has gravitated around modelling inter-rater variability and the inherent aleatoric uncertainty associated to the dataset, which can be caused by noise or inter-class ambiguities, alongside modelling the uncertainty in parameters \citep{czolbe2021segmentation}. However, less focus has been placed on how models behave when processing unexpected inputs which differ from the characteristics of the training data. Such inputs, often called anomalies, outliers or out-of-distribution samples, could possibly lead to deleterious effects in healthcare applications where predictive models may encounter data that is corrupted or from patients with diseases that the model is not trained for \citep{curth2019transferring, maartensson2020reliability}.

Out-of-distribution (OOD) detection in medical imaging has been mostly approached through the lens of reconstruction-based techniques involving some form of encoder-decoder network trained on normative datasets \citep{chen2019unsupervised, chen2021normative}. Conversely, we focus on enhancing task-specific models (e.g., a segmentation model) with reliable uncertainty quantification that enables outlier detection. Standard deep neural networks (DNNs), despite their high predictive performance, often exhibit unreasonably high confidence in predictions estimates when processing unseen samples that are not from the data manifold of the training set (e.g., in the presence of pathology under the hypothesis of training data being composed of normal subjects or in a more general setting the presence of motion artifacts never seen in training images). To alleviate this, Bayesian approaches that assign posteriors over weights ( MC Dropout \citep{gal2016dropout} included ) or in function space ( Repulsive Deep Ensembles \citep{d2021repulsive} included ) have been proposed \citep{wilson2020bayesian}. However, either assigning priors on weights or in function space does not necessarily lend itself to reliable OOD detection capabilities by virtue of inspecting the predictive variance of the model as was shown in \cite{henning2021bayesian}. The authors argue that both infinite-width networks, trained via the Neural Network Gaussian Process (NNGP) kernel \citep{lee2017deep}, or finite-width networks trained via Hamiltonian Monte Carlo \citep{neal2011mcmc} are not reliable for OOD detection since they show that the associated NNGP kernel is not correlated with distances between objects in input space. This loss of \emph{distance-awareness} after encoding data has catastrophic effects on OOD detection, as we will soon see.  Similarly, \cite{foong2019between} describe a limitation in the expressiveness of the predictive uncertainty estimate given by mean-field variational inference (MFVI) when applied as the inference technique for Bayesian Neural Networks (BNNs). Concretely, MFVI fails in offering quality uncertainty estimates in regions between well-separated clusters of data, which the authors coin as \emph{in-between} uncertainty, with potentially catastrophic consequences for active learning, Bayesian optimisation or robustness to out-of-distribution data. In this paper we follow an alternative approach, using Gaussian Processes (GP) as the building block for deep Bayesian networks. 

The use of GPs for image classification has garnered interest in the past years. Hybrid approaches, whereby a DNN's embedding mechanism is trained end-to-end with a GP as the classification layer, were the first attempts to unify the two approaches \citep{bradshaw2017adversarial}. The first convolutional kernel was proposed in \cite{van2017convolutional}, constructed by aggregating patch response functions. This approach was stacked on feed forward GP layers applied in a convolutional manner, with promising improvements in accuracy compared to their shallow counterpart \citep{blomqvist2018deep}. 

We expand on the aforementioned work, by introducing a simpler convolutional mechanism, which does not require convolving GPs at each layer and hence alleviates the computational cost of optimizing over inducing points' locations residing in high dimensional spaces alongside the issues brought upon by multi-output GPs. We propose a plug-in Bayesian layer more amenable to CNN architectures. More concretely, we seek to replace each individual component of a standard convolutional layer in convolutional neural networks (CNNs), respectively the convolved filters and the non-linear activation function. Firstly, we impose constraints on the filter such that we have an upper bound on distances after the convolution with regards to distances between the same objects beforehand. This will ensure that objects which were close in previous layers will remain close going forward, which as we shall see later on is a fundamental property for reliable OOD detection. Moreover, directly using convolved filters as opposed to convolved GPs \citep{blomqvist2018deep} solves the issue with optimizing high-dimensional inducing points' locations alongside introducing a simpler mechanism by which we can introduce correlations between channels \citep{nguyen2014collaborative}. Secondly, we replace the element-wise non-linear activation functions with Distributional Gaussian Processes (DistGP) \citep{bachoc2017gaussian} used in one-to-one mapping manner, essentially acting as a non-parametric activation function. A variant of DistGP used in a hierarchical setting akin to Deep Gaussian Processes (DGP) \citep{damianou2013deep} was shown to be better at detecting OOD due to both kernel and architecture design \citep{popescu2020hierarchical}. In this paper we will show that our proposed module is also suited for OOD detection on both toy/image data and biomedical scans.

In the remainder of this section we provide a deeper exploration of uncertainties used in literature for biomedical image segmentation, subsequently introducing the concept of \emph{distance-awareness} and imposing smoothness constraints on learned representations in a deep network as prerequisites for reliable OOD detection. These two properties will be key to motivate the imposed constraints and architecture choice of our proposed probabilistic module later on.

\subsection{Uncertainty quantification for biomedical imaging segmentation}

While prediction uncertainty can be computed for standard neural networks by using the softmax probability, these uncertainty estimates are often overconfident \citep{guo2017calibration,mcclure2019knowing}. Research into Bayesian models has focused on a separation of uncertainty into two different types, aleatoric (data intrinsic) and epistemic (model parameter uncertainty). To formalize this difference, we consider a multi-class classification problem, with classes denoted as $\{y_{1}, \cdots, y_{C} \}$ and model parameters denoted by $\theta$. We have the following predictive equation at testing time:
\begin{equation}
    p(y_{c}|x^{*},\mathbf{D}) = \int \underbrace{p(y_{c} \mid x^{*},\Theta)}_{\text{Aleatoric~Uncertainty}} \underbrace{p(\theta|\mathbf{D})}_{\text{Epistemic~Uncertainty}}d\theta \label{eqn:aleatoric_epistemic_unc_classification}
\end{equation}
Aleatoric uncertainty is irreducible, given by noise in the data acquisition process and has been considered in medical image segmentation \citep{monteiro2020stochastic}, whereas epistemic uncertainty can be reduced by providing more data during model training. This has also been studied in segmentation tasks \citep{nair2020exploring}. Previous work proposed to account for the uncertainty in the learned model parameters using an approximate Bayesian inference over the network weights \citep{kendall2015bayesian}. However, it was shown that this method may produce samples that vary pixel by pixel and thus may not capture complex spatially correlated structures in the distribution of segmentations maps. The probabilistic U-Net \citep{kohl2018probabilistic} produces samples with limited  diversity due to the fact that stochasticity is introduced in the highest resolution level of the U-Net. To solve this issue, \cite{baumgartner2019phiseg} introduce a hierachical structure between the different levels of the U-Net, hence introducing stochasticity at each level. Another improvement on the Probabilistic U-Net comes by adding variational dropout \citep{kingma2015variational} to the last layer to gain epistemic uncertainty quantification properties \citep{hu2019supervised}. All the models previously introduced relied on multiple annotations of the images with the intended goal of capturing this uncertainty in annotations with the aid of sampling from some form of latent variables which encode information about the whole image at varying scales of the U-Net. However, none of these previous works test how their models behave in the presence of outliers.

\subsection{Distributional Uncertainty as a proxy for OOD detection} \label{sec:motivation_dist_unc}

Besides the dichotomy consisting of aleatoric and epistemic uncertainty, reliably highlighting certain inputs which have undergone a domain shift \citep{lakshminarayanan2017simple} or out-of-distribution samples \citep{hendrycks2016baseline} has garnered a lot of interest in the past years. Succinctly, the aim is to measure the degree to which a model knows when it does not know, or more precisely if a network trained on a specific dataset is evaluated at testing time on a completely different dataset (potentially from a different modality or another application domain), then the expectation is that the network should output high predictive uncertainty on this set of data points that are very far from the training data manifold.

A problem with introducing this new type of uncertainty is how to disentangle it from epistemic uncertainty. For example, in the Deep Ensembles paper \citep{lakshminarayanan2017simple}, the authors propose to measure the disagreement between different sub-models of the deep ensemble $\sum\limits_{m=1}^{M} KL\left[p(y \mid x;\theta_{m}) \| \mathbb{E}\left[p(y \mid X)\right]\right]$ for $M$ sub-models with associated sub-model parameters $\theta_{m}$ and $\mathbb{E}\left[p(y \mid x)\right] = \frac{1}{M}\sum\limits_{m=1}^{M}p(y \mid x;\theta_{m})$ is the prediction of the ensemble. We remind ourselves that epistemic uncertainty can be reduced by adding more data. By this logic, epistemic uncertainty cannot be reduced outside the data manifold of our dataset since we don't add data points which do not stem from the same data generative pipeline (this is not true in the case of OOD detection models which explicitly use OOD samples during training/testing \citep{liang2017enhancing, hafner2020noise}). Hence, epistemic uncertainty can only be reduced inside the data manifold and should be zero outside the data manifold (assuming model is \emph{distance-aware}, which we will subsequently define). Conversely, our chosen measure for OOD detection should grow outside the data manifold and be close to or 0 inside the data manifold. With this in mind, the disagreement metric introduced in \cite{lakshminarayanan2017simple} cannot achieve this separation, confounding the two types of uncertainty.

\cite{malinin2018predictive} introduced for the  first time the separation of total uncertainty into three components: epistemic, aleatoric and distributional uncertainty. To make the distinction clearer, the authors argue that aleatoric uncertainty is a "known-unknown", whereby the model confidently states that an input data point is hard to classify (class overlap). Contrary, distributional uncertainty is an "unknown-unknown" due to the fact that the model is unfamiliar with the input space region that the test data comes from, thereby not being able to make confident predictions.

\begin{figure}[htb]
%  \vskip 0.2in
  \centering
    \includegraphics[width=\linewidth]{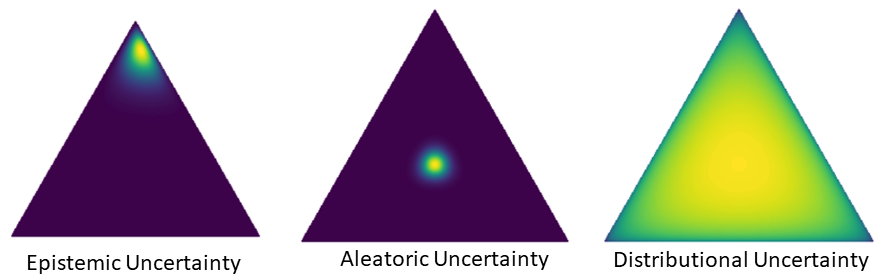}
    \caption[Decomposition of uncertainty in Dirichlet Networks]{Probability simplex for a 3 class classification problem, where corners corresponds to a class; Each point represents a categorical distribution, with brighter colors indicating higher density of the underlying ensemble. \textbf{Epistemic Uncertainty} captures uncertainty in model parameters caused by lack of data or model non-identifiability, with the ensemble of the predictions being concentrated in a corner of the probability simplex albeit with an increased diversity; \textbf{Aleatoric Uncertainty} captures class overlap, with the ensemble of predictions being confidently mapped to the highest predictive entropy; \textbf{Distributional Uncertainty} captures domain shift, with the ensemble of predictions being centred in the middle with highest possible degree of diversity;}
    \label{fig:dirichlet_uncertainty}  
%    \vskip -0.2in
\end{figure}

We briefly introduce the uncertainty decomposition mechanism introduced in \cite{malinin2018predictive}. Considering equation \eqref{eqn:aleatoric_epistemic_unc_classification}, by using Monte Carlo integration of above equation and computing the predictive entropy, we would not be able to discern between high predictive entropy due to aleatoric uncertainty (class overlap) or distributional uncertainty (dataset/domain shift). Hence, \cite{malinin2018predictive} propose to introduce a latent variable $\mu$ over the categorical variables corresponding to each class, parametrized as a distribution over distributions on a simplex, $p(\mu|x^{*},\theta)$. The intuition behind this Dirichlet distribution over the probability simplex is that OOD points should be scattered, whereas in-distribution points should concentrate. We can now re-write our predictive equation as:
\begin{equation}
    p(y_{c}|x^{*},\mathbf{D}) = \int\int \underbrace{p(y_{c}|x^{*},\mu)}_{\text{Aleatoric~Uncertainty}} 
    \underbrace{p(\mu|x^{*},\theta)}_{\text{Distributional~Uncertainty}} 
    \underbrace{p(\theta|\mathbf{D})}_{\text{Epistemic~Uncertainty}}d\mu~d\theta
\end{equation}
The authors argue that using a measure of spread of the ensemble ( after sampling from $p\left( \theta \mid \mathbf{D} \right)$) will be more informative. We remind ourselves that Mutual Information between variable $X$ and $Y$ can be expressed in terms of the difference between entropy and conditional entropy: $I(X;Y) = H(P(X)) - H(P(X|Y))$. Hence we can use the Mutual Information measure between model predictions and Dirichlet parameters to obtain a better measure of uncertainty.
We integrate out over $\theta$ in the main equation and we get:
\begin{equation}
    \underbrace{I[y,\mu|x^{*},\mathbf{D}]}_{\text{Distributional~Uncertainty}} = \underbrace{H[\mathbb{E}_{p(\mu|\mathbf{D})}p(y|x^{*},\mu)]}_{\text{Total~Uncertainty}} - 
    \underbrace{\mathbb{E}_{p(\mu|\mathbf{D})}[H[P(y|x^{*},\mu)]]}_{\text{Aleatoric~Uncertainty}}
\end{equation}
Connections between this uncertainty disentanglement framework specialized for DNN-parametrized Dirichlet distributions and uncertainty disentanglement in GP will be subsequently made clearer in subsection \ref{sec:evidential_learning}. Distributional uncertainty will be the key uncertainty score used throughout this paper to assess whether input data points are inside or outside the data manifold.

\subsection{Distance awareness and smoothness for out-of-distribution detection} 

Perhaps the greatest inspiration and motivation for this paper resides in the theoretical framework introduced in \cite{liu2020simple}, by which the authors outline what are some key mathematical conditions for provably reliable OOD detection in DNNs. We commence by briefly outlining the ideas introduced in aforementioned paper.

For an abstract data generating distribution $p\left( y \mid x \right)$, where $y$ is scalar, respectively $x \in \mathbb{X} \subset \mathbb{R}^{D}$ with the input data manifold being equipped with a suitable metric $\| \cdot \|_{X}$. We consider our training data $D = \{ (x_{i},y_{i})_{1:n}\}$ to be sampled from a subset of the full input space $x_{in-d} \in \mathbb{X}$, where the $\textit{in-d}$ abbreviation stems from in-distribution. With this in mind, we can consider the in-distribution data generating distribution $p_{in-d}\left(y\mid x \right) = p\left( y \mid x,~x\in \mathbb{X}_{in-d} \right)$, respectively the out-of-distribution data generating distribution $p_{ood}\left(y\mid x \right) = p\left( y \mid x,~x\notin \mathbb{X}_{in-d} \right)$. Hence, it is safe to assume the full data generating distribution $p\left(y \mid x \right)$ is composed as a mixture of the in-distribution and OOD generating distributions:
\begin{align}
    p\left( y \mid x \right) &=  p\left( y , x \in \mathbb{X}_{in-d} \mid x \right) + p\left( y , x \notin \mathbb{X}_{in-d} \mid x \right) \\
    &= p\left( y \mid x, x \in \mathbb{X}_{in-d}\right) p\left(x \in \mathbb{X}_{in-d} \right) + p\left( y \mid x, x \notin \mathbb{X}_{in-d}\right) p\left(x \notin \mathbb{X}_{in-d} \right) \\
    &= p_{in-d}\left(y\mid x \right) p\left(x \in \mathbb{X}_{in-d} \right) + p_{ood}\left(y\mid x \right) p\left(x \notin \mathbb{X}_{in-d} \right)
\end{align}
Evidently, during training we are only learning $p_{in-d}\left(y\mid x \right)$ since we only have access to $D \subset \mathbb{X}_{in-d}$. Therefore, our model is completely in the dark with regards to $p_{ood}\left(y\mid x \right)$. These two data generating distributions more often than not are fundamentally different (e.g., having trained a model on T1w MRI scans, subsequently feeding it with T2w MRI scans, an imaging modality which has an almost inverse scaling to represent varying brain tissue). With this in mind, \cite{liu2020simple} argue that the optimal strategy is for $p_{ood}\left(y\mid x \right)$ to be predicted as a uniform distribution, thus signalling the lack of knowledge of the model on this different input domain. We can now recall the distinction made by \cite{malinin2018predictive}, between "known-unknowns" (aleatoric uncertainty, e.g., class overlap)  and "unknown-unknowns" (distributional uncertainty, e.g., domain shift), both of which have an uninformative predictive distribution (maximum predictive entropy). However, to disentangle these two types of uncertainty, we need a second-order type of uncertainty that basically scatters logit samples when distributional uncertainty is high, respectively accurately samples logits to maximum predictive entropy in the case of high aleatoric uncertainty. We now formalize this desiderata by a condition called "distance awareness" in \cite{liu2020simple}.

\begin{definition}[Definition 1 in \cite{liu2020simple}] \label{def:distance_awareness}
    We consider the predictive distribution for unseen point $p\left( y^{*} \mid x^{*} \right)$ at testing time, for model trained on $\mathbb{X}_{in-d} \in \mathbb{X}$, with the data manifold being equipped with metric $\| \cdot \|_{X}$. Then, we can affirm that $p\left( y^{*} \mid x^{*} \right)$ is \emph{distance-aware} if there exists summary statistic $u\left( x^{*} \right)$ of $p\left( y^{*} \mid x^{*} \right)$ that embeds the distance between $\mathbb{X}_{in-d}$ and $x^{*}$:
    \begin{equation}
        u\left( x^{*} \right) = v\left[ \mathbb{E}_{x \sim \mathbb{X}_{in-d}} \left[ \| x^{*} - x \|_{X}^{2}\right]\right]
    \end{equation}
    , where $v$ is a monotonic function that increases with distance.    
    \end{definition}
    
Definition \ref{def:distance_awareness} does not make any assumptions related to the architecture of the model from which the predictive distribution stems. In practice we would have the following composition to arrive at the logits $logit(x^{*}) = f \circ enc\left( x^{*}\right)$, where $enc\left( \cdot \right)$ represents a network that outputs the representation learning layer and $f\left( \cdot \right)$ is the output layer. In \cite{liu2020simple} the authors propose the following two conditions to ensure that the composition is \emph{distance-aware}:
\begin{itemize}
    \item $f(\cdot)$ is \emph{distance-aware}
    \item $\mathbb{E}_{x \sim \mathbb{X}_{in-d}} \left[ \| x^{*} - X \|_{X}^{2}\right] \approx \mathbb{E}_{x \sim \mathbb{X}_{in-d}} \left[ \| enc(x^{*}) - enc(X) \|_{enc(X)}^{2}\right]$
\end{itemize}
The last condition means that distances between data points in input space should be correlated with distances in learned representation, which is equipped with a $\| \cdot \|_{enc(X)}$ metric. In our work, we will use GPs as $f$, because as we will see in section \ref{sec:primer_gp}, GPs are \emph{distance-aware} functions. This enables us to build a \emph{distance-aware} model that is more appropriate for OOD detection. Whereas GPs satisfy the \emph{distance-aware} condition for the last layer predictor, we are still left with the question on how to maintain distances in the learned representation correlated to distances in the input layer. This will be subsequently dealt with.

\paragraph{Network smoothness constraints} Throughout this paper we will consider the general term of "smoothness" of a model to mean the degree to which changes in the input have an effect on the output at a particular layer. The question now shifts into how can we quantify the smoothness of a network/function? In mathematical analysis a function $f : \mathbb{X} \to \mathbb{Y}$ is said to be k-smooth if the first k derivatives exist $\{f_{'}, f^{''}, \cdots, f^{(k)} \}$ and are continuous. We denote functions which have these properties as being of class $C^{k}$. For example, Gaussian Processes using squared exponential kernels are $C^{\infty}$ since the squared exponential kernel is infinitely differentiable. However, such a definition and quantification of \emph{smoothness} wouldn't aid us in ensuring the second condition of \emph{distance-awareness}. For this, we shall use Lipschitz continuity, which is defined as follows: considering two metric spaces $\mathbb{X}$ and $\mathbb{Y}$ equipped with metrics $\| \cdot \|_{X}$ and  $\| \cdot \|_{Y}$ and $f : \mathbb{X} \to \mathbb{Y}$ is Lipschitz continuous if $\exists$ real $K\geq 0$ such that $\forall x,y \in \mathbb{X}$ we have $\|f(x), f(y) \|_{Y} \leq K \|x,y \|_{X}$. Intuitively, for Lipschitz functions there is an upper limit in how much outputs can change with respect to distances in input space. It is perhaps better to highlight now that Lipschitz functions represent a global property. There are also locally Lipschitz continuous functions which respect the aforementioned condition just in a neighbourhood of $x$, respectively $B_{r}(x) = \{ y \in \mathbb{X} : \|x,y \|_{X} \leq r \}$. Lastly, bi-Lipschitz continuity is defined as $ \frac{1}{K} \|x,y \|_{X} \geq \|f(x), f(y) \|_{Y} \leq K \|x,y \|_{X}$, which is a property that avoids learning trivially smooth functions and maintains useful information \citep{rosca2020case}. With this in mind, recent work \citep{liu2020simple, van2021feature} have enforced the bi-Lipschitz property on feature extractors, thereby ensuring strong correlation between distances between data points in input space, respectively in the representation learning layer.

\subsection{Contributions}

This work makes the following main contributions:
 \begin{itemize}
  \item We introduce a Bayesian layer that can act as a drop-in replacement for standard convolutional layers. Operating on stochastic layers with Gaussian distributions, we upper bound the convolved affine operators in Wasserstein-2 space, thus ensuring Lipschitz continuity. To introduce non-linearities, we apply DistGP element-wise on the output of the constrained affine operator, thereby obtaining non-parametric ``activation functions'' which ensure adequate quantification of distributional uncertainty at each layer. 
  \item We derive theoretical requirements for the model to not suffer from \emph{feature collapse}, with additional empirical results to support the theory. 
  \item We demonstrate that a GP-based convolutional architecture can achieve competitive results in segmentation tasks in comparison to a U-Net.
  \item We show improved OOD detection results on both general OOD tasks and on medical images compared to previous OOD approaches such as reconstruction-based models.
\end{itemize}

\section{Background}

In this section we provide a brief review of the theoretical toolkit required for the remainder of the paper. We commence by laying out foundational material on GPs, followed by an introduction to attempts to sparsify GPs. Subsequently, we introduce an uncertainty disentanglement framework for sparse Gaussian Processes. We briefly define Wasserstein-2 distances and show how they can be used to define kernels operating on Gaussian distributions. Lastly, we introduce recent re-formulations of deep GPs through the lens of OOD detection.

\subsection{Primer on Gaussian Processes} \label{sec:primer_gp}

A Gaussian Process can be seen as a generalization of multivariate Gaussian random variables to infinite sets. We define this statement in more detail now. We consider $f(x)$ to be a stochastic field, with $x \in \mathbb{R}^{d}$ and we define $m(x) = \mathbb{E}\left[ f(x)\right]$ and $C(x_{i}, x_{j}) = Cov\left[f(x_{i}), f(x_{j}) \right]$. We denote a Gaussian Process (GP) $f(x)$ as:
\begin{equation}
    f(x) \sim GP\left(m(x), C\left(x_{i}, x_{j}\right) \right)
\end{equation}
The covariance function have the condition to generate non-negative-definite covariance matrices, more specifically they have to satisfy:
$
    \sum_{i,j}a_{i}a_{j}C\left( x_{i}, x_{j} \right) \geq 0
$
for any finite set $\{x_{1}, \cdots, x_{n} \}$ and any real valued coefficients $\{a_{1}, \cdots, a_{n} \}$. Throughout this paper we will only consider second-order stationary processes which have constant means and $Cov\left[f(x_{i}), f(x_{j}) \right] = C\left( \| x_{i} - x_{j} \| \right)$. We can see that such covariance functions are invariant to translations. 

Squared exponential/radial basis function kernel defines a general class of stationary covariance functions:
\begin{equation}
    k^{SE}(x_{i},x_{j}) = \sigma^{2} \exp{\left[\sum_{d=1}^{D}-\frac{\left(x_{i,d} - x_{j,d} \right)^{2}}{l^{2}_{d}} \right]} \label{eqn:euclidean_kernel}
\end{equation}
, where we have written its definition in the anisotropic case. The emphasis on the domain will make more sense in subsequent subsections where we will introduce kernels operating on Gaussian measures. Intuitively, the lengthscale values $\{l_{1}^{2}, \cdots, l_{D}^{2} \}$ represent the strength along a particular dimension of input space by which successive values are strongly correlated with correlation invariably decreasing as the distance between points increases. Such a covariance function has the property of Automatic Relevance Determination (ARD) \citep{neal2012bayesian}. Lastly, the kernel variance $\sigma^{2}$ controls the variance of the process, more specifically the amplitude of function samples.

A GP has the following joint distribution over finite subsets $\mathbb{X}_{1} \in \mathbb{X}$ with function values $f(X_{1}) \in \mathbb{Y}$. Analogously for $\mathbb{X}_{2}$, with their union being denoted as $x = \{x_{1}, \cdots, x_{n} \}$.
\begin{equation}
    \begin{pmatrix} 
        f(x_{1})     \\
        f(x_{2})
        \end{pmatrix} = \mathcal{N}\left[
        \begin{pmatrix} 
            m(x_{1})     \\
            m(x_{2})
        \end{pmatrix}
        ,
        \begin{pmatrix} 
            k(x_{1},x_{1}),  k(x_{1},x_{2})   \\
            k(x_{2},x_{1}),  k(x_{2},x_{2})
        \end{pmatrix}        
        \right] \label{eqn:cite_this_now}
\end{equation}

The following observation model is used:
\begin{equation}
    p(y|f,x) = \prod_{i=1}^{N}p(y_{n}|f(x_{n}))
\end{equation}
, where given a supervised learning scenario, the dataset $D = \{x_{i},y_{i} \}_{i=1,\cdots,n}$ can be shorthand denoted as $D = \{x, y\}$. In the case of probabilistic regression, we make the assumption that the noise is additive, independent and Gaussian, such that the latent function $f(x)$ and the observed noisy outputs $y$ are defined by the following equation:
\begin{equation}
    y_{i} = f(x_{i}) +\epsilon_{i} \text{, where }~ \epsilon_{i} \sim \mathcal{N}\left(0, \sigma^{2}_{noise} \right)
\end{equation}

To train a GP for regression tasks, one performs Marginal Likelihood Maximization of Type 2 over the following equation:
\begin{align}
    p(y) &= \mathcal{N}\left(y \mid m, K_{ff}+\sigma^{2}_{noise}\mathbb{I}_{n} \right) \\
    &\propto - \left(y - m \right)^{\top}\left(K_{ff} + \sigma^{2}_{noise}\mathbb{I}_{n} \right)^{-1}\left(y - m \right)- \log{\mid K_{ff} + \sigma^{2}_{noise}\mathbb{I}_{n} \mid }
\end{align}
by treating the kernel hyperparameters as point-mass. 

\begin{figure}[!htb]
%  \vskip 0.2in
  \centering
    \includegraphics[width=\linewidth]{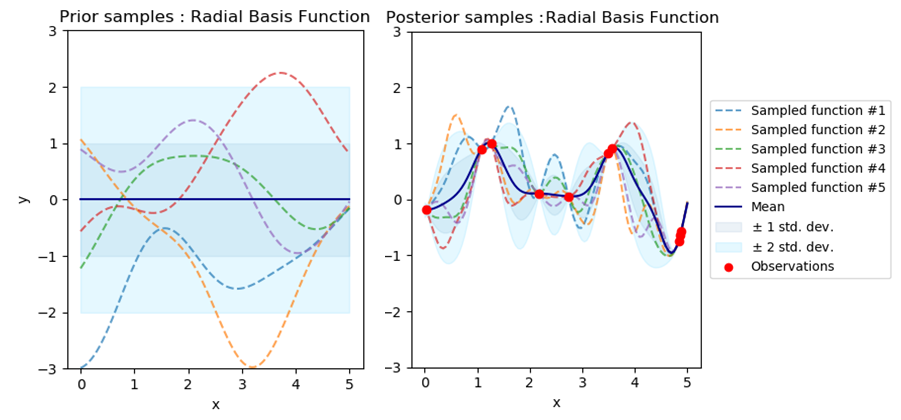}
    \caption{\textbf{Left :} GP prior samples using radial basis function kernel, which ensures a smooth function space hypothesis space; \textbf{Right :} GP samples conditioned on observations using radial basis function kernel. Predictive variance increases as input is further away from observations.}
    \label{fig:gp_properties}  
%    \vskip -0.2in
\end{figure}

We are interested in finding the posterior $p\left(f(x^{*}) \mid y \right)$ since the goal is to predict for unseen data points $x^{*}$ which are different than the training set. We know that the joint prior over training set observations and testing set latent functions is given by:
\begin{equation}
    \begin{pmatrix} 
        y     \\
        f(x^{*})
        \end{pmatrix} = \mathcal{N}\left[
        \begin{pmatrix} 
            m(x)     \\
            m(x^{*})
        \end{pmatrix}
        ,
        \begin{pmatrix} 
            k(x,x) + \sigma^{2}_{noise}\mathbb{I}_{n} &  k(x, x^{*})   \\
            k(x^{*},x) &  k(x^{*},x^{*})
        \end{pmatrix}        
        \right]
\end{equation}
Now we can simply apply the conditional rule for multivariate Gaussians to obtain:
\begin{align}
    p\left(f(x^{*}) \mid y \right) &= \mathcal{N}(f(x^{*} \mid m(x^{*}) +K_{f^{*}f}\left[K_{ff}+\sigma^{2}_{noise}\mathbb{I}_{n}\right]^{-1}\left[ y -m(x)\right], \\ & \nonumber \hspace{1cm} K_{f^{*}f^{*}} - K_{f^{*}f}\left[K_{ff}+\sigma^{2}_{noise}\mathbb{I}_{n}\right]^{-1}K_{ff^{*}})
\end{align}
An illustration of this predictive distribution is given in Figure \ref{fig:gp_properties}.

\paragraph{GP predictive variance as distributional uncertainty.}
GPs are clearly \emph{distance-aware} provided we use a translation-invariant kernel. The summary statistics (see definition \ref{def:distance_awareness}) for an unseen point is given by $u(x^{*}) = K_{f^{*}f^{*}} - K_{f^{*}f}K_{ff}^{-1}K_{ff^{*}}$, which is monotonically increasing as a function of distance (see Figure \ref{fig:gp_properties}). Throughout this paper, we will use the non-parametric variance of sparse variants of GPs as a proxy for distributional uncertainty, which will be used to assess if inputs are in or outside the distribution.

The usage of GP in real-world datasets is hindered by matrix inversion operations which have $\mathbb{O}(n^{3})$ time, $\mathbb{O}(n^{2})$ memory for training, where $n$ is the number of data points in the training set. In the next subsection we will see how to avert having to incur these expensive computational budgets.

\subsection{Sparse Variational Gaussian Processes} \label{sec:svgp}

In this subsection we succinctly review commonly used probabilistic sparse approximations for Gaussian process regression. \cite{quinonero2005unifying} provides a unifying view of sparse approximations by placing each method into a common framework of analyzing their posterior and their \emph{effective prior}, which will be shortly defined.

One way to avert the computationally expensive operators associated to the $K_{ff}$ matrix is to modify the joint prior over $p(f,f^{*})$ so that the respective terms depend on a matrix of lower rank, where the joint prior is defined as:
\begin{equation}
    \begin{pmatrix} 
        f(x)     \\
        f(x^{*})
        \end{pmatrix} = \mathcal{N}\left[
        \begin{pmatrix} 
            m(x)     \\
            m(x^{*})
        \end{pmatrix}
        ,
        \begin{pmatrix} 
            k(x,x),  k(x,x^{*})   \\
            k(x^{*},x),  k(x^{*},x^{*})
        \end{pmatrix}        
        \right]
\end{equation}

We introduce an additional set of M latent variables $U = \left[U_{1},\cdots,U_{m} \right] \in \mathbb{Y}$ with associated input locations $Z = \left[Z_{1},\cdots, Z_{m} \right] \in \mathbb{X}$. Throughout this paper, the former will be entitled inducing point values, respectively the latter inducing point locations.

Due to the consistency property of Gaussian Processes (i.e., for probabilistic model as defined in equation \eqref{eqn:cite_this_now} we have $p\left(f(x_{1})\right) = \int p\left(f(x) \right)~df(x_{2})$, which ensures that if we marginalize a subset of elements, the remainder will remain unchanged.), one can marginalize out $U$ to recover the initial joint prior over $p(f,f^{*})$:
\begin{equation}
    p(f,f^{*}) = \int p(f,f^{*},U) dU = \int p(f,f^{*} \mid U)p(U)~dU
\end{equation}
, where $p(U) = \mathcal{N}\left(U \mid 0,K_{uu}\right)$ and $K_{uu}$ is the kernel covariance matrix evaluated at $Z$.

All sparse approximations to GPs originate from the following approximation:
\begin{equation}
    p(f,f^{*}) \approx q(f,f^{*}) = \int q(f^{*} \mid U)q(f \mid U) p(U)dU
\end{equation}
which translates into a conditional independency between training and testing latent variables given $U$. Intuitively, the name "inducing points" for $\{Z, U\}$ was given for this precise property, that $U$ induces the values for the training and testing set.

\cite{titsias2009variational} introduced the first variational lower bound comprising a probabilistic regressiom model over inducing points. More specifically, the authors applied variational inference in an augmented probability space that comprises training set latent function values $F$ alongside inducing point latent function values $U$, more specifically using the following generative process in the case of a regression task:
\begin{align}
    p\left( U \right) &= \mathcal{N}\left(U \mid 0, K_{uu}; Z \right) \\
    p\left( F \mid U \right) &= \mathcal{N}\left(F \mid K_{fu}K_{uu}^{-1}U, K_{ff} - Q_{ff}; Z,X \right) \\
    p\left(y \mid F \right) &= \mathcal{N}\left(y \mid F, \sigma^{2}_{noise} \right)
\end{align}
, where $Q_{ff} = K_{fu}K_{uu}^{-1}K_{uf}$. We explicitly denoted the dependence of either $Z$ or $X$, however for decluttering reasons these notations will be dropped unless its not evident on what certain distributions depend on.

In terms of doing exact inference in this new model, respectively computing the posterior $p(f|y)$ and the marginal likelihood $p(y)$, it remains unchanged even with the augmentation of the probability space by $U$ as we can marginalize $p(F) = \int p(F,U)dU$ due to the marginalization properties of Gaussian processes. Succintely, $p(F)$ is not changed by modifying the values of $U$, even though $p(F|U)$ and $p(U)$  do indeed change. This translates into the fundamental difference between variational parameters $U$ and hyperparameters of the model $\{\sigma^{2}_{noise}, \sigma^{2}, l^{2}_{1}, \cdots, l^{2}_{D}  \}$, whereby the introduction of more variational parameters does not change the fundamental definition of the model before probability space augmentation.

Stochastic Variational Inference (SVI) \citep{hoffman2013stochastic} enables the application of VI for extremely large datasets, by virtue of performing inference over a set of global variables, which induce a factorisation in the observations and latent variables, such as in the Bayesian formulation of Neural Networks with distributions (implicit or explicit) over matrix weights. GP do no exhibit these properties, but by virtue of the approximate prior over testing and training latent functions for SGP approximations with inducing points $U$, which we remind here:
\begin{equation}
    p(f,f^{*}) \approx q(f,f^{*}) = \int p(f\mid U) p(f^{*}\mid U) p(U)~dU
\end{equation}
this translates into a fully factorized model with respect to observations at training and testing time, conditioned on the global variables $U$. 

 Our goal is to approximate the true posterior distribution $p(F,U \mid y) = p(F \mid U,Y)p(U \mid Y)$ by introducing a variational distribution $q(F,U)$ and minimizing the Kullback-Leibler divergence:
\begin{equation}
    KL\left[ q(F,U) \| p(F,U\mid y)\right] = \int q(F,U) \log \frac{q(F,U)}{p(F,U\mid y)}~dF~dU
\end{equation}
, where the approximate posterior factorized as $q(F,U) = p(F\mid U)q(U)$ and $q(U)$ is an unconstrained variational distribution over $U$. Following the standard VI framework we need to maximize the following variational lower bound on the log marginal likelihood:
\begin{align}
    \log p(y) &\geq \int p(F\mid U)q(U) \log\frac{p(y \mid F)p(F\mid U)p(U)}{p(F\mid U)p(U)}~dF~dU \\
    &\geq \int q(U) \left[\int \log p(Y\mid F)p(F\mid U)~dF + \log\frac{p(U)}{q(U)}\right]~dU
\end{align}
We can now solve for the integral over $F$:
\begin{align}
    \int \log p(y|F) p(F|U)~dF &=  \mathbb{E}_{p(F|U)} \left[-\frac{n}{2}\log (2\pi \sigma^{2}_{noise}) - \frac{1}{2\sigma^{2}_{noise}}Tr\left[yy^{\top} - 2yF^{\top} + FF^{\top}\right] \right] \\
    &= -\frac{n}{2}\log (2\pi \sigma^{2}_{noise}) - \frac{1}{2\sigma^{2}_{noise}}Tr[yy^{\top}  - 2y\left(K_{fu}K_{uu}^{-1}U\right)^{\top} + \\ & \nonumber \hspace{1cm} \left(K_{fu}K_{uu}^{-1}U\right)\left(K_{fu}K_{uu}^{-1}U\right)^{\top} + K_{ff} - Q_{ff} ] \\
    &= \log \mathcal{N}\left(y|K_{fu}K_{uu}^{-1}U, \sigma^{2}_{noise}\mathbb{I}_{n} \right) - \frac{1}{2\sigma^{2}_{noise}}Tr\left[ K_{ff} - Q_{ff} \right]
\end{align}
We can now rewrite our variational lower bound as follows:
\begin{equation}
    \log p(y) \geq \int q(U) \log \frac{\mathcal{N}\left(y \mid K_{fu}K_{uu}^{-1}U, \sigma^{2}_{noise}\mathbb{I}_{n} \right) p(U)}{q(U)}~dU - \frac{1}{2\sigma^{2}_{noise}}Tr\left[ K_{ff} - Q_{ff} \right]
\end{equation}

\begin{figure}[!htb]
    \centering
    \subfigure[SVGP]{\includegraphics[width=0.48\linewidth]{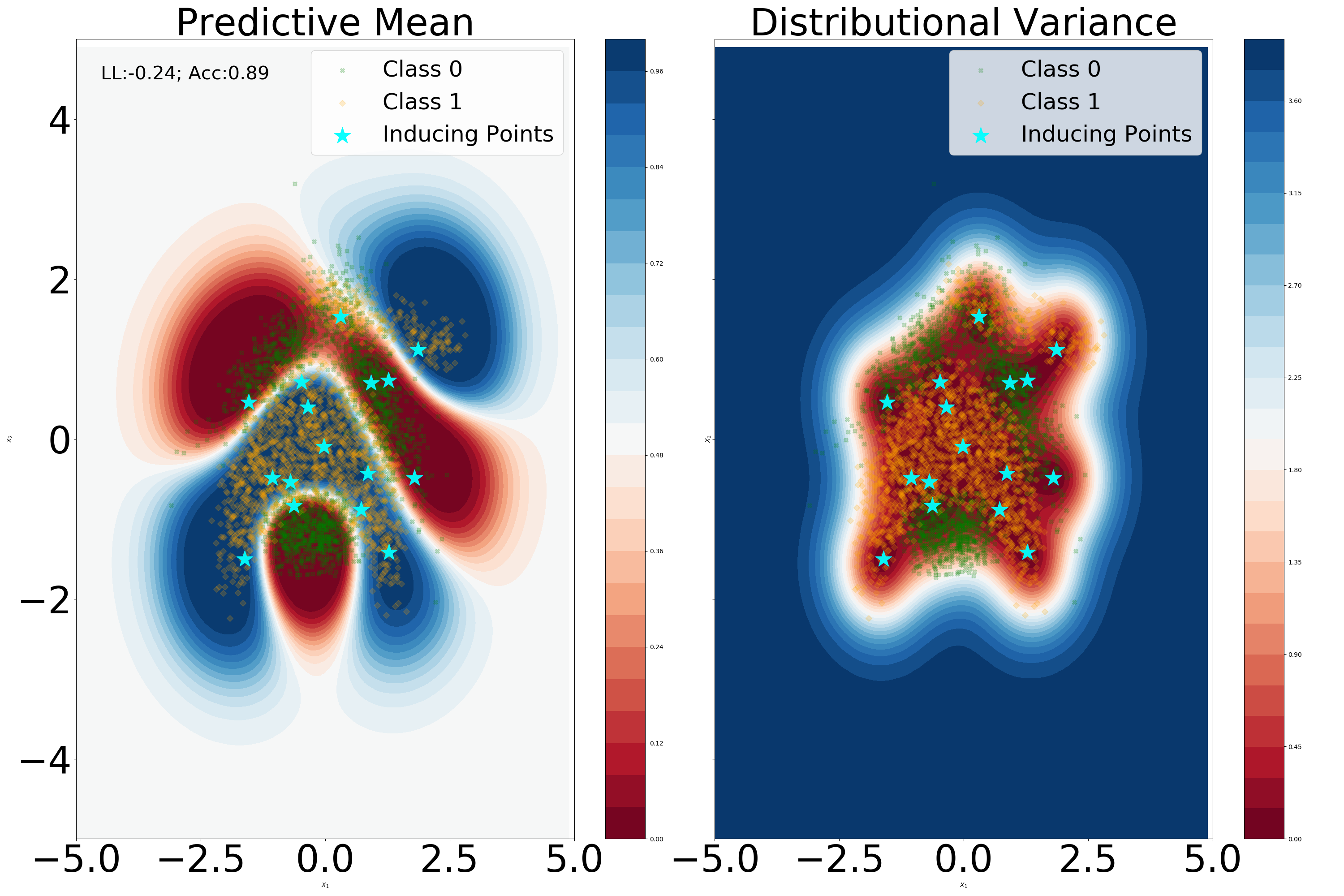}}
    \quad
    \subfigure[GP]{\includegraphics[width=0.48\linewidth]{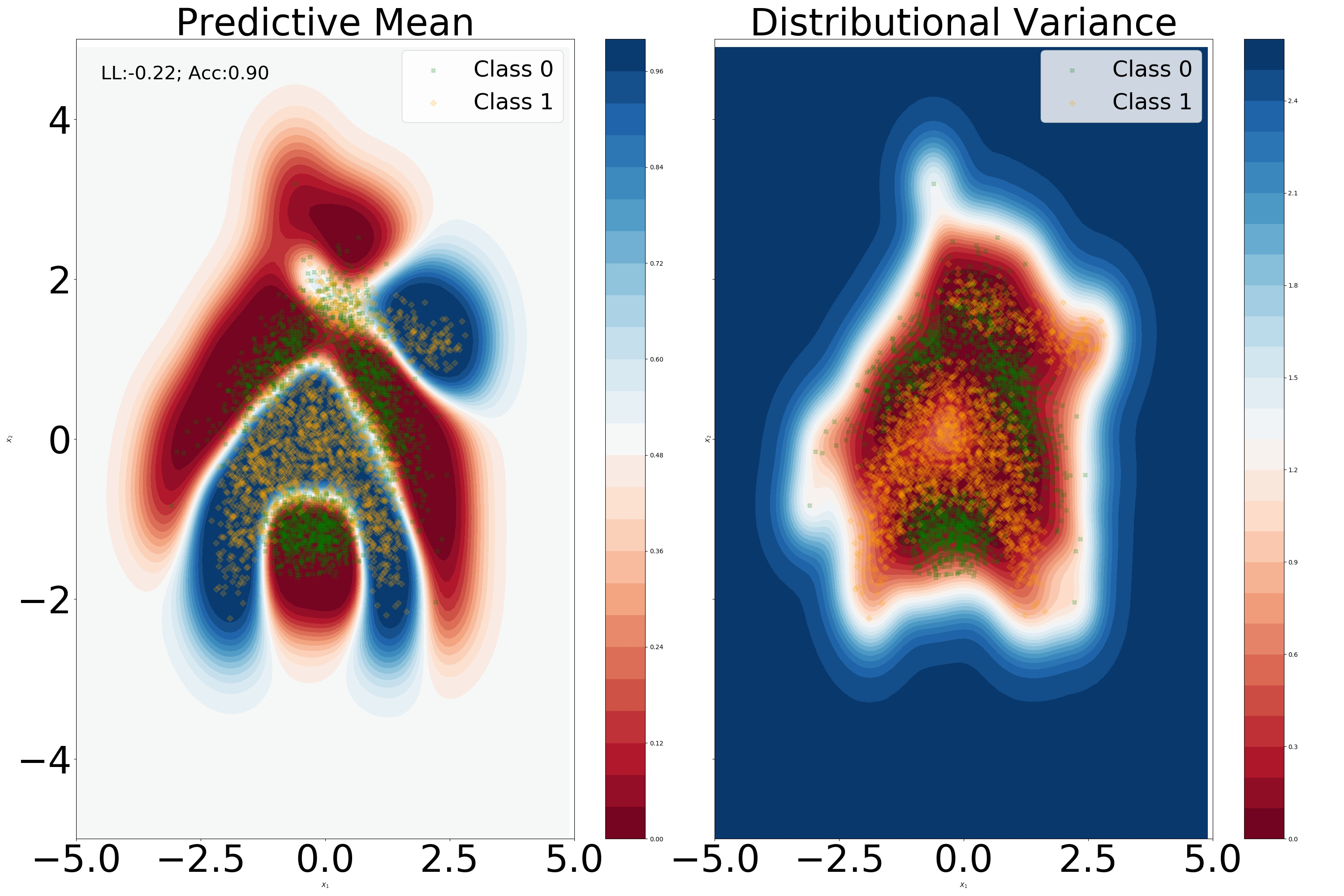}}
  \caption{\textbf{Left}: Predictive mean and variance of SVGP. Inducing points (teal stars) are tasked to compress the information present in the entire training set such that predictive equations conditioned on them are similar to ones conditioned on entire training set;  \textbf{Right}: Predictive mean and variance of GP. Not all training points are crucial in devising the decision boundary.}
  \label{fig:svgp_vs_gp}
\end{figure}

The variational posterior is explicit in this case, respectively $q(F,U) = p(F\mid U;X,Z)q(U)$, where $q(U)= \mathcal{N}(U \mid m_{U},S_{U})$. Here, $m_{U}$ and $S_{U}$ are free variational parameters. Due to the Gaussian nature of both terms we can marginalize $U$ to arrive at $q(F) = \int p(F\mid U)q(U) = \mathcal{N}(F\mid \tilde{U}(x), \tilde{\Sigma}(x) )$, where:
\begin{align}
    \tilde{U}(x) &= K_{fu}K_{uu}^{-1}m_{U} \label{eqn:posterior_mean_svgp} \\
    \tilde{\Sigma}(x) &= K_{ff} - K_{fu}K_{uu}^{-1} \left[ K_{uu}-S_{U} \right] K_{uu}^{-1}K_{uf} \label{eqn:posterior_variance_svgp}
\end{align}
The lower bound can be re-expressed as follows:
\begin{equation}
    \log p(y) \geq \int q(U) \log \mathcal{N}_{y}\left(K_{fu}K_{uu}^{-1}U, \sigma^{2}_{noise}\mathbb{I}_{n} \right) ~dU -KL\left[q(U) \| p(U)\right] -  \frac{1}{2\sigma^{2}_{noise}}Tr\left[ K_{ff} - Q_{ff} \right]
\end{equation}
We proceed to integrate out $U$, arriving at the following lower bound:
\begin{align}
    \mathcal{L}_{SVGP} &=  \mathcal{N}\left(y \mid K_{fu}K_{uu}^{-1}m_{U}, \sigma^{2}_{noise}\mathbb{I}_{n} \right) -  \frac{1}{2\sigma^{2}_{noise}}Tr\left[ K_{fu}K_{uu}^{-1}S_{U}K_{uu}^{-1}K_{uf} \right]  \\ & \nonumber -  \frac{1}{2\sigma^{2}_{noise}}Tr\left[ K_{ff} - Q_{ff} \right] - KL\left[q(U) \| p(U)\right]
\end{align}
, where we can easily see that the last equation is factorized with respect to individual observations. This lower variational bound will be denoted as the sparse variational GP (SVGP). This bound is maximized with respect to variational parameters $U$ and hyperparameters of the model $\{Z, \sigma^{2}_{noise}, \sigma^{2}, l^{2}_{1}, \cdots, l^{2}_{D}  \}$. An illustration of SVGP trained on the ``banana'' dataset is given in Figure \ref{fig:svgp_vs_gp}, showing similar behaviour to a GP only using a fraction of training set to obtain similar predictive distribution at testing time.

\subsection{Uncertainty decomposition in SVGP through evidential learning lens } \label{sec:evidential_learning}

In subsection \ref{sec:motivation_dist_unc} we have introduced the rationale behind the uncertainty decomposition framework introduced in \cite{malinin2018predictive}. We now expand on this topic on how to separate uncertainties in deep evidential learning models \citep{amini2019deep} and make an analogy to how uncertainties are decomposed in SVGP.

For multi-class classification tasks, evidential learning directly parametrizes predictive distributions over the probability simplex. Hence, in comparison to Bayesian Deep Learning or Deep Ensembles it averts parametrizing the logit space, subsequently feeding it through a softmax function. Dirichlet distributions provide an obvious choice for defining a distribution over the $K-1$ dimensional probability simplex, having the following p.d.f.: $Dir\left( \mu, \alpha \right) = \frac{1}{\beta(\alpha)} \prod\limits_{c=1}^{K} \mu_{c}^{\alpha_{c}-1}$, where $\beta(\alpha) = \frac{\prod\limits_{c=1}^{K}\Gamma(\alpha_{c}) }{\Gamma(\alpha_{0})}$ and $\alpha_{0} = \sum\limits_{c=1}^{K}\alpha_{c}$ with $\alpha_{c} \geq 0$. $\alpha_{0}$ is called the precision, being similar to the precision of a Gaussian distribution, where larger $\alpha_{0}$ will indicate a sharper distribution.

Dirichlet networks involve having a NN predict the concentration parameters of the Dirichlet distribution $\alpha = f_{\theta}(x)$, where predictions are made as follows: $\tilde{y} = \underset{c}{\arg \max }\{ \frac{\alpha_{c}}{\alpha_{0}} \}_{c=1}^{K}$.

We remind ourselves the uncertainty decomposition framework laid out in subsection \ref{sec:motivation_dist_unc}:
\begin{equation}
    \underbrace{I[y,\mu|x^{*},\mathbf{D}]}_{\text{Distributional~Uncertainty}} = \underbrace{H[\mathbb{E}_{p(\mu|\mathbf{D})}p(y|x^{*},\mu)]}_{\text{Total~Uncertainty}} - 
    \underbrace{\mathbb{E}_{p(\mu|\mathbf{D})}[H[P(y|x^{*},\mu)]]}_{\text{Aleatoric~Uncertainty}}
\end{equation}
In the case of Dirichlet networks these uncertainty measures have analytic formulas:
\begin{align}
    \mathbb{E}_{p(\mu|\mathbf{D})}[H[P(y|x^{*},\mu)]] &= -\sum\limits_{c=1}^{K} \frac{\alpha_{c}}{\alpha_{0}} \left[ \psi (\alpha_{c} + 1) - \psi(\alpha_{0} +1) \right] \label{eqn:aleatoric_uncertainty_dirichlet} \\
    I[y,\mu|x^{*},\mathbf{D}] &=  -\sum\limits_{c=1}^{K} \frac{\alpha_{c}}{\alpha_{0}} \left[ \log{\frac{\alpha_{c}}{\alpha_{0}}} -\psi (\alpha_{c} + 1) + \psi(\alpha_{0} +1) \right] \label{eqn:distributional_uncertainty_dirichlet}
\end{align}
, where $\psi$ is the digamma function. Epistemic uncertainty quantifies the spread in the Dirichlet distribution, hence $\alpha_{0}$ can be used to measure it \citep{charpentier2021natural}.

Exact inference is not tractable in GP on classification tasks due to the non-conjugacy between the GP prior
and the non-Gaussian likelihood (Categorical or Bernoulli). Therefore, approximation are required such as the Laplace approximation \citep{williams1998bayesian}, Expectation Propagation \citep{minka2013expectation} or VI \citep{hensman2015scalable}. \cite{milios2018dirichlet} have proposed a method that circumvents these approximate inference techniques by re-branding the classification problem into a regression one, for which exact inference is possible. We commence to briefly lay out the Dirichlet-based GP Classification algorithm.

We consider the probability simplex $\pi = [\pi_{1}, \cdots, \pi_{K} ] \sim Dir(\alpha)$. We can transform a multi-class classification task into a multi regression scenario where if $y_{c} = 1$ in a one-hot-encoding, then we can assign $\alpha_{c} = 1 + \alpha_{\epsilon}$, respectively $\alpha_{c} = \alpha_{\epsilon}$ for $0 \leq \alpha_{\epsilon} << 1$. The model has the following generative process:
\begin{align}
    \pi &\sim Dir(\alpha) \\
    p(y \mid \alpha) &= Cat(\pi)
 \end{align}
To sample from the Dirichlet distribution we use the following routine: $\pi_{c} = \frac{x_{c}}{\sum\limits_{k=1}^{K}x_{k}}$ with $x_{c} \sim \Gamma(\alpha_{c}, 1)$ following the Gamma distribution. From this sampling procedure, we can see that the generative process translates to independent Gamma likelihoods for each class. Intuitively, at this point in the derivation we need a GP to produce $x_{c} \geq 0$, since Gamma distributions are only defined on $\mathbb{R}^{+}$. Since the marginal GP over a subset of data is governed by a multivariate normal it will not satisfy this constraint. To obtain GP sampled functions that respect this constraints, we can use an $\exp$ function to transform it. With this in mind, we know that $x \sim \text{log-normal}\left(x \mid \mu, \sigma^{2} \right) \overset{d}{=} exp{\left(\mathcal{N}\left(x \mid \mu, \sigma^{2} \right)\right)}$, with $\mathbb{E}\left[ x \right] = \exp{\left( \mu + \frac{\sigma^{2}}{2} \right)}$ and $V\left[ x \right] = \left[\exp{\sigma^{2}} -1 \right] \exp{\left( 2\mu + \sigma^{2}\right)}$. Hence, we can approximate $x_{c} \sim \gamma(\alpha_{c}, 1) $ with $\tilde{x_{c}} \sim \text{log-normal}\left( \tilde{\mu_{c}}, \tilde{\sigma^{2}_{c}} \right)$. To ensure a good approximation, the authors in \cite{milios2018dirichlet} propose using moment matching:
\begin{align}
    \mathbb{E}\left[ x_{c} \right] = \alpha_{c} &= \exp{\left( \tilde{\mu_{c}}, \frac{\tilde{\sigma^{2}_{c}}}{2} \right)} = \mathbb{E}\left[ \tilde{x_{c}} \right] \\
    V\left[ x_{c} \right] = \alpha_{c} &= \left[\exp{\tilde{\sigma^{2}_{c}}} -1 \right] \exp{\left[2\tilde{\mu_{c}} + \tilde{\sigma^{2}_{c}} \right]} = V\left[ \tilde{x_{c}} \right]
\end{align}
with equality if $\tilde{\mu_{c}} = \log{\alpha_{c}} - \frac{\tilde{\sigma^{2}_{c}}}{2}$ and $\tilde{\sigma^{2}_{c}} = \log{\left( \frac{1}{\alpha_{c}} + 1 \right)}$. We can re-express this approximation by taking a natural logarithm, obtaining $\log{\tilde{x_{c}}} \sim \mathcal{N}\left(\tilde{\mu_{c}}, \tilde{\sigma^{2}_{c}} \right)$. This translates into a heteroskedastic regression model $\tilde{\mu_{c}} = f_{c} + \mathcal{N}\left(0, \tilde{\sigma^{2}_{c}}\right)$, where $f_{c} \sim GP\left(0, K_{ff} \right)$. Hence, one can now apply the standard inference scheme for full GP or we can sparsify the model and apply the SVGP framework. At testing time, the expectation of class probabilities will be:
\begin{equation}
    \mathbb{E}\left[ \pi_{i,c} \right] = \int \frac{\exp{f_{i,c}}}{\sum\limits_{k=1}^{C} \exp{f_{i,k}}} q(f_{i,c})~df_{i,c}
\end{equation}
which can be approximated via Monte Carlo integration. In the sparse scenario, $q(f_{i,c}) \sim \mathcal{N}\left(\tilde{U}(x_{i}), \tilde{\Sigma}(x_{i}) \right)$ similar to the predictive equations introduced in subsection \ref{sec:svgp}. In conclusion, if using Dirichlet-based GP for Classification one can obtain similar estimates of aleatoric and distributional uncertainty in the space of the probability simplex as in equations \eqref{eqn:aleatoric_uncertainty_dirichlet} and \eqref{eqn:distributional_uncertainty_dirichlet} specific to Dirichlet Networks. However, for the purposes of this paper we intend to measure distributional uncertainty in the space of logits, as the formulas are simpler to compute and more intuitive from the viewpoint of \emph{distance-awareness}.

As we have previously stated, GPs are \emph{distance-aware}. Thus, they can reliably notice departures from the training set manifold. For SVGP we decompose the model uncertainty into two components:
\begin{align}
    h(\cdot) &=  \mathcal{N}(h \mid 0, K_{ff} - K_{fu}K_{uu}^{-1}K_{uf}) \\  
    g(\cdot) &= \mathcal{N}(g \mid K_{fu}K_{uu}^{-1}m_{U}, K_{fu}K_{uu}^{-1} S_{U} K_{uu}^{-1}K_{uf})
\end{align}
The $h(\cdot)$ variance captures the shift from within to outside the data manifold and will be denoted as \emph{distributional uncertainty}. The variance $g(\cdot)$ is termed here as \emph{within-data uncertainty} and encapsulates uncertainty present inside the data manifold. A visual depiction of the two is provided in Figure \ref{fig:schematic_seg_net_and_uncertainties} (bottom). To capture the overall uncertainty in $h(\cdot)$, thereby also capturing the spread of samples from it, we can calculate it's differential entropy as:
\begin{equation}
    h(h) = \frac{n}{2}\log{2\pi} + \frac{1}{2}\log{\mid K_{ff} - K_{fu}K_{uu}^{-1}K_{uf} \mid } + \frac{1}{2}   
\end{equation}
In practice we only use the diagonal terms of the Schur complement, hence the log determinant term will considerably simplify. Intuitively, if terms on the diagonal of the Schur complement have higher values, so will the distributional differential entropy. This OOD measure in logit space will be used throughout the rest of the paper.

\subsection{Deep Gaussian Processes fail in propagating distributional uncertainty}

Deep Gaussian Processes (DGP) were first introduced in \cite{damianou2013deep}, as a multi-layered hierarchical formulation of GPs. Composition of processes has retains theoretical properties of underlying stochastic process (such as Kolmogorov extension theorem) while also ensuring a more diverse hypothesis space of process priors, or at least in theory as we shall later see.

We can view the DGP as a composition of functions, keeping in mind that this is only one way of defining this class of probabilistic models \citep{dunlop2018deep}: 
\begin{equation}
    f_{L}(x) = f_{L} \circ ... \circ f_{1}(x) 
\end{equation}
with $f_{l} = \mathcal{GP}\left(m_{l}, k_{l}\left(\cdot, \cdot \right) \right)$.
Assuming a likelihood function we can write the joint prior as:
\begin{equation}
 p\left(y, \{f_{l} \}_{l=1}^{L}; X \right)=\underbrace{p(y \mid f_{L})}_{\text{likelihood}}\underbrace{\prod_{l=1}^{L} p(f_{l} \mid f_{l-1})}_{\text{prior}}
\end{equation}
with $p\left(f_{l} \mid f_{l-1}\right) \sim \mathcal{GP}\left(m_{l}(f_{l-1}), k_{l}\left(f_{l-1}, f_{l-1} \right) \right)$, where in the case we choose squared exponential kernels we have the following formula for the l-th layer:
\begin{equation}
    k^{SE}(f_{l,i},f_{l,j})_{l} = \sigma^{2}_{l} \exp{\left[\sum_{d=1}^{D_{l}}-\frac{\left(f_{l,i,d} - f_{l,j,d} \right)^{2}}{l^{2}_{l,d}} \right]} \nonumber
\end{equation}
where $D_{l}$ represents the number of dimensions of $F_{l}$ and we introduce layer specific kernel hyperparameters $\{\sigma^{2}_{l}, l^{2}_{l,1}, \cdots, l^{2}_{l,D_{l}}\}$.

Analytically integrating this Bayesian hierarchical model is intractable as it requires integrating Gaussians which are present in a non-linear way. Moreover, to enable faster inference over our model we can augment each layer $l$ with $M_{l}$ inducing points' locations $Z_{l-1}$, respectively inducing points' values $U_{l}$ resulting in the following augmented joint prior:
\begin{equation}
 p\left(y, \{f_{l} \}_{l=1}^{L}, \{U_{l} \}_{l=1}^{L}; X, \{Z_{l} \}_{l=0}^{L-1} \right) = \underbrace{p(y|f_{L})}_{\text{likelihood}}\underbrace{\prod_{l=1}^{L} p(f_{l}|f_{l-1},U_{l};Z_{l-1})p(U_{l})}_{\text{prior}}
\end{equation}
, where $p(f_{l} \mid f_{l-1},U_{l};Z_{l-1}) = \mathcal{N}\left(f_{l} \mid  m_{l}(f_{l-1}) + K_{fu}K_{uu}^{-1}\left(U_{l} - m_{l}(Z_{l-1}), K_{ff} - K_{fu}K_{uu}^{-1}K_{uf} \right)\right)$. To perform SVI we introduce a factorised variational approximate posterior $q\left( \{U_{l} \}_{l=1}^{L}\right) = \prod\limits_{l=1}^{L} \mathcal{N}\left(U_{l} \mid m_{U_{l}}, S_{U_{l}} \right)$. Using a similar derivation as in the uncollapsed evidence lower bound for SVGPs, we can arrive at our ELBO for DGPs:
\begin{equation}
    \mathcal{L}_{DGP} = \mathbb{E}_{q(\{ f_{l} \}_{l=1}^{L})}\left[\log{p\left( y \mid f_{L}\right)} \right] - \sum\limits_{l=1}^{L}KL\left[q(U_{l}) \| p(U_{l}) \right]
\end{equation}
where $q\left( \{f_{l} \}_{l=1}^{L} \right) = \prod\limits_{l=1}^{L} q\left(f_{l} \mid f_{l-1} \right)$ and $q\left(f_{l} \mid f_{l-1} \right) = \mathcal{N}\left(f_{l} \mid \tilde{U_{l}}(f_{l-1}), \tilde{\Sigma_{l}}(f_{l-1}) \right)$, respectively:
\begin{align}
    \tilde{U_{l}}(f_{l-1}) &= m_{l}(f_{l-1}) + K_{fu}K_{uu}^{-1}\left[m_{U_{l}} - m_{l}(Z_{l-1})\right] \\
    \tilde{\Sigma_{l}}(f_{l-1}) &= K_{ff} - K_{fu}K_{uu}^{-1}\left[K_{uu} - S_{U_{l}} \right]K_{uu}^{-1}K_{uf} 
\end{align}
This composition of functions is approximated via Monte Carlo integration as introduced in the doubly stochastic variational inference framework for training DGPs \citep{salimbeni2017doubly}.

In \cite{popescu2020hierarchical} the authors argued that total uncertainty in the hidden layers of a DGP will be higher for OOD data points in comparison to in-distribution data points only under a set of conditions. We briefly lay out the details here.

Without loss of generality for deeper architectures, we can consider the case of a DGP with two layers and zero mean functions which has the following posterior predictive equation:
\begin{equation}
    q(F_{2})(x)  = \int p(F_{2}|F_{1}) q(F_{1}(x))dF_{1}
\end{equation}
, where $q(F_{1}(x)) = \mathcal{N}_{f_{1}}\left(\tilde{U_{1}}(x), \tilde{\Sigma_{1}}(x) \right)$. This is similar to the case of approximating GPs with uncertain inputs, in this case Multivariate Normals. In \cite{girard2004approximate} they lay out a framework for obtaining Gaussian approxiations of GPs with uncertain inputs (in our case the uncertainty stems from the previous layer of the DGP), which when adapted to our case we obtain the following approximate moments for $q(F_{2})(x)$:
\begin{align}
    m(F_{2}) &= \tilde{U}_{2}(\tilde{U}_{1}(x)) \\
    v(F_{2}) &=  \tilde{\Sigma}_{2}(\tilde{U}_{1}(x)) + \tilde{\Sigma}_{1}(x) \Bigg[ \frac{1}{2} \frac{\partial^{2} \tilde{\Sigma}_{2}(F_{1})}{\partial^{2} F_{1}}\Bigr|_{\substack{F_{1}=\tilde{U}_{1}(x)}}  + \left(\frac{\partial \tilde{U}_{2}(F_{1})}{\partial F_{1}}\right)^{2}\Bigr|_{\substack{F_{1}=\tilde{U}_{1}(x)}} \Bigg] \label{eqn:approximation_variance}
\end{align}

In \cite{popescu2020hierarchical} they propose a realistic scenario which occurs frequently in practice, whereby the inducing points $Z_{l}$ of particular layer are spread out such as to cover the entire spectrum of possible samples from the previous layer $F_{l-1}$. More precisely, we can consider an OOD data point $x_{ood}$ in input space such that $\tilde{\Sigma_{1}}(x_{ood}) = \sigma^{2}$ and $\tilde{U_{1}}(x_{ood}) = 0$, respectively an in-distribution point $x_{in-d}$ such that $\tilde{\Sigma}_{1}(x_{in-d}) = V_{in} \leq \sigma^{2} $ and $\tilde{U}_{1}(x_{in-d}) = M_{in}$. We also assume that $Z_{2}$ are equidistantly placed between $[-3\sigma, 3\sigma]$. The authors go on to show that the total variance in the second layer of $x_{ood}$ will be higher than $x_{in-d}$ if the following holds
$
    \frac{ \left(\frac{\partial \tilde{U}_{2}(F_{1})}{\partial F_{1}}\right)^{2}\Bigr|_{\substack{F_{1}=M_{in}}}}{\left(\frac{\partial \tilde{U}_{2}(F_{1})}{\partial F_{1}}\right)^{2}\Bigr|_{\substack{F_{1}=0}}} \leq \frac{\sigma^{2}}{V_{in}}
$. One can rapidly infer that this inequality holds if the absolute first order derivative of the parametric component of the SVGP around 0 is higher compared to any other value which might be evaluated at. This observation is to be made in conjunction with the fact that $\frac{\sigma^{2}}{V_{in}} \geq 1.0$, since the total variance of in-distribution points will be reduced compared to the prior variance.

\begin{figure}[!htb]
%  \vskip 0.2in
  \centering
    \includegraphics[width=\linewidth]{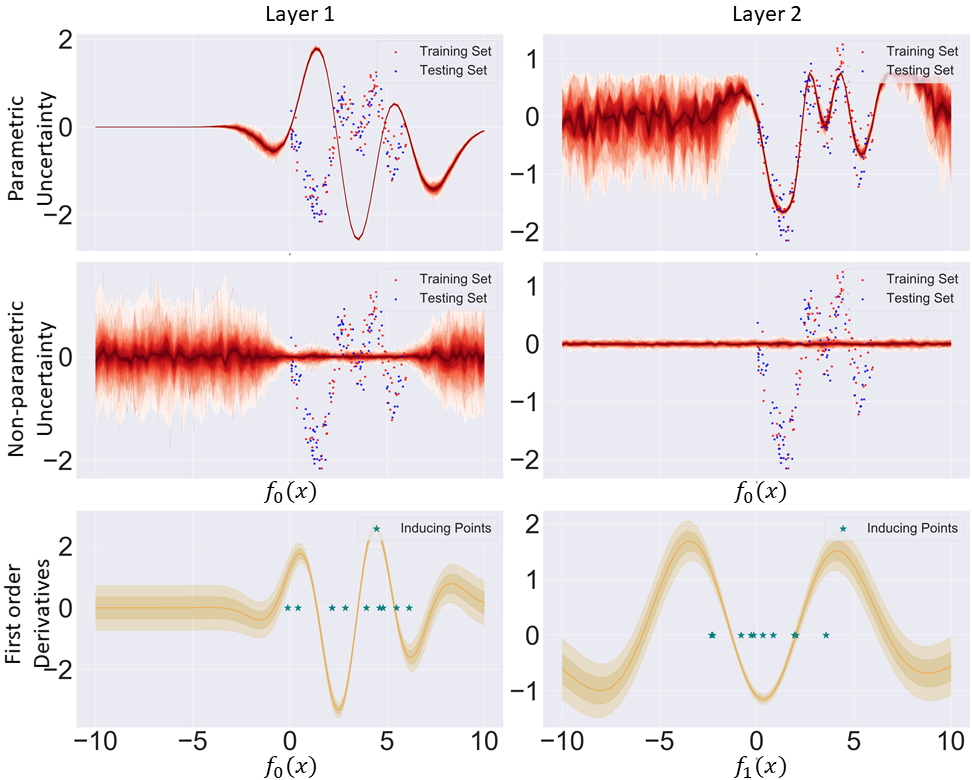}
    \caption{Layer-wise decomposition of uncertainty into parametric/epistemic and non-parametric/distributional for a zero mean function DGP, alongside first order derivatives. OOD points in input space $x_{\text{out}}$ get mapped on average to $0$ in $f_{1}(x_{\text{out}})$, which has a high absolute first order derivative causing the parametric uncertainty in $f_{2}$ to be high for $x_{\text{out}}$.}
    \label{fig:snelson_derivatives}  
%    \vskip -0.2in
\end{figure}

To gain some intuition as to what occurs in practice, we can consider a 2 layer DGP trained on a toy regression task, where we decompose the resulting posterior SVGP predictive equation into its parametric and non-parametric components for each layer with respect to input space (first two rows of Figure \ref{fig:snelson_derivatives}). To investigate whether our trained DGP respects the above inequality for propagating higher total uncertainty for OOD data points in comparison to in-distribution data points, we also need to predict what are the first-order derivatives with respect to the input stemming from the previous layer (last row of Figure \ref{fig:snelson_derivatives}). We encourage the reader to inspect \cite{mchutchon2013differentiating} for an in-depth introduction to first order derivative of GPs. We can notice that the total variance is indeed higher for OOD data points in the final layer, as this was brought upon by the high absolute value of the first order derivative around 0 in the second layer (OOD data points in the first hidden layer will have an expected value of 0). Intuitively, OOD data points in input space will have higher total uncertainty in output space due to the higher diversity of function values in the second layer. The diversity is caused by the high non-parametric uncertainty in the first hidden layer. Conversely, we can see that for in-distribution points the total variance in the first hidden layer is relatively small, hence the sampling will be close to deterministic, implicitly meaning that it will access only a very restricted set of function values in the second layer thus causing a relatively small total variance. Lastly, we remind ourselves that for GPs we can consider the non-parametric/distributional uncertainty as a proxy for OOD detection. From Figure 
\ref{fig:snelson_derivatives} we can see that distributional uncertainty collapses in the second layer for any value in input space. This implies that DGPs are not \emph{distance-aware}.

\begin{figure}[!htb]
%  \vskip 0.2in
  \centering
    \includegraphics[width=0.95\linewidth]{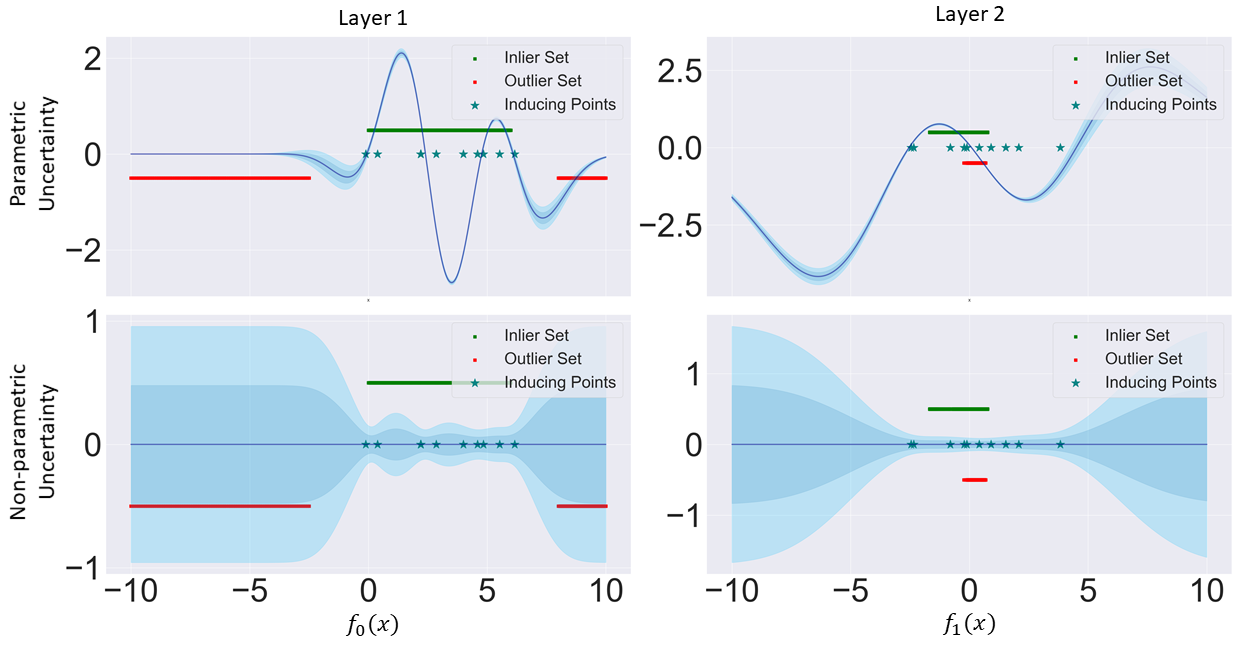}
    \caption{Layer-wise decomposition of uncertainty into parametric/epistemic and non-parametric/distributional for a zero mean function DGP. Outlier points are sampled close to inliers points in $f_{1}$, thereby causing their collapse of non-parametric variance since inducing points in $f_{1}$ are close to both outlier and inlier samples.}
    \label{fig:visual_explanation_collapse_of_variance}  
%    \vskip -0.2in
\end{figure}

To understand what is causing this pathology, we take a simple case study of a DGP (zero mean function) with two hidden layers trained on a toy regression dataset (Figure \ref{fig:visual_explanation_collapse_of_variance}). Taking a clear outlier in input space, say the data point situated at -7.5, it gets correctly identified as an outlier in the mapping from input space to hidden layer space as given by its distributional variance. However, its outlier property dissipates in the next layer after sampling, as it gets mapped to regions where the next GP assigns inducing point locations. This is due to points inside the data manifold getting confidently mapped between -2.0 and 1.0 in hidden layer space. Consequently, what was initially correctly identified as an outlier will now have its final distributional uncertainty close to zero. Adding further layers, will only compound this pathology.

\subsection{Wasserstein-2 kernels for probability measures}

As we have seen in the previous subsection, analytically integrating out the prior of a DGP is intractable in the case of using kernels operating in Euclidean space. However, the hidden layers of a DGP are intrinsically defined over probability measures (Gaussian in this case). This leads us to ponder whether we can obtain an analytically tractable formulation of DGPs by using kernels operating on probability measures, thereby we need a metric on probability measures which we subsequently introduce.

The Wasserstein space on $\mathbb{R}$ can be defined as the set $W_{2}(\mathbb{R})$ of probability measures on $\mathbb{R}$ with a finite moment of order two. We denote by $\Pi(\mu,\nu)$  the  set  of  all  probability measures $\Pi$ over the product set $\mathbb{R}\times\mathbb{R}$ with marginals $\mu$ and $\nu$, which are probability measures  in $W_{2}(\mathbb{R})$. The transportation cost between two measures $\mu$ and $\nu$ is defined as:
\begin{equation}
T_{2}(\mu,\nu) = \inf_{\pi\in\Pi(\mu,\nu)}\int[x-y]^{2}d\pi(x,y)
\end{equation}
This transportation cost allows us to endow the set $W_{2}(\mathbb{R})$ with a metric by defining the quadratic Wasserstein distance between
$\mu$ and $\nu$ as:
\begin{equation}
W_{2}(\mu,\nu) =T_{2}(\mu,\nu)^{1/2}
\end{equation}

\begin{theorem}[Theorem \RNum{4}.1. in \cite{bachoc2017gaussian}]
Let $k_{W} : W_{2}(\mathbb{R}) \times W_{2}(\mathbb{R}) \rightarrow \mathbb{R}$ be the Wasserstein-2 RBF kernel defined as following:
\begin{equation}
k^{W_{2}}(\mu,\nu) = \sigma^{2} \exp \frac{-W_{2}^{2}(\mu,\nu)}{l^{2}}
\end{equation}
then $k^{W_{2}}(\mu,\nu)$ is a positive definite kernel for any $\mu,\nu \in  W_{2}(\mathbb{R})$, respectively $\sigma^{2}$ is the kernel variance, $l^{2}$ being the lengthscale. 
\end{theorem}
A detailed proof of this theorem can be found in \cite{bachoc2017gaussian}.

Multiplication of positive definite kernels results again in a positive definite kernel, hence we arrive at the automatic relevance determination kernel based on Wasserstein-2 distances: 
\begin{equation}
    k^{W_{2}}([\mu_{d}]_{d=1}^D,[\nu_{d}]_{d=1}^D) = \sigma^{2} \exp \sum_{d=1}^D \frac{-W_{2}^{2}(\mu_{d},\nu_{d})}{l_{d}^{2}}\label{eqn:wasserstein_kernel}    
\end{equation}

\paragraph{Wasserstein-2 Distance between Gaussian distributions.} 

Gaussian measures fulfill the condition of finite second order moment, thereby being a clear example of probability measures for which we can compute Wasserstein metrics. The Wasserstein-2 distance between two multivariate Gaussian distributions $\mathcal{N}( m_{1},\Sigma_{1})$ and $\mathcal{N}( m_{2},\Sigma_{2})$, which have associated Gaussian measures and implicitly the Wasserstein metric is well defined for them, has been shown to have the following form
$\parallel m_{1} \minus m_{2} \parallel_{2}^{2}+
Tr\Big[\Sigma_{1} + \Sigma_{2} \minus 2\Big(\Sigma_{1}^{1/2} \Sigma_{2}\Sigma_{1}^{1/2}\Big)^{1/2}\Big]$ \citep{Dowson1982TheFD}, which in the case of univariate Gaussians simplifies to $| m_{1} - m_{2} |^{2}+|\sqrt{\Sigma_{1}} \minus \sqrt{\Sigma_{2}} |^{2}$. This last formulation will be used throughout this paper.

\subsection{Distributional Deep Gaussian Processes \& OOD detection} \label{sec:hybrid_kernel}

In the previous subsection we have seen that DGPs can easily fail in propagating distributional uncertainty forward. We now focus on the variant of DGPs introduced in \cite{popescu2020hierarchical} that was proven both theoretically and empirically to propagate distributional uncertainty throughout the hierarchy, thus ensuring \emph{distance-awareness} properties. The insights gained from this subsection will constitute the departure point for our proposed model in the next section.

Distributional Gaussian Processes (DistGP) were first introduced in \citep{bachoc2017gaussian} to describe a shallow GP that operates on probability measures using a Wasserstein-2 based kernel as defined in equation \eqref{eqn:wasserstein_kernel}.  

 We introduce the generative process of Distributional Deep Gaussian Processes (DDGP) for 2 layers:
\begin{align}
    p(F_{1}) &\sim \mathcal{N}\left(0, K_{ff} \right) \\
    F_{1}^{sth} &= m(F_{1}) + \sqrt{v(F_{1})} \epsilon, ~ \epsilon \sim \mathcal{N}\left(0, \mathbb{I}_{n} \right) \\
    F_{1}^{det} &= \mathcal{N}\left(m(F_{1}), diag\left[ v(F_{1}) \right] \right) \\
    p(F_{2}) &\sim \mathcal{N}\left(0, k_{hybrid}\left(\{F_{1}^{sth}, F_{1}^{det} \}, \{F_{1}^{sth},F_{1}^{det}\} \right) \right)
\end{align}
, where the hybrid kernel is defined as follows:
\begin{equation}
k^{hybrid}\left(\mu_{i},\mu_{j}\right) = k^{E}(x_{i},x_{j})\exp \sum_{d=1}^D\frac{-W_{2}^{2}(\mu_{i,d},\mu_{j,d})}{l_{d}^{2}} \label{eqn:hybrid_kernel}
\end{equation}
, where we denoted $\mu_{i} = \mathcal{N}(m(F_{1}(x_{i})), \sigma_{1}^{2})$; $\mu_{j} = \mathcal{N}(m(F_{1}(x_{j})),\sigma_{1}^{2})$ as the first two moments which are obtained through the $F_{1}^{det}$ operation in the generative process. Intuitively this generative process implies keeping track of a \emph{stochastic}, respectively \emph{deterministic} component of the same SVGP at any given hidden layer, while the first layered is governed by a standard SVGP operating on Euclidean data. It is worthy to point out that for this probabilistic construction, the inducing points $\{ Z_{l} \}_{l=1}^{L}$ have to reside in the space of multivariate Gaussians, hence $Z_{l} \sim \mathcal{N}\left( Z_{l} \mid \mu_{Z_{l}}, \Sigma_{Z_{l}} \right)$. The first two moments are treated as hyperparameters that are optimized during training.

\begin{figure}[!htb]
    %\vskip 0.2in
    \centering
    \includegraphics[width=\linewidth]{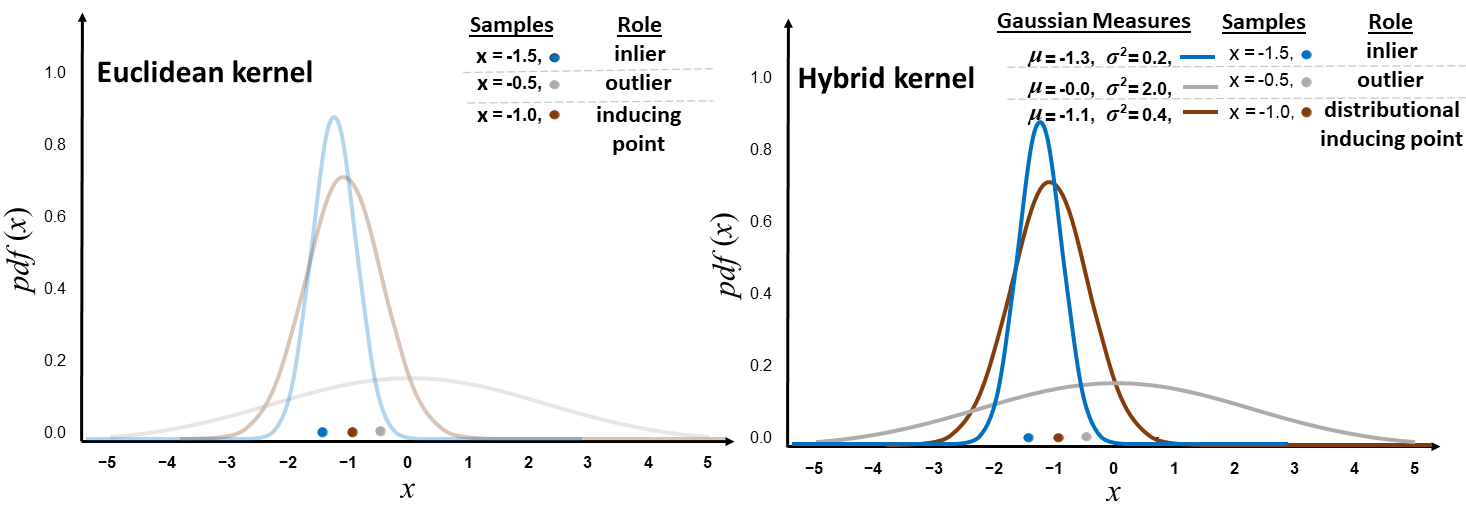}
    \caption{Conceptual difference between euclidean and hybrid kernel.}
    \label{fig:schematic_hybrid_kernel}  
    %\vskip -0.2in
\end{figure}

We can consider an OOD data point $x_{ood}$ in input space such that $\tilde{\Sigma_{1}}(x_{ood}) = \sigma^{2}$ and $\tilde{U_{1}}(x_{ood}) = 0$, respectively an in-distribution point $x_{in-d}$ such that $\tilde{\Sigma}_{1}(x_{in-d}) = V_{in} \leq \sigma^{2} $ and $\tilde{U}_{1}(x_{in-d}) = M_{in}$. We also assume that $Z_{2}$ are equidistantly placed between $[-3\sigma, 3\sigma]$. The authors go on to show that the total variance in the second layer of $x_{ood}$ will be higher than $x_{in-d}$ if the following holds $\sigma^{2} >> Z_{2,var}$  and  $V_{in} \approx Z_{2,var}$. To better understand this behaviour, we can consider a two-layered DGPs and DDGPs, we assume an in-distribution point to have low total variance in hidden layer $F_{1}$, respectively an OOD point to have high total variance. In the DGP case, upon sampling from $q(F_{1}(x_{in-d}))$  and  $q(F_{1}(x_{ood}))$ we can end up with samples which are equally distance with respect to inducing points' location $Z_{1}$. If this occurs, then non-parametric variance (proxy for distributional variance) will be equal for $x_{in-d}$ and $x_{ood}$ in $F_{2}$. Hence, what was initially flagged as OOD in the first hidden layer will be considered as in-distribution by the second hidden layer. onversely, in the DDGP case and under the assumption that the variance of distributional inducing points' locations $Z_{2}$ is almost equal in distribution to the total variance of in-distribution points in $F_{2}$, the Wasserstein-2 component of the hybrid kernel will notice that there is a higher distance between the now distributional inducing point location and $x_{ood}$, as opposed of former with $x_{in-d}$. Then, the non-parametric variance of $x_{ood}$ will be higher than that of $x_{in-d}$. A visual depiction of this case study is illustrated in Figure \ref{fig:schematic_hybrid_kernel}.

\section{Distributional GP Layers}

\paragraph{Shift towards single-pass uncertainty quantification}  
Early methods for uncertainty quantification in Bayesian deep learning (BDL) have focused on estimating the variance of sample from difference sub-models, such as in using dropout \citep{gal2016dropout}, deep ensembles \citep{lakshminarayanan2017simple} or in sampling posterior network weights from a hypernetwork \citep{pawlowski2017implicit}. This results in slow uncertainty estimation at testing time, which can be critical in high-risk domains where speed is of essence (e.g., self-driving cars). Recent work in OOD detection has focused on estimating proxies for distributional uncertainty in a single-pass, such as in bi-Lipschitz regularized feature extractors for GP \citep{van2021feature, liu2020simple} or in parametrizing second-order uncertainty via neural networks within the framework of evidential learning \citep{charpentier2020posterior, amini2019deep}. With this shift towards single-pass uncertainty quantification, DDGPs and the hybrid kernel introduced in subsection \ref{sec:hybrid_kernel} are no longer appropriate since they involved sampling the features at each hidden layer. In next subsection we detail a deterministic variant which still preserves correlations between data points in the hidden layers.

\paragraph{Integrating GPs in convolutional architectures}
GP for image classification has garnered interest in the past years, with hybrid approaches, whereby a deep neural network embedding mechanism is trained end-to-end with GPs as the classification layer, being the first attempt to unify the two approaches \citep{wilson2016deep, bradshaw2017adversarial}. \cite{garriga2018deep} provided a conceptual framework by which classic CNN architectures are translated into the kernel of a shallow GP by exploiting the mathematical properties of the variance of weights matrices. \cite{van2017convolutional} proposed the first convolutional kernel, constructed by aggregating patch response functions. \cite{dutordoir2019translation} have attempted to solve the issue with complete spatial invariance of the convolutional kernel by adding an additional squared exponential kernel between the locations of two patches to account for spatial location, obtaining improvements in accuracy. To extend this shallow GP model to accommodate deeper architectures, \cite{blomqvist2018deep} have proposed to use the convolutional GP on top of a succession of feed-forward GP layers which process data in a convolutional manner akin to standard convolutional layers. However, scaling this framework to modern convolutional architectures with large number of channels in each hidden layer is problematic for two reasons. Firstly, this would imply training high-dimensional multi-output GPs which still represents a research avenue on how to make it more efficient \citep{bruinsma2020scalable}. Ignoring correlations between channels would severely diminish the expressivity of the model. Secondly, the hidden layer GP which process data in a convolutional manner implies taking inducing points, with a dimensionality which scales linearly with the number of channels. This would imply optimization over high-dimensional spaces for each hidden layer, potentially leading to local minima. We will see later on an alternative to this framework for integrating GP in a convolutional architecture, one that is more amenable to modern convolutional architectures.

\subsection{Deep Wasserstein Kernel Learning}

\subsubsection{Generative Process}

We now write the generative process of this new probabilistic framework coined Deep Wasserstein Kernel Learning (DWKL) for 2 layers:
\begin{align}
    p(F_{1}) &= \mathcal{N}\left(PCA(F_{0}), diag\left[K_{ff}\right] \right) \\
    p(F_{2}) &= \mathcal{N}\left[0, k^{W_{2}}\left( p(F_{1}), p(F_{1}) \right) \right]
\end{align}
Due to the introduction of PCA mean functions, data points in the hidden layer are now correlated. To make this clear, we can explicitly calculate it:
\begin{align}
    p\begin{pmatrix}
        F_{2,i} \\ F_{2,j}
    \end{pmatrix}
    &\sim \mathcal{N}\left[
        \begin{pmatrix}
            0 \\
            0
        \end{pmatrix},
        \begin{pmatrix}
            \sigma_{2}^{2} \exp -\frac{-W_{2}^{2}(\mu_{i},\mu_{i})}{l^{2}} & \sigma_{2}^{2} \exp -\frac{-W_{2}^{2}(\mu_{i},\mu_{j})}{l^{2}} \\ 
            \sigma_{2}^{2} \exp -\frac{-W_{2}^{2}(\mu_{j},\mu_{i})}{l^{2}} & \sigma_{2}^{2} \exp -\frac{-W_{2}^{2}(\mu_{j},\mu_{j})}{l^{2}}
        \end{pmatrix}
        \right] \\
    & \sim \mathcal{N}\left[
        \begin{pmatrix}
            0 \\
            0
        \end{pmatrix},
        \begin{pmatrix}
            \sigma_{2}^{2}  & K_{i,j}^{W_{2}} \\ 
            K_{j,i}^{W_{2}} & \sigma_{2}^{2} 
        \end{pmatrix}
        \right]
\end{align}
where $\mu_{i} = \mathcal{N}\left(PCA(F_{0,i}), \sigma^{2} \right)$ and $\mu_{j} = \mathcal{N}\left(PCA(F_{0,j}), \sigma^{2} \right)$. If the PCA embeddings of $x_{i}$ and $x_{j}$ are different, then the Wasserstein-2 distance will be different than zero, hence introducing correlations.

\subsubsection{Evidence lower bound}

Deep Kernel Learning (DKL) \citep{wilson2016deep} is defined as a shallow GP with the input encoded by a neural network:
\begin{equation}
p(Y,F_{L},U_{L}) = \underbrace{p(Y \mid F_{L})}_{\text{likelihood}}\underbrace{ p(F_{L} \mid U_{L};Z_{L-1},\emph{Enc}(X))p(U_{L})}_{\text{prior}}
\end{equation}
,where $\emph{Enc}(X)$ represents the input passed through a neural network encoder, providing a deterministic transformation of the data which is then fed into a SVGP operating on Euclidean data (using equation \eqref{eqn:euclidean_kernel}).

We diverge from this approach by utilising stacked DistGP with Wasserstein-2 kernels as the encoder network, hence our transformed input given by a Gaussian distribution $q(F_{L-1})$. Using the first two moments of the penultimate layer,  we introduce a DistGP so as to obtain the final predictions. The conditional equation for DistGP at arbitrary layer $l \geq 2$ is written as:
\begin{equation}
   p(F_{l} \mid U_{l};Z_{l-1},F_{l-1}) = \mathcal{N}(F_{l} \mid K_{fu}^{W_{2}}\inv{K^{W_{2}}_{uu}}U, K_{ff}^{W_{2}} - Q_{ff}^{W_{2}}) \label{eqn:conditional_distributional_gp} 
\end{equation}
, where we have inducing points $Z_{l} \sim \mathcal{N}(Z_{l} \mid \mu_{Z_{l}},\Sigma_{Z_{l}})$ and uncertain input $F_{l-1} \sim \mathcal{N}\left(F_{l-1} \mid \tilde{U_{l-1}}(F_{l-2}), \tilde{\Sigma_{l-1}}(F_{l-2}) \right)$. For computational reasons we take both $\tilde{\Sigma_{l-1}}(F_{l-2})$ and $\Sigma_{Z_{l}}$ to be diagonal matrices. For $l=1$ we have $q(F_{1}) \sim \mathcal{N}\left(F_{1} \mid \tilde{U_{1}}(x), \tilde{\Sigma_{1}}(x) \right)$ which are the standard predictive equations for SVGP as given in equations \eqref{eqn:posterior_mean_svgp} and \eqref{eqn:posterior_variance_svgp} since the first layer is governed by a standard SVGP operating on Euclidean data.

The joint density prior of Deep Wasserstein Kernel Learning (DWKL) is given as:
\begin{equation}
\underbrace{p(Y|F)}_{\text{likelihood}} \underbrace{ p(F_{L} \mid U_{L};Z_{L-1},Enc(X))\prod_{l=1}^{L}p(U_{l})}_{\text{prior}}
\end{equation}
, where in our case $Enc(X) \sim \mathcal{N} \left[\tilde{U}_{L-1}(F_{L-2}),\tilde{\Sigma}_{L-1}(F_{L-2})\right]$ that acts as the uncertain input for the final distributional GP. We introduce a factorized posterior between layers and dimensions $q(F_{L},\{U_{l}\}_{l=1}^{L}) =  p(F_{L}|U_{L};Z_{L-1})\prod_{l=1}^{L}q(U_{l})$, where $q(U_{l})$ is taken to be a multivariate Gaussian with mean $m_{U_{l}}$ and variance $S_{U_{l}}$. This gives the DWKL variational lower bound:
\begin{equation}
\mathcal{L}_{DKWL} = \textbf{E}_{q(F_{L},\{U_{l}\}_{l=1}^{L})} p(Y \mid F_{L}) - \sum_ {l=1}^{L}KL\left[q(U_{l}) \| p(U_{l})\right]
\end{equation}
, where $q(F_{L}) = \mathcal{N}(\tilde{U}_{L}(Enc(X)), \tilde{\Sigma}_{L}(Enc(X)))$. For $1\leq l \leq L-1$, $F_{l}$ act as features for the next kernel as opposed to random variables that need to be integrated out. We provide pseudo-code of the previously mentioned operations (see Algorithm \ref{alg:dwkl}).

\begin{algorithm*}
   \caption{Deep Wasserstein Kernel Learning}
   \label{alg:dwkl}
\begin{algorithmic}
   \STATE {\bfseries Input:} Euclidean data $X=F_{0}$
   \STATE First layer is standard sparse variational GP
    \STATE {\bfseries Variational Parameters:} $U_{1} \sim \mathcal{N}(m_{U_{1}},~\Sigma_{U_{1}})$
    \STATE {\bfseries Inducing Points:} Euclidean space $Z_{0}$ 
    \STATE $q(F_{1})= \mathcal{N}(F_{1} \mid K_{fu}^{SE}\inv{K^{SE}_{uu}}m_{U_{1}}, K_{ff}^{SE} - K_{fu}^{SE}\inv{K^{SE}_{uu}}(K^{SE}_{uu} - S_{U_{1}})\inv{K^{SE}_{uu}}K_{uf}^{SE}$
   \FOR{$l=2$ {\bfseries to} $L$}
        \STATE Hidden layers are distributional sparse variational GP
        \STATE {\bfseries Variational Parameters:} $U_{l} \sim \mathcal{N}(m_{U_{l}},~S_{U_{l}})$
        \STATE {\bfseries Inducing Points:}  $Z_{l-1} \sim \mathcal{N}(\mu_{Z_{l-1}},~\Sigma_{Z_{l-1}})$  
        \STATE {\bfseries Compute $K_{fu}^{W_{2}}$:} $\sigma^{2}_{l}\exp \sum_{d=1}^{D_{l}}\frac{-W_{2}^{2}(q(F_{l-1}[:,d]),Z_{l-1}[:,d])}{l_{l,d}^{2}}$
        \STATE {\bfseries Compute $K_{uu}^{W_{2}}$:} $\sigma^{2}_{l}\exp \sum_{d=1}^{D_{l}}\frac{-W_{2}^{2}(Z_{l-1}[:,d],Z_{l-1}[:,d])}{l_{l,d}^{2}}$        
        \STATE $q(F_{l})= \mathcal{N}(F_{l} \mid K_{fu}^{W_{2}}\inv{K^{W_{2}}_{uu}}m_{U_{l}}, K_{ff}^{W_{2}} - K_{fu}^{W_{2}}\inv{K^{W_{2}}_{uu}}\left[K^{W_{2}}_{uu} - S_{U_{l}}\right]\inv{K^{W_{2}}_{uu}}K_{uf}^{W_{2}}$
   \ENDFOR
   \STATE {\bfseries Maximize ELBO:} $\mathbb{E}_{q(F_{L}),\{q(U_{l}\}_{l=1}^{L})} p(Y \mid F_{L}) - \sum_{l=1}^{L}KL\left[q(U_{l}) \| p(U_{l})\right]$
\end{algorithmic}
\end{algorithm*}

\subsection{Module Architecture} \label{sec:dist_gp_layers}

\begin{figure}[!htb]
\centering
  \includegraphics[width=0.7\linewidth]{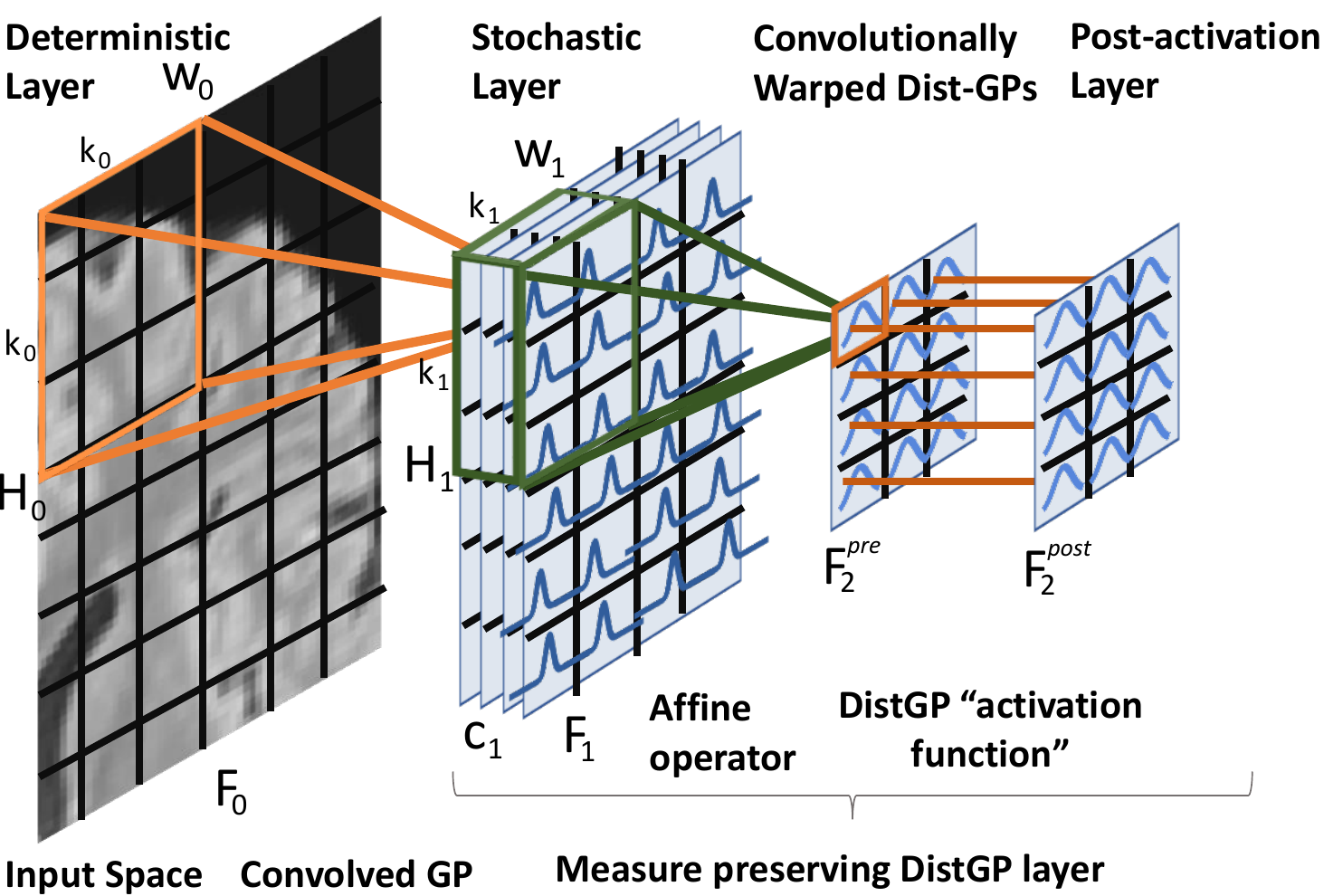}
  \caption{Schematic of measure-preserving DistGP layer. Sparse variational GP is convolved on input data to obtain first hidden layer. Affine operator is convolved on stochastic layer, propagating both mean and variance to obtain the pre-activation of the second hidden layer. Distributional GP is applied element-wise to introduce non-linearities and to propagate distributional uncertainty in the post-activation of the second hidden layer. }
  \label{fig:schematic_algorithm}
\end{figure}

For ease of notation and graphical representation we describe the case of the input being a 2D image, with no loss of generality. We denote the image's representation $F_{l} \in \R^{H_{l},W_{l},C_{l}}$ with width $W_{l}$, height $H_{l}$ and $C_{l}$ channels at the l-th layer of a multi-layer model. $F_{0}$ is the image.
Consider a square kernel of size $k_{l}\!\times\!k_{l}$. We denote with $F^{[p,k_{l}]}_{l} \in \R^{k_{l},k_{l},C_{l}}$ the $p$-th patch of $F_l$, which is the area of $F_l$ that the kernel covers when overlaid at position $p$ during convolution (e.g., orange square for a $3\!\times\!3$ kernel in Figure \ref{fig:schematic_algorithm}).
We introduce the convolved $GP_{0} : F^{[p,k_{0}]}_{0} \rightarrow \mathcal{N}(m,k)$ with $Z_{0} \in \R^{k_{0},k_{0},C_{0}}$ to be the SGP operating on the Euclidean space of patches of the input image in a similar fashion to the layers introduced in \cite{blomqvist2018deep}. For $ 1 \leq l \leq L $ we introduce affine operators
$A_{l} \in \R^{k_{l},k_{l},C_{l-1},C_{l,pre}}$ which are convolved on the previous stochastic layer in the following manner:

\begin{align}
    m(F^{pre}_{l}) &= \text{Conv}_{\text{2D}}(m(F_{l-1}),A_{l}) \label{eq:conv_mu} \\
    var(F^{pre}_{l}) &= \text{Conv}_{\text{2D}}(var(F_{l-1}),A_{l}\odot A_{l}) \label{eq:conv_var}
\end{align}
, where $\odot$ represents the Hadamard product.
The affine operator is sequentially applied on the mean, respectively variance components of the previous layer $ F_{l-1}$ so as to propagate the Gaussian distribution to the next pre-activation layer $ F^{pre}_{l}$. To obtain the post-activation layer, we apply a $DistGP_{l} : F^{pre,[p,1]}_{l} \rightarrow \mathcal{N}(m,k)$ in a many-to-one manner on the pre-activation patches to arrive at $F^{post}_{l}$. Figure \ref{fig:schematic_algorithm} depicts this new module, entitled ``Measure preserving DistGP'' layer with pseudo-code offered in Algorithm \ref{alg:dist_gp_layers}. In \cite{blomqvist2018deep} the convolved GP is used across the entire hierarchy, thereby inducing points are in high-dimensional space ($k_{l}^{2}*C_{l}$). In our case, the convolutional process is replaced by an inducing points free affine operator, with inducing points in low-dimensional space ($C_{l,pre}$) for the DistGP activation functions. The affine operator outputs $C_{l,pre}$, which is taken to be higher than the associated output space of DistGP activation functions $C_{l}$. Hence, the affine operator can cheaply expand the channels, in constrast to the layers in \cite{blomqvist2018deep} which would require high-dimensional multi-output GP. We motivate the preservation of distance in Wasserstein-2 space in the following section. Previous research has highlighted the importance of having an upper bound on $||h(x_{1}) - h(x_{2})||_{h} \leq L_{upper} || x_{1} - x_{2}||_{x}$, as it ensures a certain degree of robustness towards adversarial examples, since it prevents the hidden forward mappings from being overly sensitive to the conceptually meaningless perturbations in input space \citep{jacobsen2018excessive, sokolic2017robust, weng2018evaluating}. Conversely, the lower bound $||h(x_{1}) - h(x_{2})||_{h} \geq L_{lower} || x_{1} - x_{2}||_{x}$ ensures that the forward mappings do not become invariant to semantically meaningful changes in the input \cite{van2020simple}.

\begin{algorithm*}[htb]
   \caption{Distributional Gaussian Processes Layers}
   \label{alg:dist_gp_layers}
\begin{algorithmic}
   \STATE {\bfseries Input:} Euclidean data $X=F_{0} \in \R^{H_{0},W_{0},C_{0}}$
   \STATE First layer is convolved sparse variational $GP_{0} : F^{[p,k_{0}]}_{0} \rightarrow F^{[p]}_{1} $
    \STATE {\bfseries Variational Parameters:} $U_{1} \sim \mathcal{N}(m_{U_{1}},~S_{U_{1}})$
    \STATE {\bfseries Inducing Points:} Euclidean space $Z_{0} \in \R^{k_{0},k_{0},C_{0}}$ 
    \STATE $q(F_{1})= \mathcal{N}(F_{1} \mid K_{fu}^{SE}\inv{K^{SE}_{uu}}m_{U_{1}}, K_{ff}^{SE} - K_{fu}^{SE}\inv{K^{SE}_{uu}}(K^{SE}_{uu} - S_{U_{1}})\inv{K^{SE}_{uu}}K_{uf}^{SE}$
   \FOR{$l=2$ {\bfseries to} $L$}
        %\STATE {\bfseries Convolutionally Warped DistGP}
        \STATE {\bfseries affine operators: } $A_{l} \in \R^{k_{l},k_{l},C_{l-1},C_{l,pre}}$
        \STATE $m(F^{pre}_{l}) = Conv_{2D}(m(F_{l-1}),A_{l})$
        \STATE $v(F^{pre}_{l}) = Conv_{2D}(var(F_{l-1}),A_{l}^{2})$
        %\STATE {\bfseries DistGP \emph{activation function}}
        \STATE Hidden layer activation functions are sparse variational GP $DistGP_{l} : F^{pre,[p,1]}_{l} \rightarrow  F^{post,[p,1]}_{l}$ 
        \STATE {\bfseries Variational Parameters:} $U_{l} \sim \mathcal{N}(m_{U_{l}},~S_{U_{l}})$
        \STATE {\bfseries Inducing Points:}  $Z_{l-1} \sim \mathcal{N}(\mu_{Z_{l-1}},~\Sigma_{Z_{l-1}})$  
        \STATE {\bfseries Compute $K_{fu}^{W_{2}}$:} $\sigma^{2}_{l}\exp \sum_{d=1}^{D_{l}}\frac{-W_{2}^{2}(q(F_{l-1}[:,d]),Z_{l-1}[:,d])}{l_{l,d}^{2}}$
        \STATE {\bfseries Compute $K_{uu}^{W_{2}}$:} $\sigma^{2}_{l}\exp \sum_{d=1}^{D_{l}}\frac{-W_{2}^{2}(Z_{l-1}[:,d],Z_{l-1}[:,d])}{l_{l,d}^{2}}$        
        \STATE $q(F_{l}^{post})= \mathcal{N}(F_{l} \mid K_{fu}^{W_{2}}\inv{K^{W_{2}}_{uu}}m_{U_{l}}, K_{ff}^{W_{2}} - K_{fu}^{W_{2}}\inv{K^{W_{2}}_{uu}}\left[K^{W_{2}}_{uu} - S_{U_{l}}\right]\inv{K^{W_{2}}_{uu}}K_{uf}^{W_{2}}$
   \ENDFOR
   \STATE {\bfseries Maximize ELBO:} $\mathbb{E}_{q(F_{L}),\{q(U_{l}\}_{l=1}^{L})} p(Y \mid F_{L}) - \sum_{l=1}^{L}KL\left[q(U_{l}) \| p(U_{l})\right]$
\end{algorithmic}
\end{algorithm*}

\subsection{Imposing Lipschitz Conditions in Convolutionally Warped DistGP} \label{sec:lipschitz}

If a sample is identified as an outlier at certain layer, respectively being flagged with high variance, in an ideal scenario we would like to preserve that status throughout the remainder of the network. As the kernels operate in Wasserstein-2 space, the distance of a data point's first two moments with respect to inducing points is vital. Hence, we would like our network to vary smoothly between layers, so that similar objects in previous layers get mapped into similar spaces in the Wasserstein-2 domain. In this section, we accomplish this by quantifying the \textit{"Lipschitzness"} of our "Measure preserving DistGP" layer and by imposing constraints on the affine operators so that they preserve distances in Wasserstein-2 space.

\begin{proposition} \label{thm:lipschitz_dist_gp}
    For a given DistGP $F$ and a Gaussian distribution $\mu \sim \mathcal{N}(m_{1},\Sigma_{1}) $ to be the centre of an annulus $B(x) = \{ \nu\sim \mathcal{N}\left(m_{2},\Sigma_{2}\right) \mid 0.125 \leq \frac{W_{2}(\mu,\nu)}{l^{2}} \leq 1.0$ and choosing any $\nu$ inside the ball we have the following Lipschitz bounds: $ W_{2}(F(\mu),F(\nu)) \leq L W_{2}(\mu,\nu)$, where  $L = (\frac{4\sigma^{2}}{l})^{2} \left[ \| K_{uu}^{-1}m \|_{2}^{2} + \| K_{uu}^{-1} \left( K_{uu} - S \right)K_{uu}^{-1} \|_{2} \right]$ and  $l,\sigma^{2}$ are the lengthscales and variance of the kernel.
\end{proposition}

Proof is given in Appendix \ref{apd:lipschitz_proofs}.

\begin{remark}
    This theoretical result shows that DistGP \emph{"activation functions"} have Lipschitz constants with respect to the Wasserstein-2 metric in both output and input domain. This will ensure that the distance between previously identified outliers and inliers will stay constant. However, it is worthy to highlight that we can only obtain locally Lipschitz continuous functions, given that we can only obtain Lipschitz constants for any Gaussian distribution $\nu$ inside the annulus $B(x) = \{ \nu\sim \mathcal{N}\left(m_{2},\Sigma_{2}\right) \mid 0.125 \leq \frac{W_{2}(\mu,\nu)}{l^{2}} \leq 1$. with respect to the centre of the annulus, $\mu$.
\end{remark}

We are now interested in finding Lipschitz constants for the affine operator $A$ that gets convolved to arrive at the pre-activation stochastic layer.

\begin{proposition} \label{thm:lipschitz_affine_layer}
    We consider the affine operator $A \in \R^{C,1}$ operating in the space of multivariate Gaussian distributions of size C. Consider two distributions $\mu \sim \mathcal{N}(m_{1}, \sigma^{2}_{1})$ and $\nu \sim \mathcal{N}(m_{2}, \sigma^{2}_{2})$, which can be thought of as elements of a hidden layer patch, then for the affine operator function $f(\mu) = \mathbb{N}(m_{1} A, \sigma^{2}A^{2})$ we have the following Lipschitz bound: $W_{2}\left(f(\mu), f(\nu)\right) \leq L W_{2}\left(\mu, \nu\right)$, where $L = \sqrt{C}\| W \|_2^{2}$.
\end{proposition}

Proof is given in Appendix \ref{apd:lipschitz_proofs}.

\begin{remark}
    We denote the l-th layer weight matrix, computing the c-th channel by column matrix $A_{l,c}$. We can impose the Lipschitz condition to Eq.~\ref{eq:conv_mu},~\ref{eq:conv_var} by having constrained weight matrices with elements of the form $A_{l,c} = \frac{A_{l,1}}{C^{\frac{1}{2}}\sqrt{\sum_{c=1}^{C}W_{l,c}^{2}}}$. 
\end{remark}

% TODO -- make it crystal clear that the above parametrization has Lipschitz constant=1.0

%\paragraph{\textbf{Definition 1}}

%Let X and Y be metric spaces with metrics $dX$ and $dY$. A map $f : X \xrightarrow{} Y$ is called an isometry or distance preserving if for any $a,b \in X$ one has

%\begin{equation}
%    d Y ( f ( a ) , f ( b ) ) = d X ( a , b )
%\end{equation}

% For a real number $p \geq 1$ the p-norm of a K-dimensional vector $x$ is defined as $\Vert x \Vert_{p} = \left[ \sum_{k=1}^{K} \vert x \vert^{p} \right]^{\frac{1}{p}}$.

\subsection{\emph{Feature-collapse} in DistGP layers} \label{sec:function_space_dist_gp_layers}

In this subsection we delve deeper into the properties of DistGP layers from a function-space view. In light of recent interest into \emph{feature collapse} \cite{van2020simple}, which is the pathological phenomenon of having the representation layer collapse to a small finite set of values, with catastrophic consequences for OOD detection, we investigate what are the necessary conditions for our proposed network to collapse in feature space. Subsequently, we investigate if feature collapse is inherently encouraged by our loss function.

We commence by introducing notation conventions. We consider $\{u_{l} \in \mathbb{R}^{D_{l}} \}_{l=0:L}$ where $D_{l}$ is the number of dimensions in the l-th layer of the hierarchy. We consider the following two functions $\Psi_{l} : u_{l-1} \to \mathcal{R}^{m_{l}}$ and $f_{l}: \mathcal{R}^{m_{l}} \to u_{l}$. To relate this notation to our construction of a DistGP layer introduced in section \ref{sec:dist_gp_layers}, $m_{l}$ represents the number of dimensions of the warped GP (warping performed by affine deterministic layer; see dark green arrows in Figure \ref{fig:schematic_algorithm}). We denote by $f_{l}$ to be the DistGP (mean function included) taking values in the space of continuous functions $C(u_{l};\mathcal{R}^{m_{l}})$, which relates to the "activation function" construction from Figure \ref{fig:schematic_algorithm}. Then we have the following composition for a given DistGP layer:
\begin{equation}
    u_{l}(x) = f_{l}\left(\Psi_{l}(u_{l-1})(x) \right)
\end{equation}
One can easily see that DWKL can be recovered by taking $\Psi_{l} = id$, instead of the affine embedding. 
The first layer prior $p\begin{pmatrix}
    u_{1}(x) \\
    u_{1}(x^{*})
\end{pmatrix}$ is defined as follows:
\begin{equation}
\mathcal{N}
    \left[    \begin{pmatrix} 
        m_{1}(x) \\
        m_{1}(x^{*})
    \end{pmatrix}    
    ,
    \begin{pmatrix} 
    \sigma^{2}_{1}  & k^{E}\left( x, x^{*}\right) \\
    k^{E}\left( x^{*}, x\right)   & \sigma^{2}_{1} 
    \end{pmatrix}
    \right]
\end{equation}
We now define the prior post-activation layers $p\begin{pmatrix}
    u_{l}(x) \\
    u_{l}(x^{*})
\end{pmatrix}$ for $l \geq 2$ in the following recursive manner:
\begin{equation}
\mathcal{N}
    \left[    \begin{pmatrix} 
        m_{l}(x) \\
        m_{l}(x^{*})
    \end{pmatrix}    
    ,
    \begin{pmatrix} 
    \sigma^{2}_{l}  & k^{W_{2}}\left( \mu_{l-1}(x), \mu_{l-1}(x^{*})\right) \\
    k^{W_{2}}\left( \mu_{l-1}(x^{*}), \mu_{l-1}(x)\right)   & \sigma^{2}_{l} 
    \end{pmatrix}
    \right]
\end{equation}

, where $ \mu_{l}\left(x\right) = \mathcal{N}\left( m_{l-1}(x)W_{l}, \sigma^{2}_{l-1} W_{l}^{2}  \right)$, where $m_{l+1}(\cdot) = \overline{\overline{m_{1}(\cdot)W_{1}} \cdots W_{l} }$ and $\overline{m_{1}(x)W_{1}}$ signifies having the Principal Component Analysis (PCA) mean function of the first layer multiplied by $W_{1}$ and averaged across its dimensions.

\begin{proposition} \label{thm:feature_collapse}
    We assume $\mu_{0}$ to be bounded on bounded sets almost-surely. If at each layer we have satisfied the following inequality $D_{l}^{2}  \langle \tilde{W_{l}},\tilde{W_{l}} \rangle \leq 1$, respectively $\left[ D_{L} * \langle \tilde{W_{L}}, \tilde{W_{L}}\rangle + \frac{ \sigma_{L}^{2}}{2l_{L}^{2}} \right] \leq 1$, where $D_{l}$ is the size of the l-th layer and $\tilde{W_{l}}$ represents a normalized version of the affine embedding $W_{l}$, we have the following result:  
    \begin{equation}
        P\left( \|  u_{n}(x) - u_{n}(x^{*}) \|_{2} \to 0 \right) = 1    
    \end{equation}    
\end{proposition}

The proof of Proposition 3 can be found in Appendix \ref{apd:feature_collapse}.

\begin{remark}
    As we have previously outlined in the above derivation, if at each layer we have satisfied the following inequality $D_{l}m_{l-1}D_{l-1}  <\tilde{W_{l}},\tilde{W_{l}}>\leq 1$, respectively $\left[ m_{l}D_{L-1} * \langle \tilde{W_{L}}, \tilde{W_{L}}\rangle + \frac{ \sigma_{L}^{2}}{2l_{L}^{2}} \right] \leq 1$ then the network collapses to constant values. Intuitively, if the norm of $W_{l}$ is not large enough, then it won't change the Gaussian random field too much. Furthermore, if $\sigma_{l}^{2}$ is larger, which translates in increased amplitude of the samples from the Gaussian random field, then the values will not collapse. As opposed to the hypothetical requirements for DGP \cite{dunlop2018deep}, we can immediately notice that for DistGP layers there is no requirement for the kernel variance and lengthscales from intermediate layers, relying solely on the last layer hyperparameters. Lastly, we can notice that as the width of the stochastic layers is increased, alongside warped layers through affine embedding, the conditions are less likely to be satisfied.
\end{remark}

\subsection{Over-correlation in latent space}

\cite{ober2021promises} has highlighted a certain pathology in DKL applied to regression problems in the non-sparse scenario. The authors provide empirical examples of this pathology, whereby features in the representation learning layer are almost perfectly correlated, which would correspond to the feature collapse phenomenon as coined in \cite{van2020simple}. We commence by briefly introducing the main results from that paper and then adapt them to the sparse scenario, which bears more resemblance to what occurs in practice.

Full GPs are trained via type-\RNum{2} maximum likelihood:
\begin{align}
    \log{p(y)} &= \log{\mathcal{N}\left(y \mid 0, K_{ff} +\sigma^{2}_{noise}\mathbb{I}_{n} \right)} \\
    &\propto -\underbrace{\frac{1}{2}\log \mid K_{ff}  + \sigma^{2}_{noise}\mathbb{I}_{n} \mid }_{\textit{complexity penalty}} - \underbrace{\frac{1}{2}y^{\top}\left(K_{ff} + \sigma^{2}_{noise}\mathbb{I}_{n} \right)^{-1}y}_{\textit{data fit}}
\end{align}
, where we define the squared exponential kernel $k^{SE}\left(x_{i}, x_{j}\right) = \sigma^{2}\exp\left[\sum\limits_{d=1}^{D}-\frac{\left(x_{i,d} - x_{j,d}\right)^{2}}{2l_{d}^{2}} \right]$ for $x_{i},x_{j} \in \mathbb{R}^{D}$.

The authors in \cite{ober2021promises} go on to show that at optimal values, the data fit term will converge towards $\frac{N}{2}$, where $N$ is the number of training points. Hence, once the model has reached convergence, it can only increase its log-likelihood score by modifications to the \emph{complexity~penalty} term, which can be broken up as follows:
\begin{equation}
    \frac{1}{2} \log \mid K_{ff} + \sigma^{2}_{noise}\mathbb{I}_{n} \mid = \frac{N}{2}\log \sigma_{f}^{2} + \frac{1}{2}\log \mid \tilde{K_{ff}} +\tilde{\sigma_{noise}^{2}}\mathbb{I}_{n} \mid
\end{equation}
, where we introduced the reparametrizations $K_{ff} = \sigma^{2}\tilde{K_{ff}}$ and $\sigma^{2}_{noise} = \sigma^{2}\tilde{\sigma^{2}_{noise}}$.
We can easily see that if this term is to be minimized, one could decrease $\sigma_{f}$ with the caveat that this would decrease model fit. Hence, the only solution is to have high correlations values in $K_{ff}$ so as to get a determinant close to 0.

In the remainder of this subsection, we derive similar results to \cite{ober2021promises} but in the sparse scenario. We introduce the collapsed bound introduced in \cite{titsias2009variational}:
\begin{align}
    \mathcal{L}_{Titsias} &= \log{\mathcal{N}\left(y \mid Q_{ff} + \sigma^{2}_{noise} \mathbb{I}_{n} \right)} - \frac{1}{2\sigma^{2}_{noise}}Tr\left[K_{ff} -  Q_{ff} \right] \\
    &\propto -\underbrace{\frac{1}{2}\log \mid Q_{ff} + \sigma^{2}_{noise}\mathbb{I}_{n} \mid}_{\textit{complexity penalty}} - \underbrace{ \frac{1}{2}y^{\top}\left(Q_{ff} + \sigma^{2}_{noise}\mathbb{I}_{n} \right)^{-1}y}_{\textit{data fit}} - \underbrace{\frac{1}{2\sigma^{2}_{noise}} Tr \left[K_{ff} -  Q_{ff} \right]}_{\textit{trace term}} \\
    &\propto -\frac{N}{2}\sigma^{2} - \frac{1}{2}\log \mid \tilde{Q_{ff}} + \tilde{\sigma^{2}_{noise}} \mathbb{I}_{n} \mid - \frac{1}{2\sigma^{2}} y^{\top}\left[ \tilde{Q_{ff}} + \tilde{\sigma^{2}_{noise}}\mathbb{I}_{n} \right]y \\ & \nonumber \hspace{1cm}- \frac{1}{2\tilde{\sigma^{2}_{noise}}}Tr\left[ \tilde{K_{ff}} - \tilde{Q_{ff}}\right]
\end{align}
, where we have used again the following notation for kernel terms $k\left(\cdot,\cdot \right) = \sigma^{2} \tilde{k\left(\cdot,\cdot \right)}$ and $\sigma^{2}_{noise} = \sigma^{2}\tilde{\sigma^{2}_{noise}}$. To obtain predictions at testing time under this framework we can make use of the optimal $q(U)$ being given by the following first two moments:
\begin{align}
    m(U^{*}) &= \sigma^{-2}_{noise} K_{uu}\left[ K_{uu} + \sigma^{-2}_{noise}K_{uf}K_{fu}\right]^{-1}K_{uf}y \\
    v(U^{*}) &= K_{uu}\left[ K_{uu} + \sigma^{-2}_{noise}K_{uf}K_{fu}\right]^{-1}K_{uu}
\end{align}
, which we can plug in to standard SVGP predictive equations (equations \eqref{eqn:posterior_mean_svgp} and \eqref{eqn:posterior_variance_svgp}).

We adapt the derivation in \cite{ober2021promises} to our framework at hand:
\begin{align}
    \frac{\partial \mathcal{L}_{Titsias}}{\partial \sigma^{2}} &= \frac{\partial -\frac{N}{2}\sigma^{2} - \frac{1}{2}\log \mid \tilde{Q_{ff}} + \tilde{\sigma^{2}_{noise}} \mathbb{I}_{n} \mid - \frac{1}{2\sigma^{2}} y^{\top}\left[ \tilde{Q_{ff}} + \tilde{\sigma^{2}_{noise}}\mathbb{I}_{n} \right]^{-1}y - \frac{1}{2\tilde{\sigma^{2}_{noise}}}\left[ \tilde{K_{ff}} - \tilde{Q_{ff}}\right]}{\partial \sigma^{2}} \\
    &= -\frac{N}{2\sigma^{2}} + \frac{1}{2\sigma^{4}} y^{\top}\left[ \tilde{Q_{ff}} + \tilde{\sigma^{2}_{noise}}\mathbb{I}_{n}\right]^{-1}y
\end{align}

Hence, if we set the derivative to 0, then we obtain that $\sigma^{2} = \frac{1}{N}y^{\top}\left[ \tilde{Q_{ff}} + \tilde{\sigma^{2}_{noise}}\mathbb{I}\right]^{-1}y$, which if we input it into the data fit term it results in $\frac{N}{2}$, similar to the non-sparse scenario analyzed in \cite{ober2021promises}. The difference between the sparse and non-sparse framework is that after convergence in the data fit term, the model now has to achieve over-correlation in $Q_{ff}$, while still minimizing $K_{ff} - Q_{ff}$.

\subsection{Pooling operations on stochastic layers}

Previous work that dealt with combining GP with convolutional architectures \cite{pmlr-v108-dutordoir20a, kumar2018deep, blomqvist2018deep} have used in their experiments simple architectures involving a couple of stacked layers. In this paper, we propose to experiment with more modern architectures such as DenseNet \cite{huang2017densely} or ResNet \cite{he2016identity}. However, both these architectures include pooling layers such as average pooling, which for Euclidean data is a straightforward operation since we have a naturally induced metric. Since we are using stochastic layers that operate in the space of Gaussian distributions, this introduces some complications as it is not desirable to sample from the stochastic layers, subsequently applying the Euclidean space average pooling operation. Nevertheless, in the remainder of this subsection we show a simple method for replicating average pooling in Wasserstein space by using Wasserstein barycentres \citep{agueh2011barycenters}.

We consider probability measures $\mu_{1}, ...,\mu_{k}$  and  fixed  weights $\theta_{1},...,\theta_{k}$ that are positive real numbers such that $\sum\limits_{k=1}^{K} \theta_{k} = 1$.   For $\nu \in \mathbb{P}_{2}(\mathbf{R}^{d})$, where $\mathbb{P}_{2}$ is the set of Borel probabilities on $\mathbb{R^{d}}$ with finite second moment and absolutely continuous with respect to Lebesque measures, we consider the following functionals: 
\begin{align}
    \mathbf{V}(\nu) &= \sum_{k=1}^{K} \theta_{k}\mathbf{W}_{2}^{2}(\mu,\mu_{k}) \\
    \mathbb{V}(\tilde{\mu}) &= \min_{\mu \in \mathbb{P}_{2}}\mathbb{V}(\mu)
\end{align}
, where $\mathbb{V}(\tilde{\mu})$ is defined as the barycentre with respect to the Wasserstein-2 distance of the set of probabilities $\{\mu_{1}, ...,\mu_{k}\}$. Intuitively, barycentres can be seen as the equivalent of averaging in Euclidean space, while still maintaining the geometric properties of the distributions at hand.

\begin{theorem}[Theorem 4.2. in \cite{alvarez2016fixed}]
    Assume  $\Sigma_{1},...,\Sigma_{K}$ are symmetric positive semidefinite matrices, with at least one of them positive definite. We take $S_{0}\in \mathbb{M}_{d \times d}^{+}$ and define:
    \begin{equation}
    S_{n+1} = S_{n}^{-1/2}(\sum_{k=1}^{K}\theta_{k}(S_{n}^{1/2}\Sigma_{k}S_{n}^{1/2})^{1/2})^{2}S_{n}^{-1/2}    
    \end{equation}
    If $\mathbb{N}(0,\Sigma_{0})$ is the barycenter of $\mathbb{N}(0,\Sigma_{1}),...,\mathbb{N}(0,\Sigma_{K})$ , then $W_{2}^{2}(\mathbb{N}(0,S_{n}),\mathbb{N}(0,\Sigma_{0}) \to 0$ as $n \to \infty$.
\end{theorem}

\begin{remark}
 In the case of computing the barycentre of univariate Gaussian measures, the iterative algorithm converges in one iteration to $\Sigma_{0} = \left( \sum_{k=1}^{K} \theta_{k} \Sigma_{k}^{\frac{1}{2}}\right)^{2}$. This provides us with a deterministic and single step equation to downsample stochastic layers, where we can additionally calculate the mean of the barycentre by $\sum\limits_{k=1}^{K}\theta_{k}m_{k}$, where $\{m_{1}, \cdots, m_{K} \}$ represent the first moments of the respective distributions. 
\end{remark}

\section{DistGP Layer Networks \& OOD detection}

An outlier can be defined in various ways \citep{ruff2021unifying}. In this paper we follow the most basic one, namely \textit{"An anomaly is an observation that deviates considerably from some concept of normality."} More concretely, it can be formalised as follows: our data resides in $X \in \mathbb{R}^{D}$, an anomaly/outlier is a data point $x \in X$ that lies in a low probability region under $\mathcal{P}$ such that the set of anomalies/outliers is defined as $A  =\{x \in X | p(x) \leq \xi \}, ~ \xi \geq 0$, with $\xi$ is a threshold under which we consider data points to deviate sufficiently from what normality constitutes. 

\paragraph{Influence of enforced Lipschitz condition.}

\begin{figure*}[!htb]
    %\vskip 0.2in
    \centering
    \subfigure[DistGP-NN without Lipschitz constraint]{\includegraphics[width=0.45\linewidth]{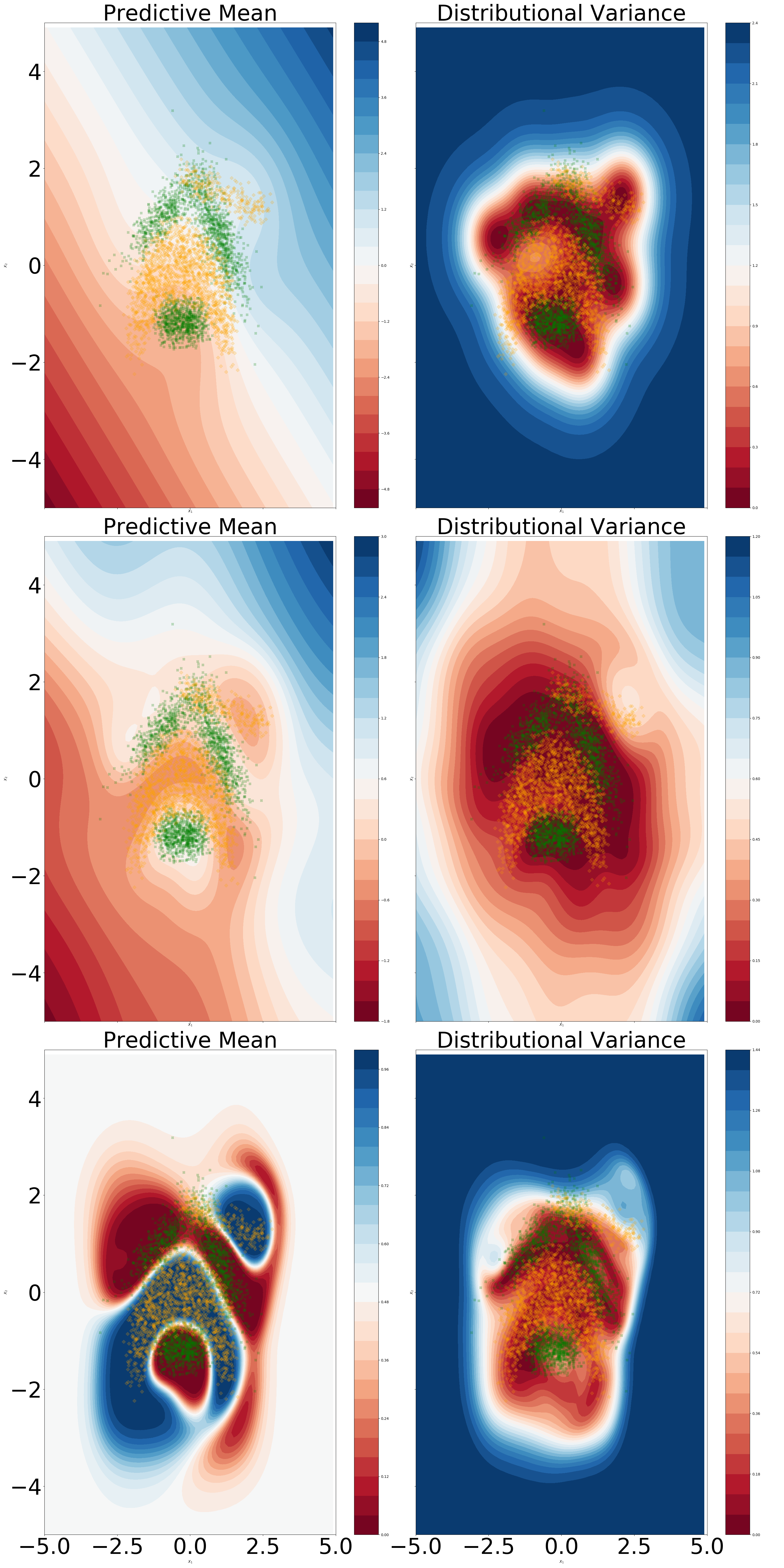}}
    ~
    \subfigure[DistGP-NN with Lipschitz constraint]{\includegraphics[width=0.45\linewidth]{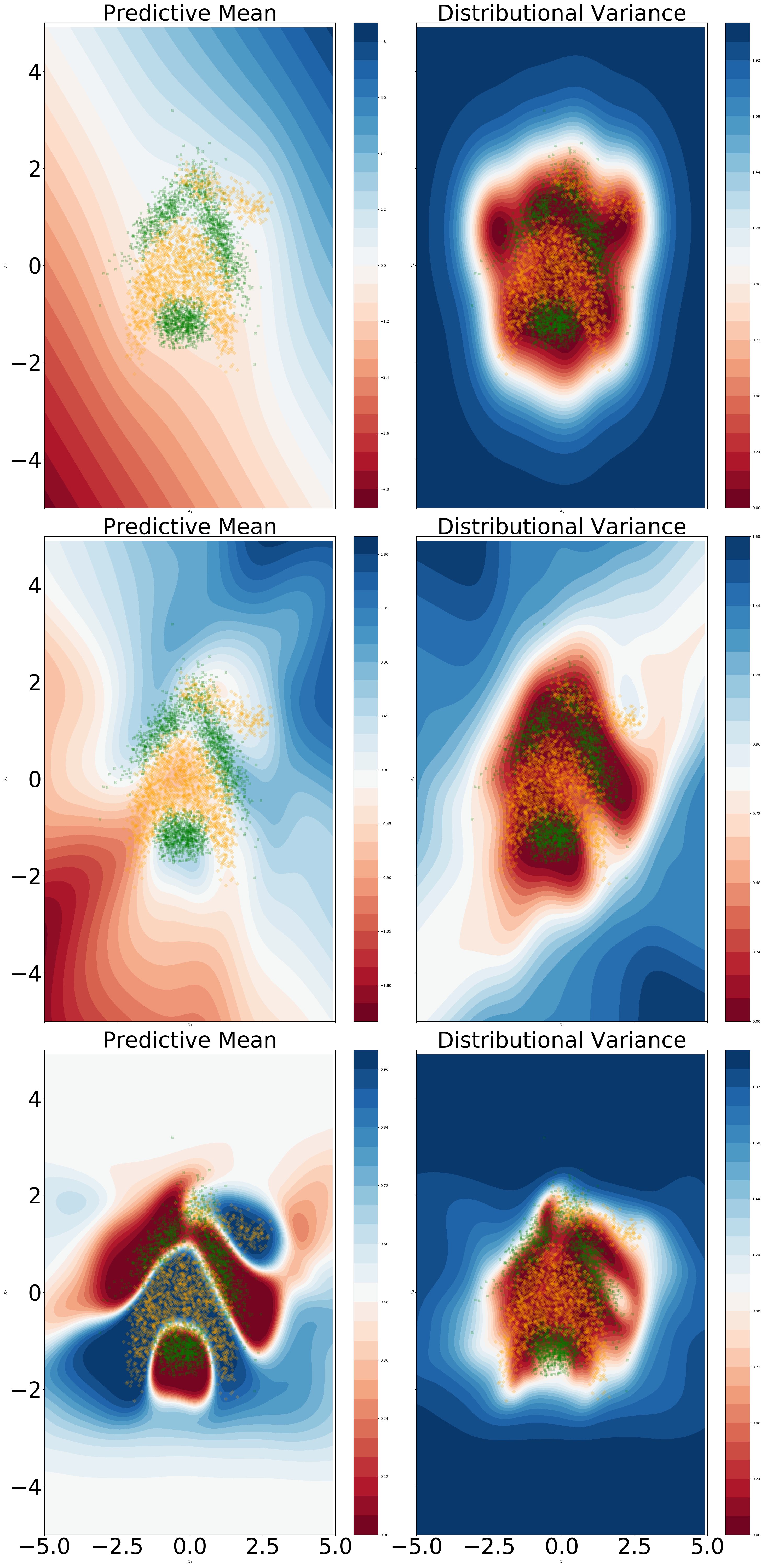}}
    \caption{Layer-wise predictive moments of DistGP-NN models (with or without unitary Lipschitz constraints) trained on toy binary classification dataset.}
    \label{fig:matrix_normalisation_effect_on_banana}  
    %\vskip -0.2in
\end{figure*}

We aim to visually assess if the Lipschitz condition imposed via Proposition \ref{thm:lipschitz_affine_layer} negatively influences the predictive capabilities. We use a standard neural network architecture with two hidden layers with 5 dimensions each, with the affine embeddings operations described in equations \eqref{eq:conv_mu} and \eqref{eq:conv_var} being replaced by a non-convolutional dense layer. From Figure \ref{fig:matrix_normalisation_effect_on_banana} we can notice that imposing a unitary Lipschitz constant does not result in the over-regularization of the predictive mean. A slight smoothing effect on the predictive mean can be noticed in output space. Moreover, for the Lipschitz constrained version we can discern a better fit of the data manifold in terms of distributional variance, with a noticeable difference in the second hidden layer.

\paragraph{Over-correlation in latent space.}

\begin{figure}[!htb]
    \centering
    \subfigure[Collapsed SGPR]{\includegraphics[width=0.45\linewidth]{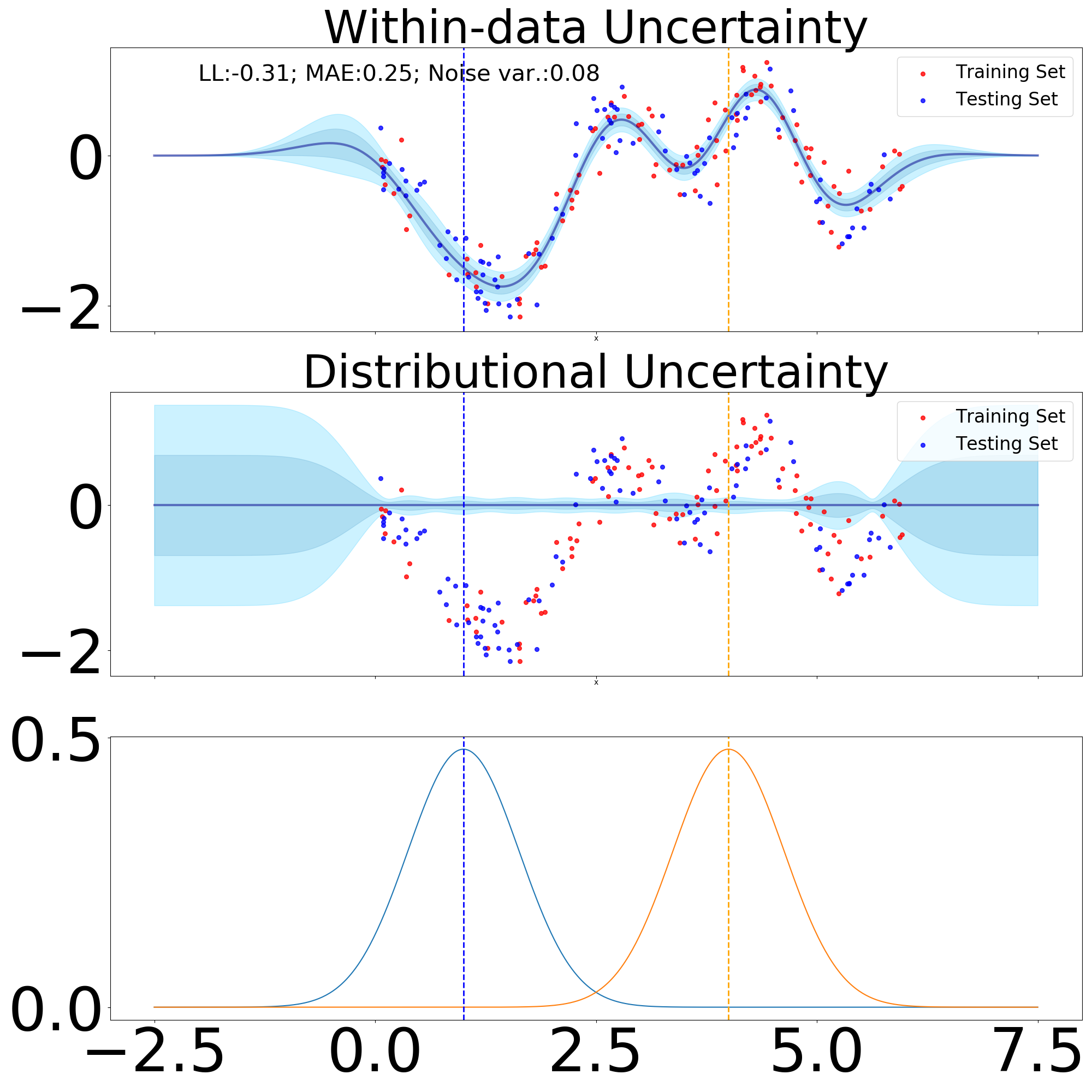}}
    \quad
    \subfigure[DKL]{\includegraphics[width=0.45\linewidth]{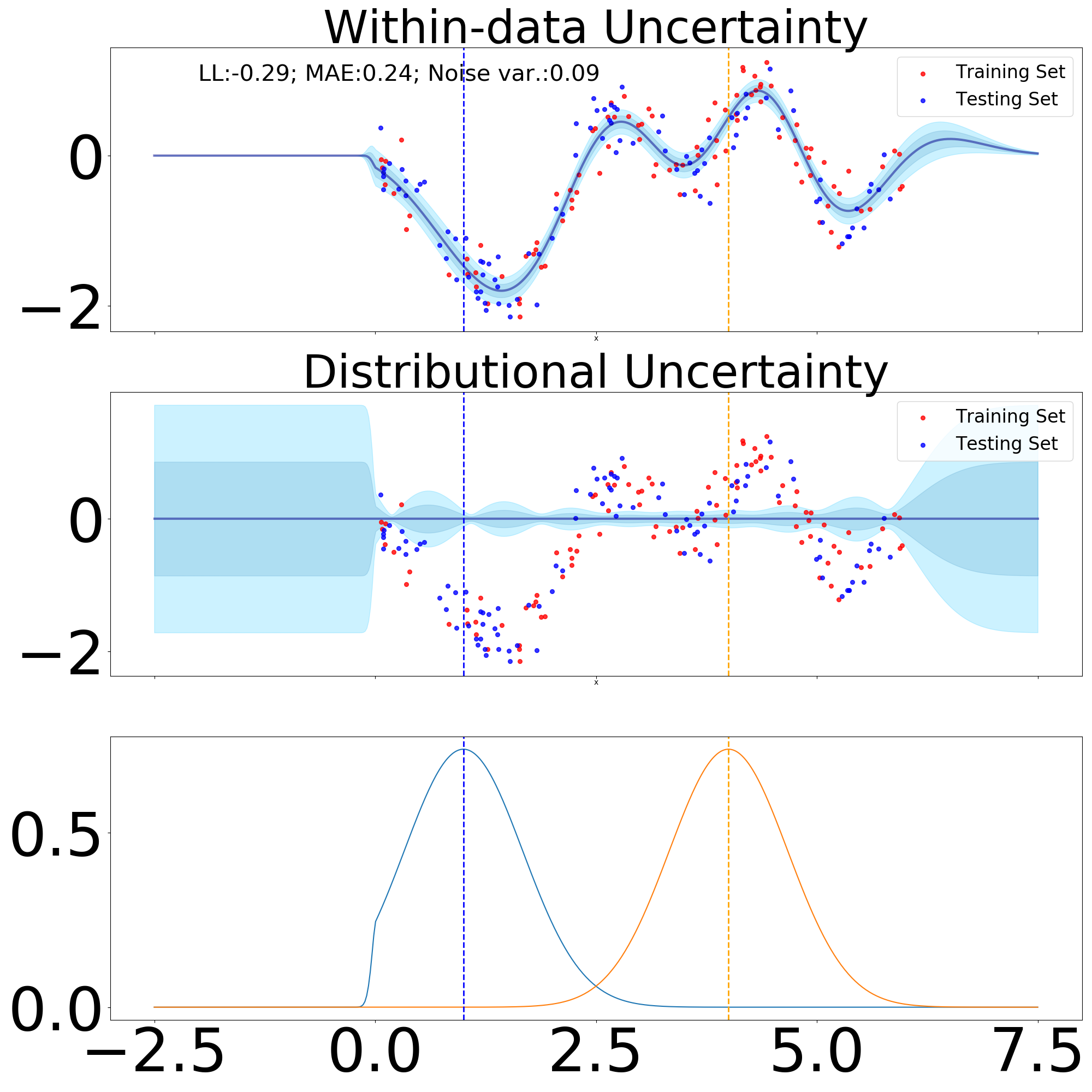}}
    \quad
    \subfigure[DistGP-NN Not Constrained]{\includegraphics[width=0.45\linewidth]{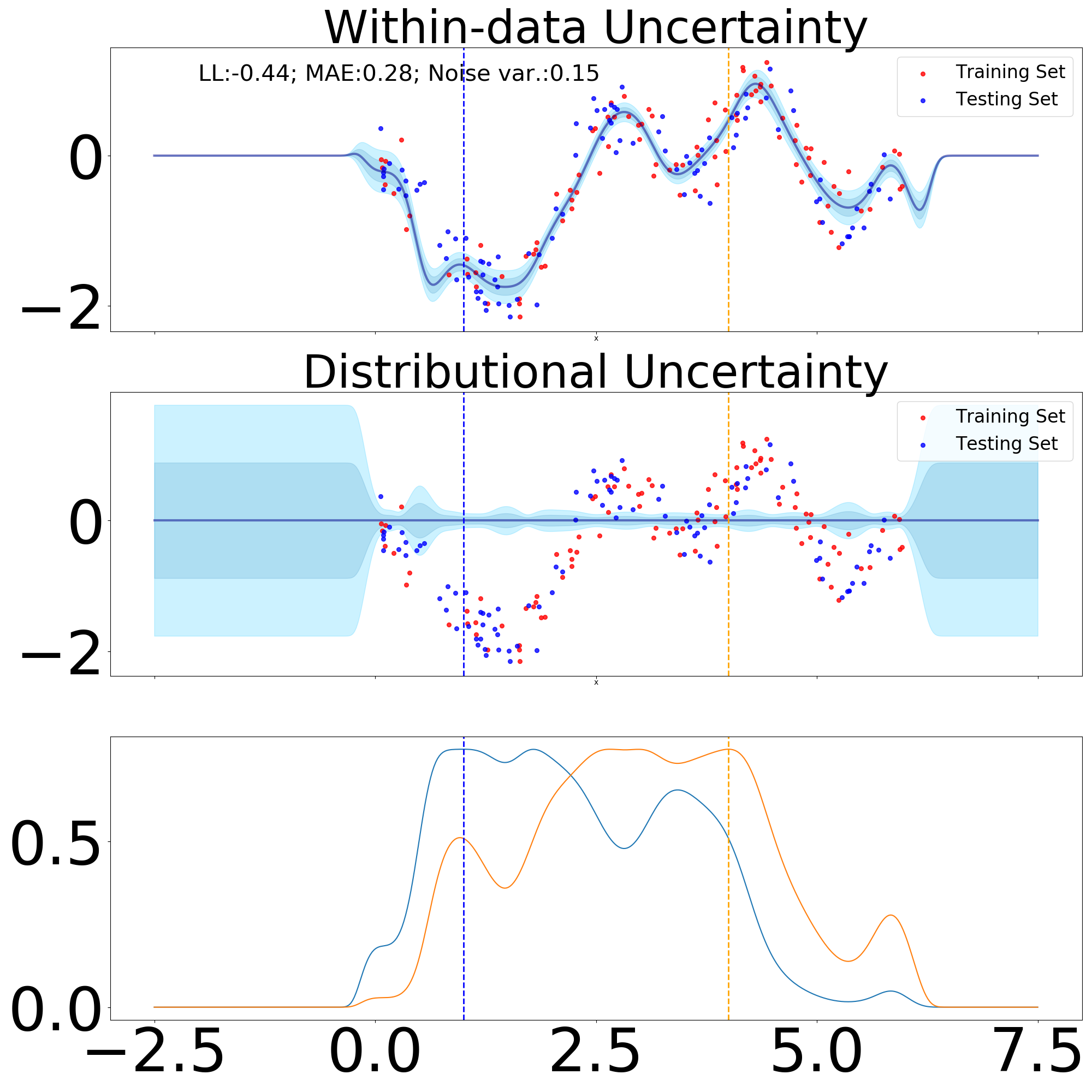}}
    \quad
    \subfigure[DistGP-NN Lipschitz Constraint]{\includegraphics[width=0.45\linewidth]{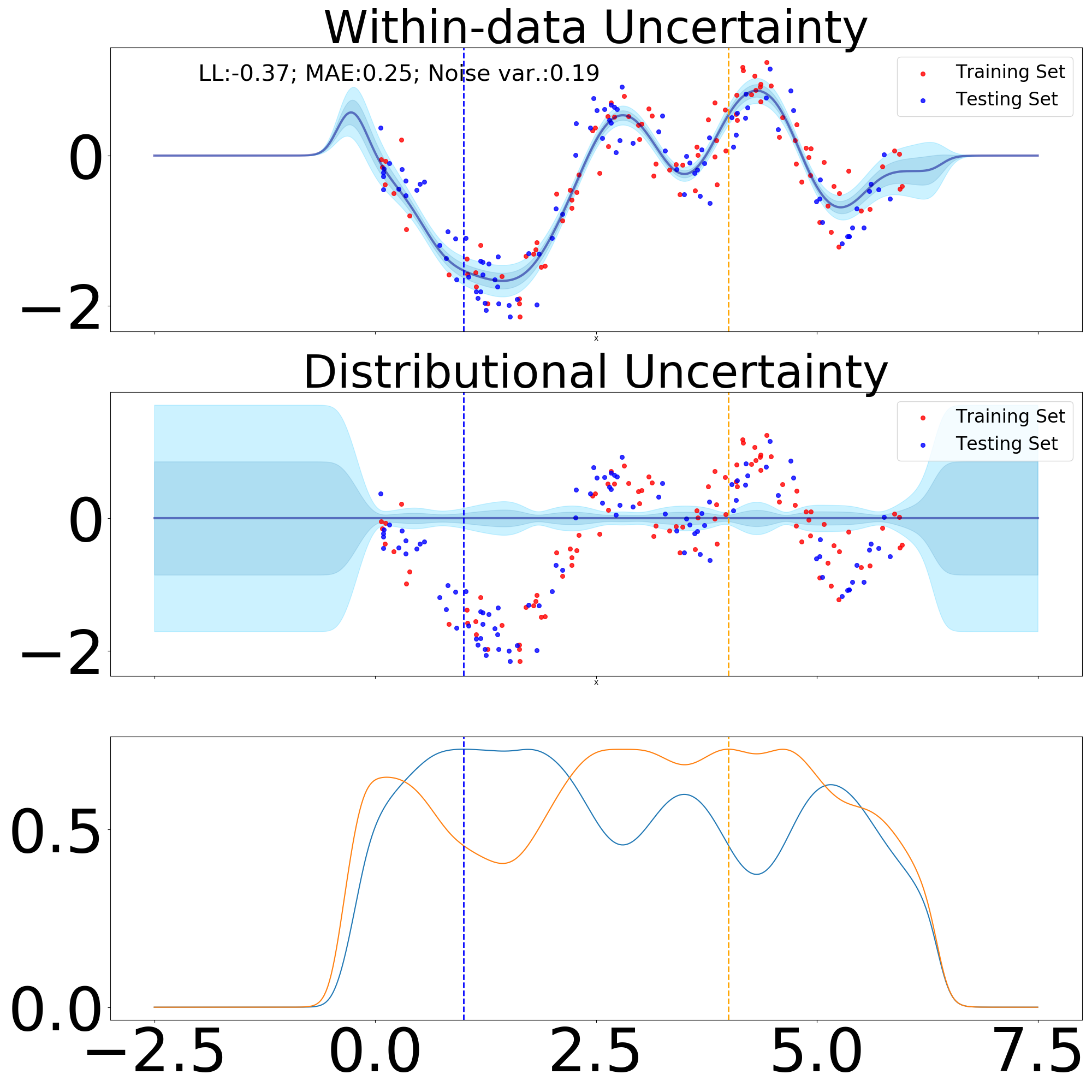}}
  \caption{\textbf{Top row}: Predictive mean and variance of parametric part of SGP; \textbf{Middle row}: Predictive mean and variance of non-parametric part of SGP; \textbf{Bottom row}: Kernel evaluated across whole input span with respect to -2.0 (blue) and 2.0 (orange).}
  \label{fig:ober_experiment_snelson}
\end{figure}

We aim to understand whether the over-correlation phenomenon occurs for our model. We consider a standard neural network architecture with two hidden layers with 50 hidden units per layer. From Figure \ref{fig:ober_experiment_snelson} we can notice that for DKL, the sparse framework does remove any unwanted over-correlations in the final hidden layer latent space. In the unconstrained model, there is a notion of locality in the final hidden layer latent space, albeit of a lower degree compared to the DKL model. With regards to OOD detection, of utmost importance is the fact that regions outside the training set manifold have a correlation value of 0. Perhaps unsurprisingly, introducing a unitary Lipschitz constraint resulted in an increased correlation in the latent space, alongside a smoother predictive mean.

\subsection{Reliability of \emph{in-between} uncertainty estimates}

 We are interested to test our newly introduced module in scenarios where \emph{in-between} uncertainty can fail. For this we use the ``snelson'' dataset, with the training set taken to comprise the intervals between 0.0 and 2.0, respectively 4.0 and 6.5. Thereby, in an ideal scenario we would expect our model to offer high distributional uncertainty estimates between 2.0 and 4.0, which constitutes our \emph{in-between} region. To benchmark our approach, we compare it to a collapsed SGPR as defined in \cite{titsias2009variational}.

\begin{figure}[!htb]
    \centering
    \subfigure[Collapsed SGPR]{\includegraphics[width=0.45\linewidth]{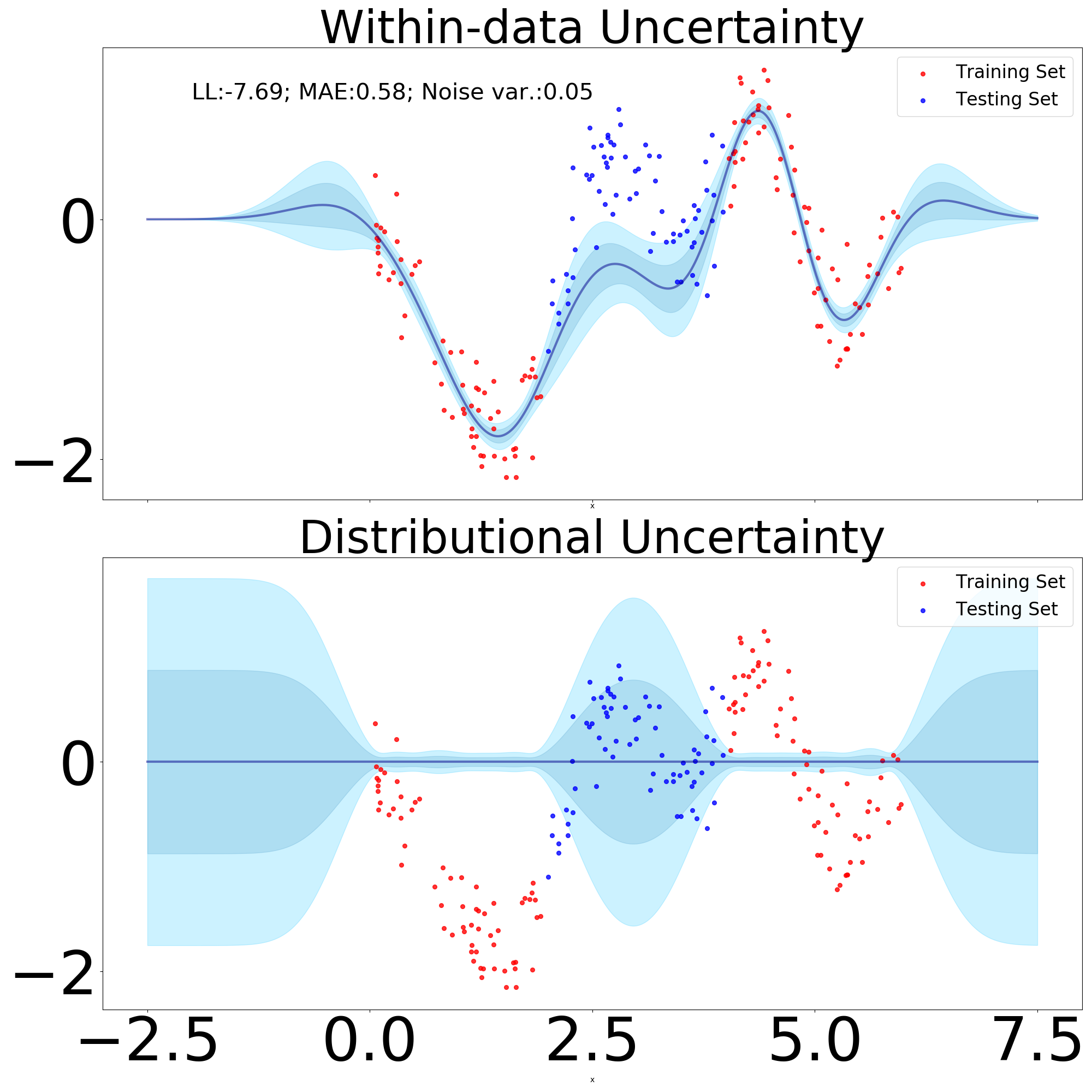}}
    \quad
    \subfigure[DistGP-NN]{\includegraphics[width=0.45\linewidth]{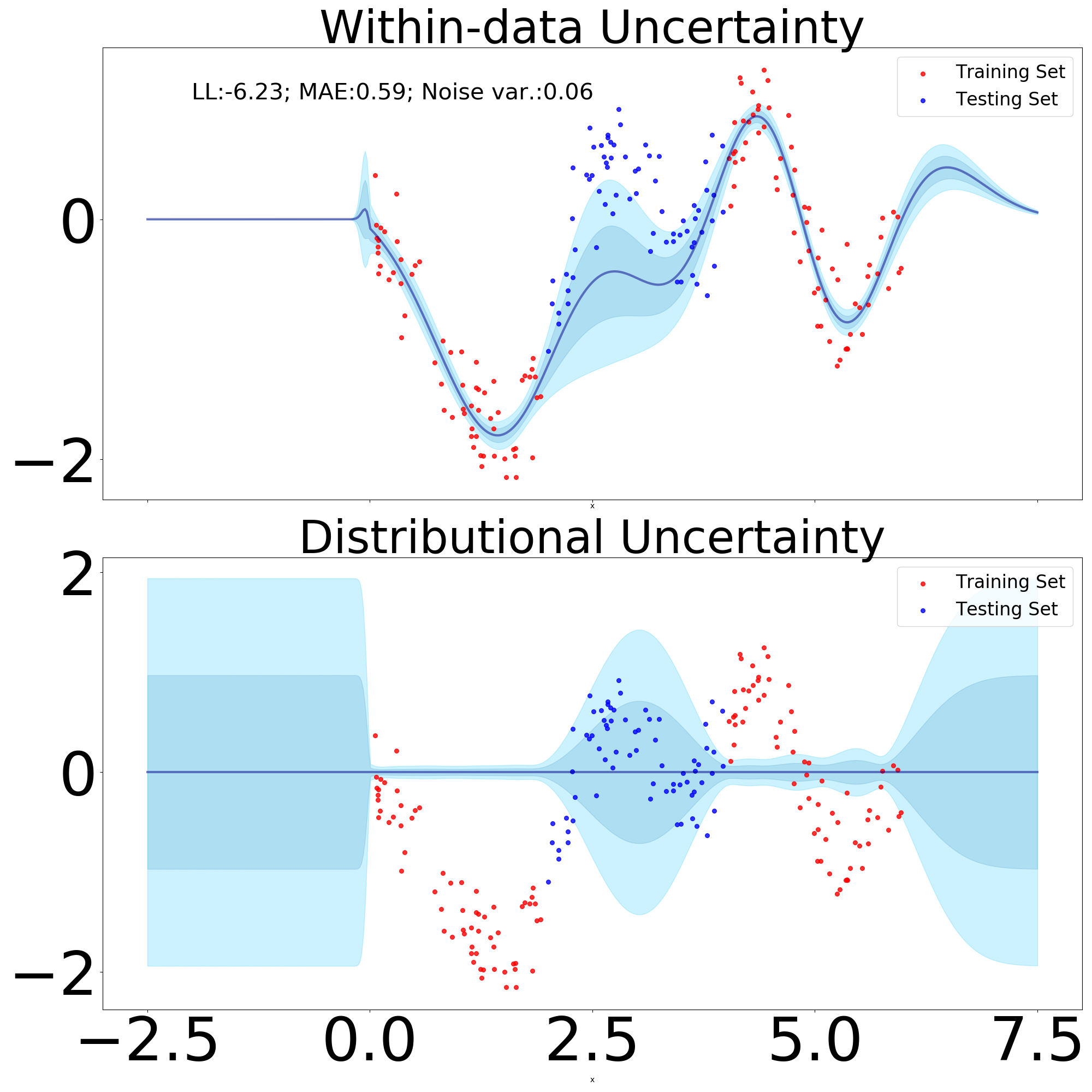}}
  \caption{Reliability of \emph{in-between} uncertainty. Top row: Predictive mean and variance of parametric part of SGP; Bottom row: Predictive mean and variance of non-parametric part of SGP.}
  \label{fig:snelson_gap}
\end{figure}

From Figure \ref{fig:snelson_gap} we can observe that the behaviour is strikingly similar between a collapsed SGPR and a three layer DistGP-Layers network.  

\paragraph{Reliability of \emph{within-data} uncertainty estimates.}

\begin{figure}[!htb]
    \centering
    \subfigure[Collapsed SGPR]{\includegraphics[width=0.45\linewidth]{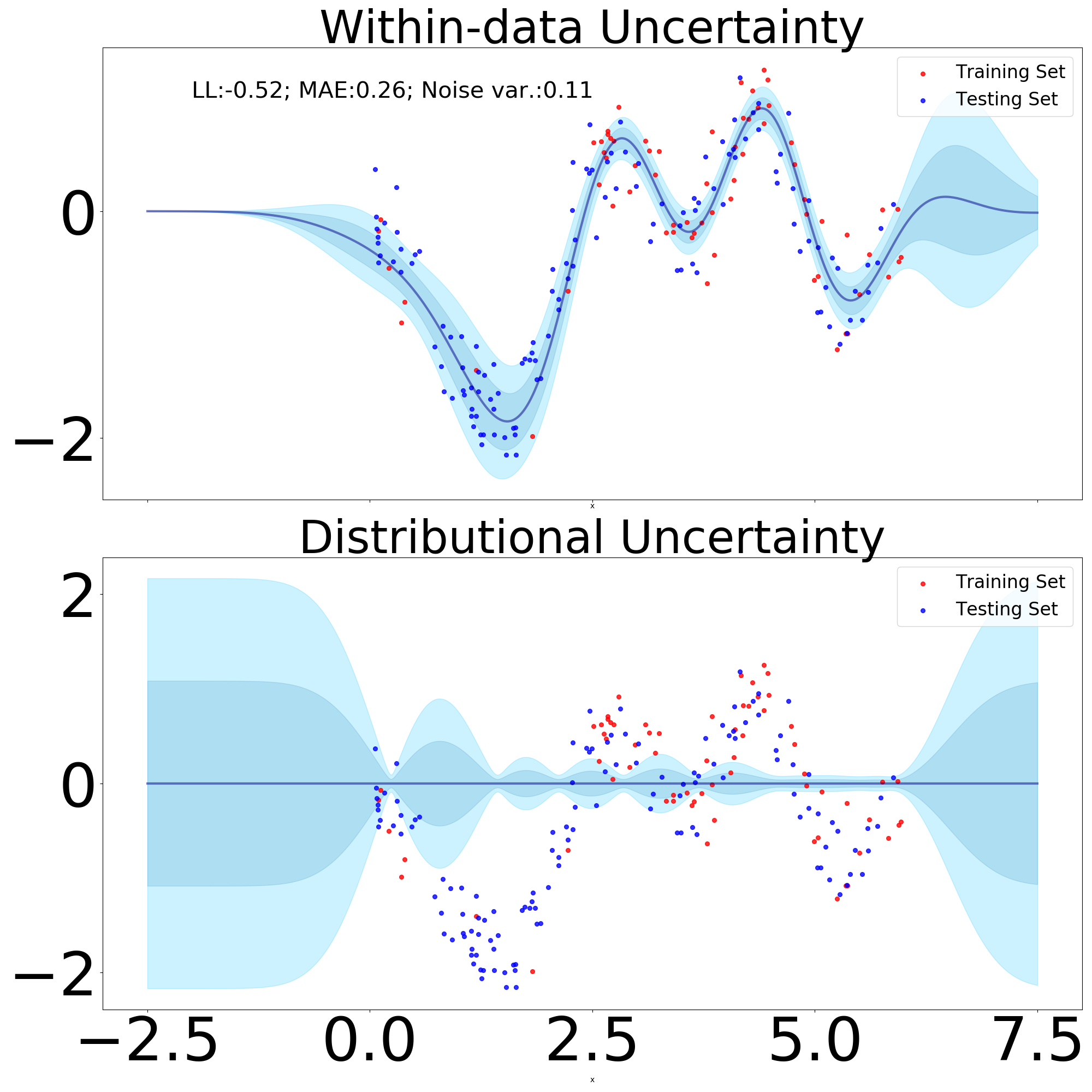}}
    \quad
    \subfigure[DistGP-NN]{\includegraphics[width=0.45\linewidth]{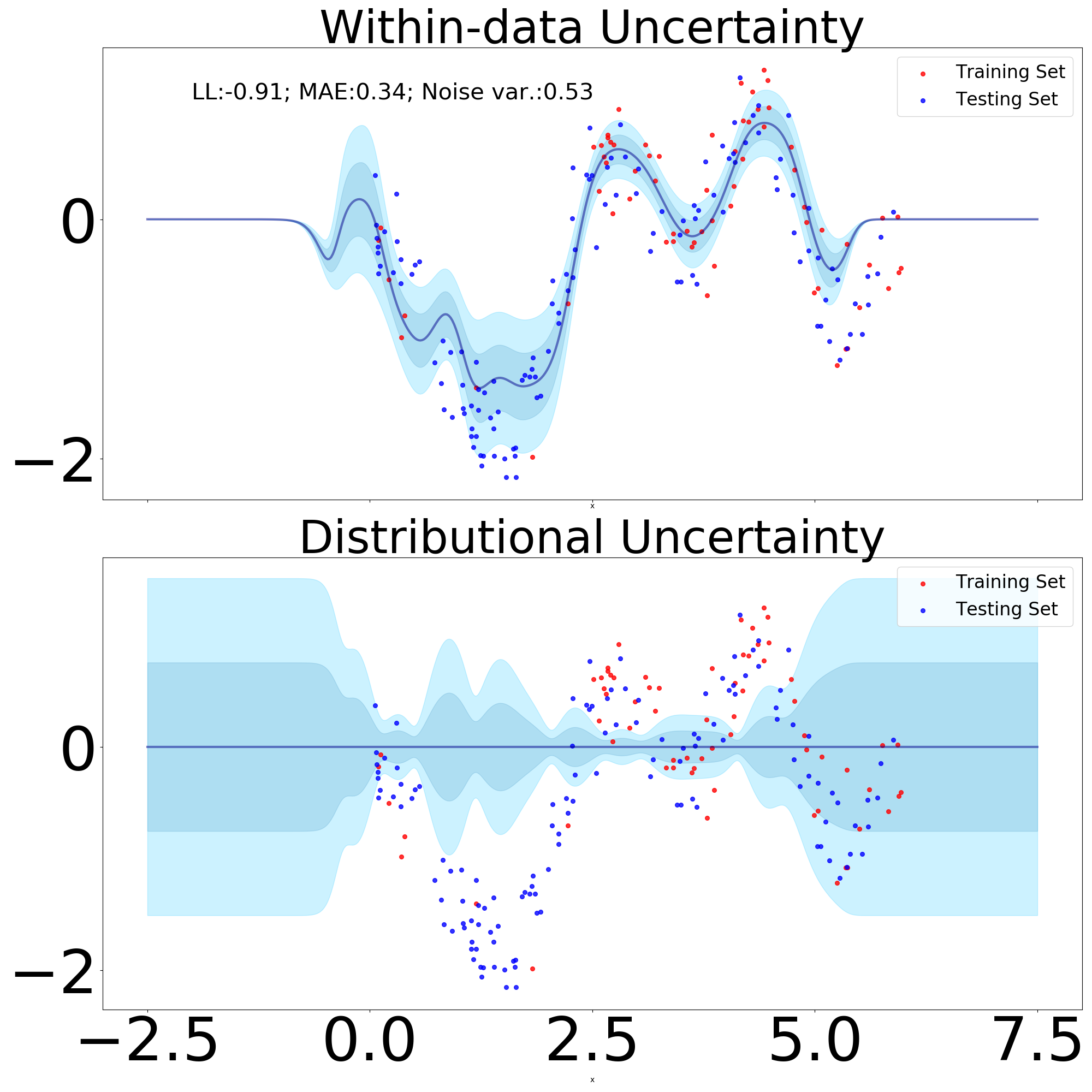}}
  \caption{Reliability of \emph{within-data} uncertainty. Top row: Predictive mean and variance of parametric part of SGP; Bottom row: Predictive mean and variance of non-parametric part of SGP.}
  \label{fig:snelson_subsampled}
\end{figure}

Within-data uncertainty or more conveniently epistemic uncertainty is responsible to detect regions of the input space where the variance in the model parameters, in this case of $U$, can be further reduced if we add more data points in said input regions. To test if our newly introduced module can provide reliable within-data uncertainty estimates one can proceed to subsample a dataset (as done in Figure \ref{fig:snelson_subsampled} subsampling in the interval $[0, 2.5]$), with the intended effect being of an increase in within-data uncertainty across the input region where we subsampled. 

From Figure \ref{fig:snelson_subsampled} we can see that despite the low number of training points, it did not result in over-fitting, with our model exhibiting a relatively smooth predictive mean. Moreover, in comparison to Figure \ref{fig:ober_experiment_snelson} we can notice that the within-data uncertainty has substantially increased in the $[0, 2.5]$ interval.

\paragraph{MNIST and CIFAR10.}

We compare our approach on the standard image classification benchmarks of MNIST \cite{MNIST} and CIFAR-10 \cite{CIFAR10}, which have standard training and test folds to facilitate direct performance comparisons. MNIST contains 60,000 training examples of $28 \times 28$ sized grayscale images of 10 hand-drawn digits, with a separate 10,000 validation set. CIFAR-10 contains 50,000 training examples of RGB colour images of size $32 \times 32$ from 10 classes, with 5,000 images per class. We preprocess the images such that the input is normalized to be between 0 and 1. We compare our model primarily against the original shallow Convolutional Gaussian process \cite{van2017convolutional} and Deep Convolutional Gaussian Process (DeepConvGP) \cite{blomqvist2018deep}. In terms of model architectures, we have used a standard stacked convolutional approach, with the model entitled ``DistGP-DeepConv'' consisting of 64 hidden units for the ``Convolutionally Warped DistGP'' part of the module, respectively, 5 hidden units for the ``DistGP activation-function'' part. For the DeepConvGP, we used 64 hidden units at each hidden layer. All models use a stride of 2 at the first layer. In all experiments we use 250 inducing points at each layer. Lastly, we also devised 18 hidden layers size versions of the ResNet \citep{he2016identity} and DenseNet \citep{huang2017densely} architectures.

\begin{table}[!htb]
\begin{center}
\begin{tabular}{c c c c } 
 \toprule
 Convolutional GP models & Hidden Layers &  MNIST & CIFAR-10 \\ [0.5ex] 
 \midrule
 ConvGP & 0 & $98.83$ & $64.6$  \\ 
DeepConvGP & 1 & $98.38$ & $58.65$ \\ 
DeepConvGP & 2 & $99.24$ & $73.85$  \\ 
DeepConvGP & 3 & $99.44$ & $75.89$ \\
DistGP-DeepConv & 1 & $99.01$ & $70.12$ \\ 
DistGP-DeepConv & 2 & $99.43$ & $76.54$  \\ 
DistGP-DeepConv & 3 & $99.67$ & $78.49$ \\
DistGP-ResNet-18 & 18 & 99.52 & 74.56  \\
DistGP-DenseNet & 18 & 99.75 & 75.29  \\ 
\midrule
 Hybrid NN-GP models & Hidden Layers &  MNIST & CIFAR-10 \\ [0.5ex] 
 \midrule
Deep Kernel Learning & 5 &  $99.2$ & $77.0$ \\
GPDNN & 40 &  $99.95$ & $93.0$ \\ 
\bottomrule
\end{tabular}
\caption{Performance on MNIST and CIFAR-10. Deep Kernel Learning are the set of models from \citet{wilson2016deep}, whereas GPDNN are the set of models published in \cite{bradshaw2017adversarial}. 
Other results than our method are taken from the respective publications}
\label{tab:results_mnist_cifar10}
\end{center}
\end{table}

Table \ref{tab:results_mnist_cifar10} shows the classification accuracy on MNIST and CIFAR-10 for different Convolutional GP models. Compared to other convolutional GP approaches, our method achieves superior classification accuracy compared to DeepConvGP \citep{blomqvist2018deep}. We find that for our method, adding more layers increases the performance significantly. This observation is only available for a couple of stacked layers, as the results from our ResNet and DenseNet variants do not support this assertion. The GPDNN models introduced in \cite{bradshaw2017adversarial} are nonetheless close to state of the art on CIFAR10 but also using a variant of DenseNet \citep{huang2017densely} as the building blocks for their GP classifier.

\paragraph{Outlier detection on different fonts of digits.}

We test if DistGP-DeepConv models outperform OOD detection models from literature such as DUQ \citep{van2020simple}, OVA-DM \citep{padhy2020revisiting} and OVVNI \citep{franchi2020one}. In these experiments we assess the capacity of our model to detect domain shift by training it on MNIST and looking at the uncertainty measures computed on the testing set of MNIST and the entire NotMNIST dataset \citep{bulatov2011notmnist}, respectively SVHN \citep{netzer2011reading}. The hypothesis is that we ought to see both higher predictive entropy and differential entropy for distributional uncertainty  (respectively higher OOD measures specific to each of the baseline models) for the digits stemming from a wide array of fonts present in NotMNST as none of the fonts are handwritten, respectively the digit fonts in SVHN exhibit different backgrounds, orientations besides not being handwritten.

\begin{table}[!htb]
\begin{center}
 \begin{tabular}{r |c  c | c  c  } 
 \toprule
 Model &  \multicolumn{2}{c}{\makecell{ MNIST vs. NotMNIST \\ AUC }}  & \multicolumn{2}{c}{\makecell{ MNIST vs. SVHN \\ AUC }} \\  
 \midrule
 AUC & Pred. Entropy & OOD measure & Pred. Entropy & OOD measure \\
 \midrule
DistGP-DeepConv & \textbf{0.92} &  0.82          &  \textbf{0.95} & 0.98             \\ 
 \midrule
0VA-DM          & 0.73          &  \textbf{1.0}  &  0.70          & \textbf{1.0}     \\ 
 \midrule
OVNNI           & 0.68          &  0.55          &  0.56          & 0.81             \\ 
 \midrule
DUQ             & 0.82          &  0.81          &  0.65          & 0.74             \\ 
\bottomrule
\end{tabular}
\caption[OOD detection results on MNIST using convolutional architectures.]{\textbf{OOD detection results.} Performance of OOD detection based on predictive entropy and distributional differential entropy (for baseline OOD models each has a different OOD measure). Models are trained on MNIST (normative data).}
\label{tab:results_ood_dist_gp_layers}
\end{center}
\end{table}

From Table \ref{tab:results_ood_dist_gp_layers} we can observe that generally all models exhibit a shift in their uncertainty measure between MNIST and notMNIST, with the notable exception of OVNNI which barely manages to better separate the two datasets compared to a random guess. Moreover, OVA-DM manages to completely separate the two datasets with the caveat that it obtains lower predictive entropy for MNIST vs. notMNIST compared to DistGP-DeepConv. The latter achieves similar results to DUQ, with the added benefit of a higher degree of separation using predictive entropy. In the case of SVHN we can observe similar patterns to notMNIST, with OVA-DM and DistGP-DeepConv managing to almost separate the two datasets (MNIST vs. SVHN) by inspecting their uncertainty measure, again with the caveat for OVA-DM that it exhibits lower predictive entropy for SVHN in comparison to MNIST.

\paragraph{Sensitivity to input perturbations.}

MorphoMNIST \citep{castro2018morpho} enables the systematic deformation of MNIST digits using morphological operations. We use MorphoMNIST to better understand the outlier detection capabilities of each method by exposing them to increasingly deformed samples. We use the first 500 MNIST digits in the testing set to generate new images with controlled morphological deformations. We use the swelling deformation with a strength of 3 and increasing radius from 3 to 14. Our hypothesis is that the predictive entropy should increase as the deformation is increased, alongside with the distributional differential entropy, which is a measure of the overall uncertainty in the logit space. This is motivated by the fact that the newly obtained images from MorphoMNIST are outside of the data manifold, which is different from the concept of having high uncertainty as expressed by entropy upon seeing a difficult digit to classify. In this case we would expect high entropy but low differential entropy. 

\begin{figure*}[!htb]
    %\vskip 0.2in
    \centering
    \subfigure[DistGP-DeepConv]{\includegraphics[width=0.45\linewidth]{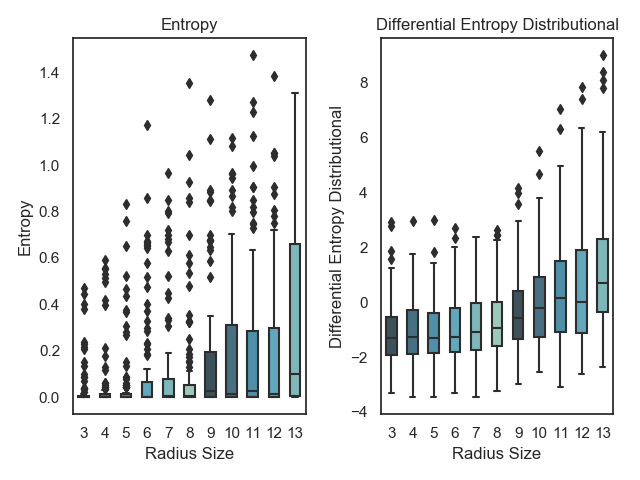}}
    \quad
    \subfigure[OVA-DM]{\includegraphics[width=0.45\linewidth]{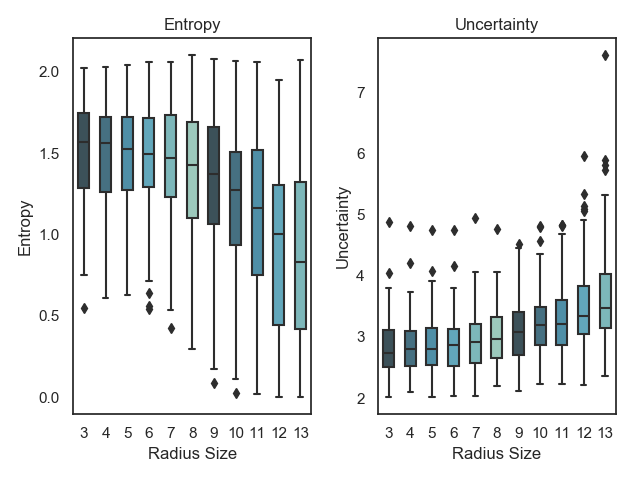}}
    \quad
    \subfigure[OVNNI]{\includegraphics[width=0.45\linewidth]{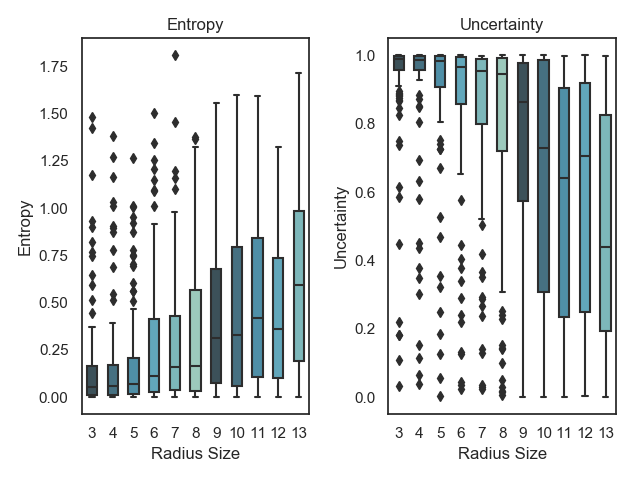}}  
    \quad
    \subfigure[DUQ]{\includegraphics[width=0.45\linewidth]{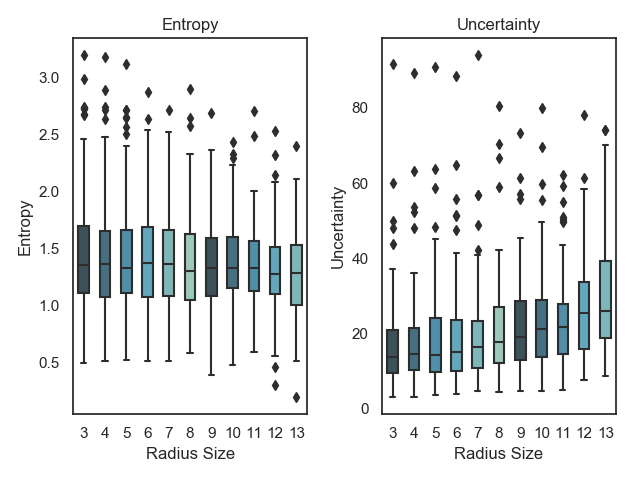}}  

    \caption{Predictive entropy and model-specific uncertainty measure for varying models as swelling of increasing radius is applied on MNIST digits. Higher values of uncertainty measure indicate outlier status, expect for OVNNI where the inverse is true. Results are shown for 3 hidden layers with DistGP-DeepConv dimensionality being set to 5, whereas the capacity of the convolutionally warped DistGPs was set to 12, whereas for OOD models we use 128 hidden units at each layer.}
    \label{fig:morpho_mnist}  
    %\vskip -0.2in
\end{figure*}

All models are able to pick up on the shift in the data manifold as swelling is applied to the original digits, with the model-specific uncertainty measure steadily increasing (for OVNNI, a decrease in the measure translates to higher uncertainty) as increasing deformation is applied. However, for OVA-DM and DUQ the predictive entropy is stable or actually decreases as more deformation is applied, which is in contrast to what one would expect (Figure \ref{fig:morpho_mnist}).

To further assess the sensitivity to input perturbations of our methods, we employ the experiments introduced in \cite{Gal2016DropoutAA} by successively rotating digits from MNIST. We expect to see an increase in both predictive entropy and distributional differential entropy as digits are rotated. For our experiment we rotate digit 6. When the digit is rotated by around 180 degrees the entropy and differential entropy should revert back closer to initial levels, as it will resembles digit 9.

\begin{figure*}[!htb]
    %\vskip 0.2in
    \centering
    \subfigure[DistGP-DeepConv]{\includegraphics[width=0.45\linewidth]{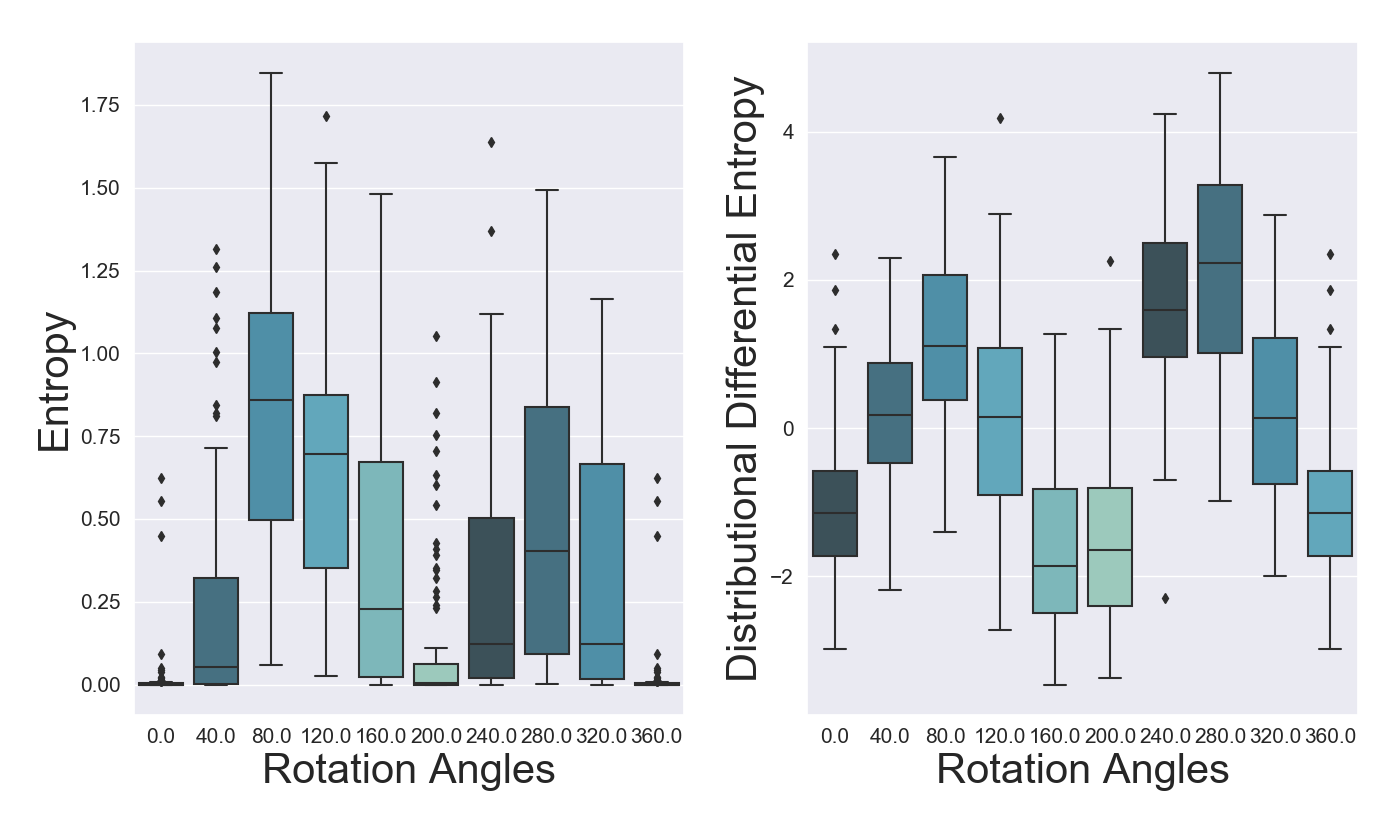}}
    \quad
    \subfigure[OVA-DM]{\includegraphics[width=0.45\linewidth]{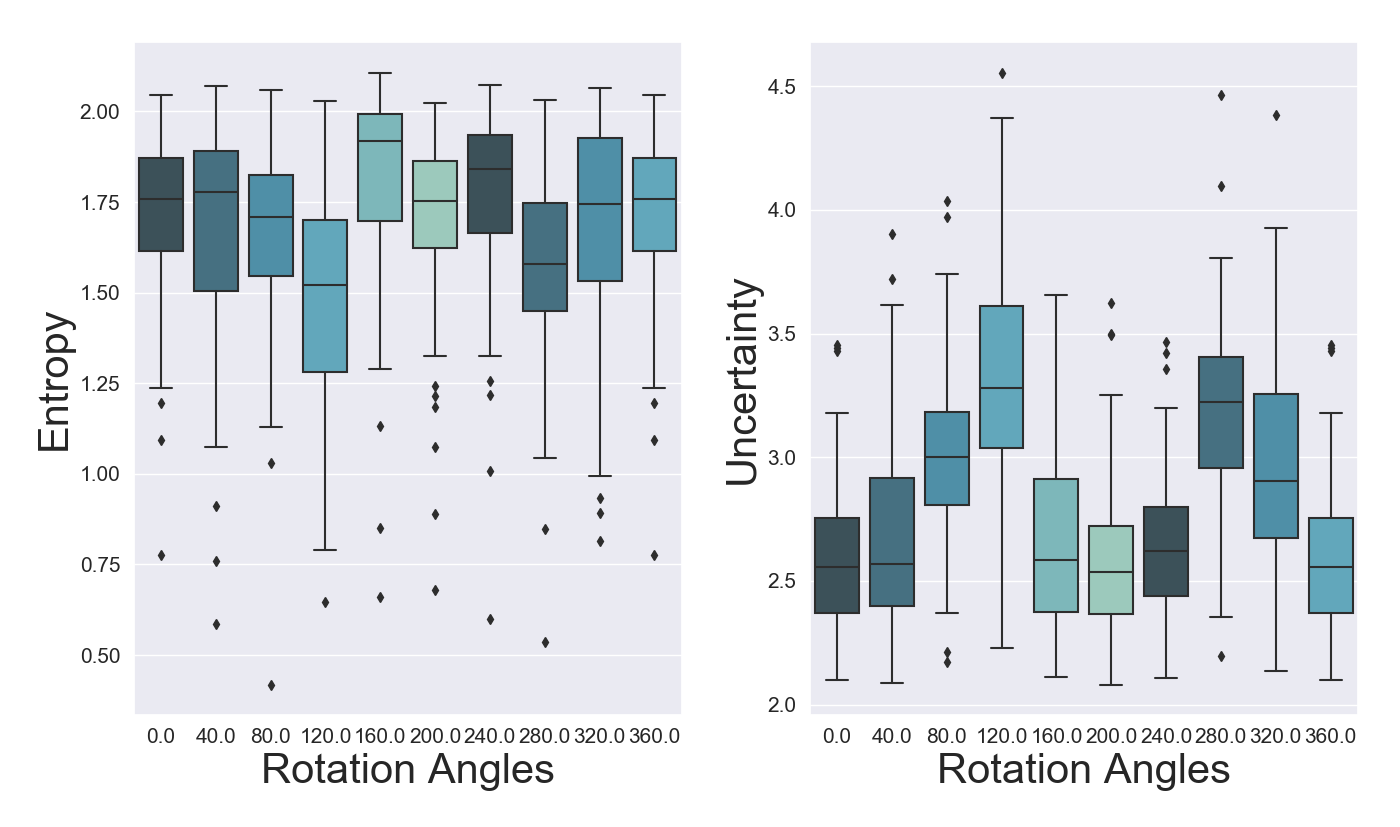}}
    \quad
    \subfigure[OVNNI]{\includegraphics[width=0.45\linewidth]{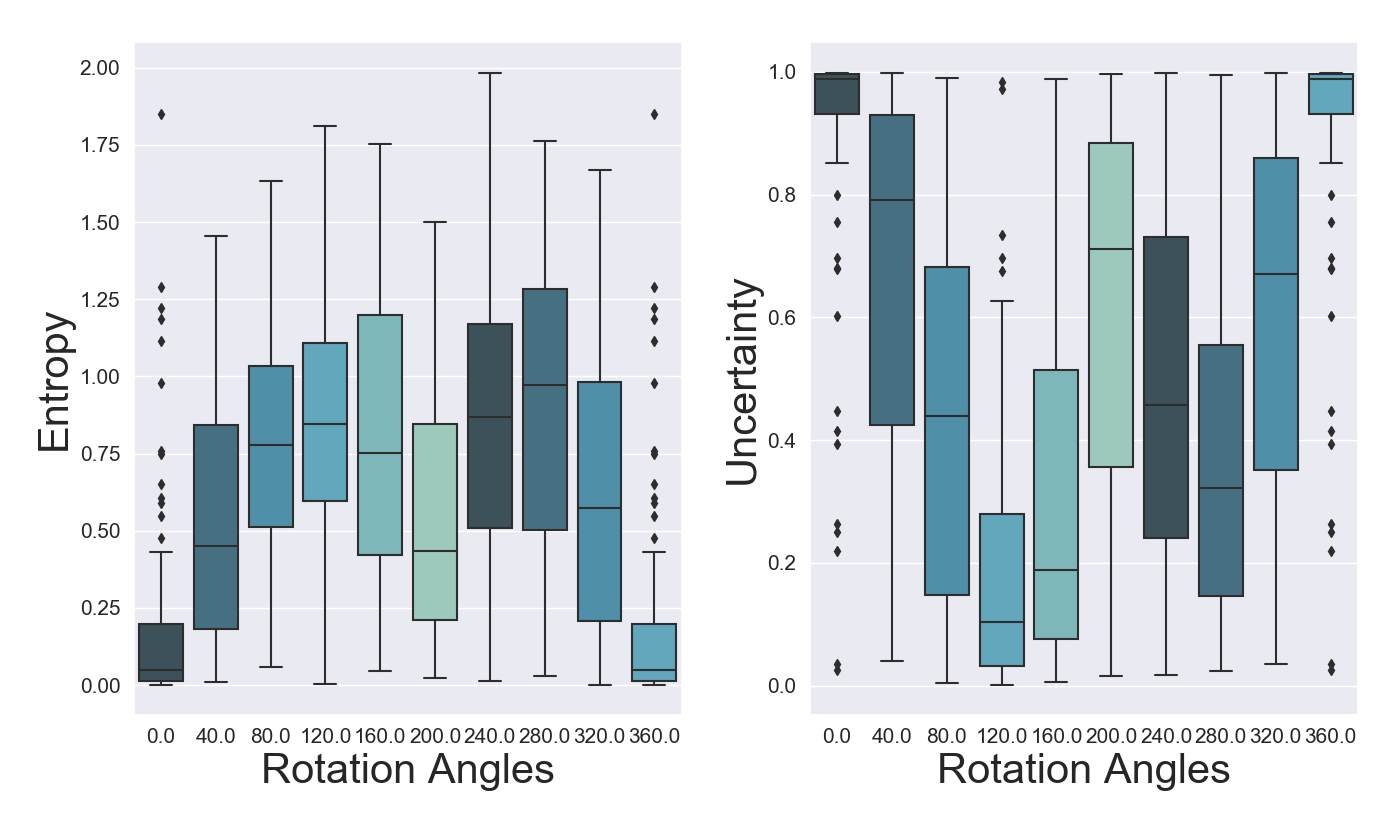}}
    \quad
    \subfigure[DUQ]{\includegraphics[width=0.45\linewidth]{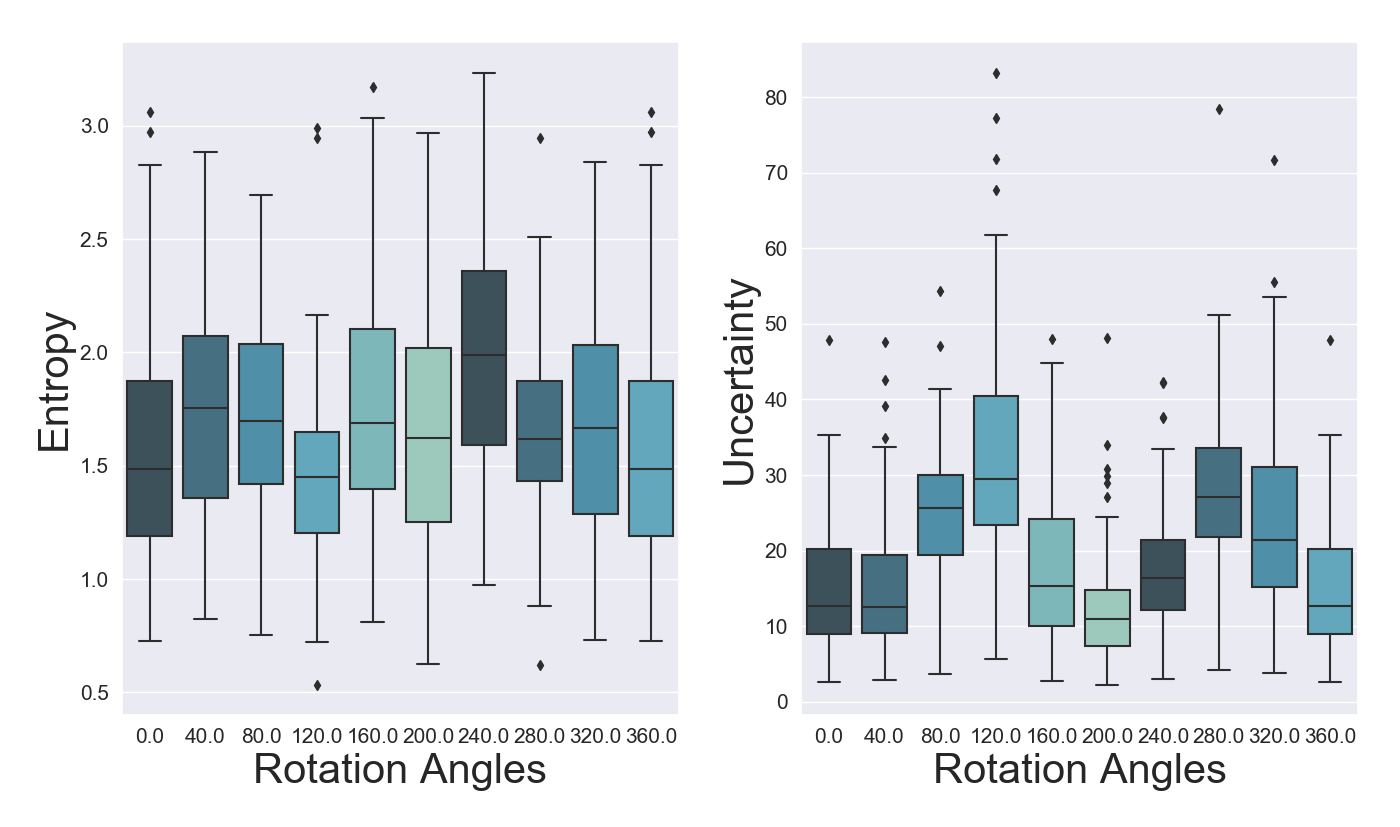}}   
    \caption{Predictive entropy and model-specific uncertainty measure for varying models as varying degrees of rotation is applied to digit 6. Higher values of uncertainty measure indicate outlier status, expect for OVNNI where the inverse is true. Results are shown for 3 hidden layers with DistGP-DeepConv dimensionality being set to 5, whereas the capacity of the convolutionally warped DistGPs was set to 12, whereas for OOD models we use 128 hidden units at each layer.}
    \label{fig:rotated_mnist}  
    %\vskip -0.2in
\end{figure*}

From Figure \ref{fig:rotated_mnist} we can notice that all models exhibit an increase (decrease for OVNNI translates into higher uncertainty) in their specific uncertainty measures for rotation angles between 40 and 160, respectively between 240 and 320 degrees. In terms of predictive entropy, we can discern relatively stable and highly overlapping values for OVA-DM and DUQ, whereas for DistGP-DeepConv and OVNNI we can observe a clear pattern of increases and decreases as what was originally a 6 becomes a 9.

\section{DistGP-based Segmentation Network \& OOD Detection in Medical Imaging} \label{sec:medical_imaging}

\begin{figure}[!htb]
  \includegraphics[width=\linewidth]{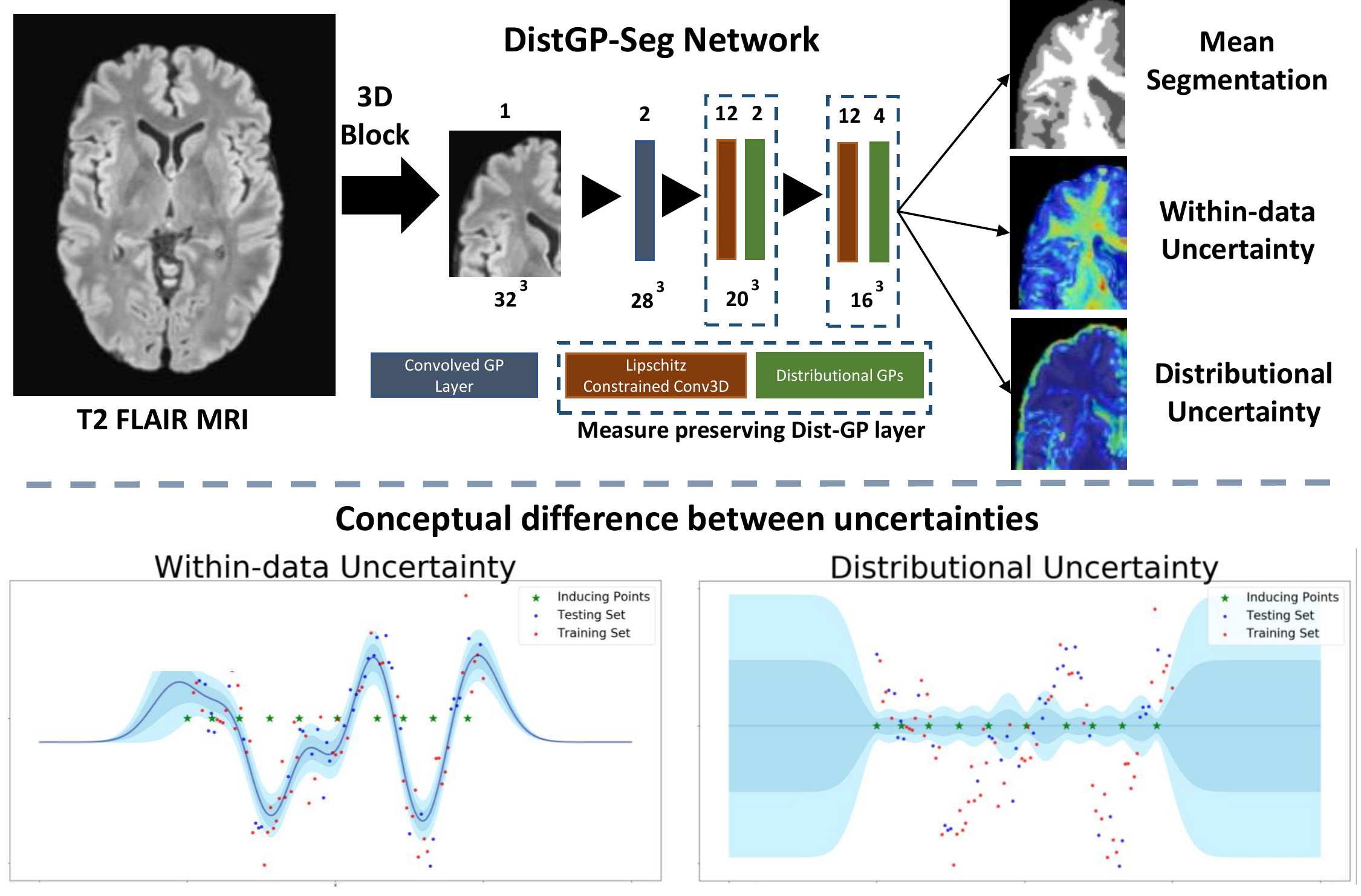}
  \caption{\textbf{Top}: Schematic of proposed DistGP activated segmentation net. Above and below each layer we show the number of channels and their dimension respectively. \textbf{Bottom}: Visual depiction of the two uncertainties in DistGP after fitting a toy regression dataset. Hyperparameters and variational approximate posteriors are optimized. Distributional uncertainty increases outside the manifold of training data and is therefore useful for OOD detection.}
  \label{fig:schematic_seg_net_and_uncertainties}
\end{figure}

The above introduced modules in Sec.~\ref{sec:dist_gp_layers} can be used to construct a convolutional network that benefits from properties of DistGP. Specifically, we construct a 3D network for segmenting volumetric medical images, which is depicted in Figure \ref{fig:schematic_seg_net_and_uncertainties} (top). It consists of a convolved GP layer, followed by two measure-preserving DistGP layers. Each hidden layer uses filters of size $5\!\times\!5\!\times\!5$. To increase the model's receptive field, in the second layer we use convolution dilated by 2. We use 250 inducing points and 2 channels for the DistGP ``activation functions''. The affine operators project the stochastic patches into a 12 dimensional space. The size of the network is limited by computational requirements for GP-based layers, which is an active research area. Like regular convolutional nets, this model can process input of arbitrary size but GPU memory requirement increases with input size. We here provide input of size $32^{3}$  to the model, which then segments the central $16^{3}$ voxels. To segment a whole scan we divide it into tiles and stitch together the segmentations.

\subsection{Evaluation on Brain MRI}

In this section we evaluate our method alongside recent OOD models \citep{van2020simple,franchi2020one,padhy2020revisiting}, assessing their capabilities to reach segmentation performance comparable to well-established deterministic models and whether they can accurately detect outliers.

\subsubsection{Data and pre-processing}

For evaluation we use publicly available datasets:

1) Brain MRI scans from the UKBB study \citep{alfaro2018image}, which contains scans from nearly 15,000 subjects. We selected for training and evaluation the bottom 10$\%$ percentile in terms of white matter hypointensities with an equal split between training and testing. All subjects have been confirmed to be normal by radiological assessment. Segmentation of brain tissue (CSF,GM,WM) has been obtained with SPM12.

2) MRI scans of 285 patients with gliomas from BraTS 2017 \citep{bakas2017advancing}. All classes are fused into a \emph{tumor} class, which we will use to quantify OOD detection performance.

In what follows, we use only the FLAIR sequence to perform the brain tissue segmentation task and OOD detection of tumors, as this MRI sequence is available for both UKBB and BraTS. All FLAIR images are pre-processed with skull-stripping, N4 bias correction, rigid registration to MNI152 space and histogram matching between UKBB and BraTS. Finally, we normalize intensities of each scan via linear scaling of its minimum and maximum intensities to the [-1,1] range.

\subsubsection{Brain tissue segmentation on normal MRI scans}

\begin{table}[!htb]
\begin{center}

 \begin{tabular}{c c c c c} 
 \toprule
 Model & Hidden Layers &  DICE CSF & DICE GM & DICE WM \\ [0.5ex] 
 \midrule
 % MC-Dropout     & 3         &   ?       & ?       & ??   & ? \\
 OVA-DM \citep{padhy2020revisiting}    & 3         &   0.72       & 0.79       & 0.77  \\
 OVNNI \citep{franchi2020one}    & 3         &   0.66       & 0.77       & 0.73   \\
 DUQ \citep{van2020simple}       & 3         & 0.745      & 0.825    & 0.781 \\ 
 DistGP-Seg (ours) & 3 & 0.829 & 0.823 & 0.867 \\
\midrule
 U-Net     & 3 scales  & 0.85 & 0.89 & 0.86 \\ 
\bottomrule
\end{tabular}
\caption{Performance on UK Biobank in terms of Dice scores per tissue.}
\label{tab:results}
\end{center}
\end{table}

\paragraph{\textbf{Task:}} We train and test our model on the task of segmenting brain tissue of healthy UKBB subjects. This corresponds to the within-data manifold in our setup.

\paragraph{\textbf{Baselines:}} 
We compare our model with recent Bayesian approaches for enabling task-specific models (such as image segmentation) to perform uncertainty-based OOD detection \citep{van2020simple,franchi2020one,padhy2020revisiting}. For fair comparison, we use these methods in an architecture similar to ours (Figure ~\ref{fig:schematic_seg_net_and_uncertainties}), except that each layer is replaced by standard convolutional layer, each with 256 channels, LeakyRelu activations, and dilation rates as in ours. We also compare these Bayesian methods with a well-established deterministic baseline, a U-Net with 3 scales (down/up-sampling) and 2 convolution layers per scale in encoder and 2 in decoder (total 12 layers).

\paragraph{\textbf{Results:}}
Table \ref{tab:results} shows that DistGP-Seg surpasses other Bayesian methods with respect to Dice score for all tissue classes. Our method approaches the performance of the deterministic U-Net, which has a much larger architecture and receptive field. We emphasize this has not been previously achieved with GP-based architectures, as their size (e.g., number of layers) is limited due to computational requirements. This supports the potential of DistGP, which is bound to be further unlocked by advances in scaling GP-based models.

\subsubsection{Outlier detection in MRI scans with tumors} \label{sec:outlier_tumors_brats}

\begin{table}[!htb]
\begin{center}
 \begin{tabular}{c c c c c} 
 \toprule
Model  & \multicolumn{1}{p{2cm}}{\centering DICE \\ FPR=0.1 }  & \multicolumn{1}{p{2cm}}{\centering DICE \\ FPR=0.5 } & \multicolumn{1}{p{2cm}}{\centering DICE \\ FPR=1.0 } & \multicolumn{1}{p{2cm}}{\centering DICE \\ FPR=5.0 } \\ [0.5ex] 
 \midrule
 % MC-Dropout     & 3         &   ?       & ?       & ??   & ? \\
 OVA-DM \citep{padhy2020revisiting}       &    0.382      &    0.428    &  0.457  & 0.410 \\
 OVNNI \citep{franchi2020one}         &    $\leq 0.001 $      & $\leq 0.001 $ & $\leq 0.001 $   & $\leq 0.001 $ \\
 DUQ \citep{van2020simple}    & 0.068     &  0.121    &   0.169    & 0.182 \\
 DistGP-Seg (ours) & 0.512 & 0.571 & 0.532 & 0.489 \\
\midrule
 VAE-LG \citep{chen2019unsupervised}          &  0.259      &  0.407      &  0.448  & 0.303 \\
 AAE-LG \citep{chen2019unsupervised}            &   0.220     &   0.395     &  0.418  & 0.302 \\
\bottomrule
\end{tabular}
\caption{Performance comparison of Dice for detecting outliers on BraTS for different thresholds obtained from UKBB.}
\label{tab:ood_results}
\end{center}
\end{table}

\paragraph{\textbf{Task:}} The previous task of brain tissue segmentation on UKBB serves as a proxy task for learning normative patterns with our network. Here, we apply this pre-trained network on BRATS scans with tumors. We expect the region surrounding the tumor and other related pathologies, such as squeezed brain parts or shifted ventricles, to be highlighted with higher distributional uncertainty, which is the OOD measure for the Bayesian deep learning models. To evaluate quality of OOD detection at a pixel level, we follow the procedure in \cite{chen2019unsupervised}, for example to get the 5.0$\%$ False Positive Ratio threshold value we compute the 95$\%$ percentile of distributional variance on the testing set of UKBB, taking into consideration that there is no outlier tissue there. Subsequently, using this value we threshold the distributional variance heatmaps on BraTS, with tissue having a value above the threshold being flagged as an outlier. We then quantify the overlap of the pixels detected as outliers (over the threshold) with the ground-truth tumor labels by computing the Dice score between them. 

\begin{figure*}[!htb]
    %\vskip 0.2in
    \centering
    \includegraphics[width=0.95\linewidth]{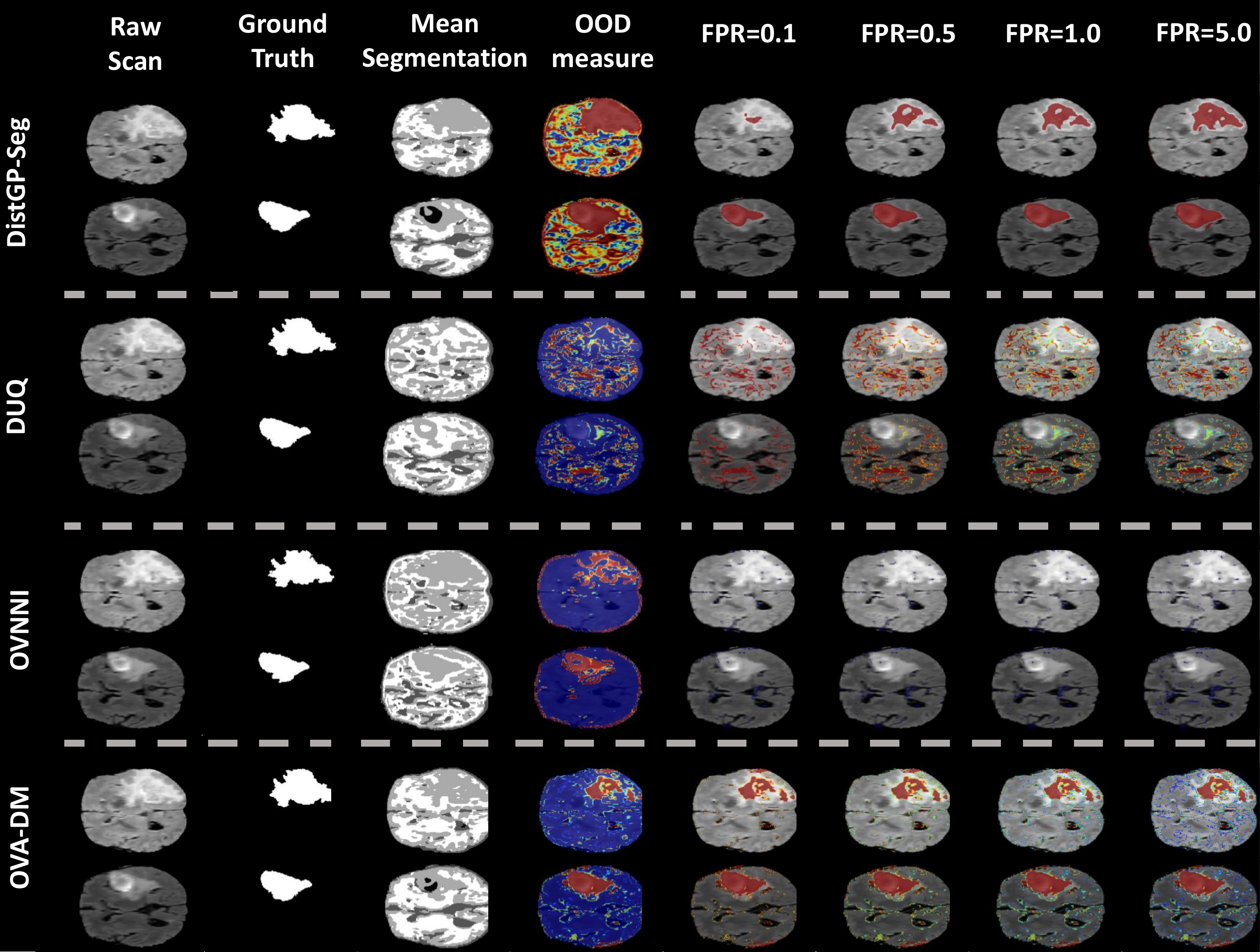}
    \caption{Comparison between models in terms of voxel-level outlier detection of tumors on BRATS scans. Mean segmentation represents the hard segmentation of brain tissues. OOD measure is the quantification of uncertainty for each model, using their own procedure. Higher values translate to appartenance to outlier status, whereas for OVNNI it is the converse. OOD measures have been normalized to be between 0 and 1 for each model in part.}
    \label{fig:tumor_class_comparison_brats}  
    %\vskip -0.2in
\end{figure*}

\paragraph{\textbf{Results:}} Table \ref{tab:ood_results} shows the results from our experiments with DistGP and compared Bayesian deep learning baselines. We also provide performance of reconstruction-based OOD detection models as reported in \cite{chen2019unsupervised} for similar experimental setup. DistGP-Seg surpasses its Bayesian deep learning counterparts, as well as reconstructed-based models. In Figure \ref{fig:tumor_class_comparison_brats} we provide representative results from the methods we implemented for qualitative assessment. Moreover, although BRATS does not provide labels for WM/GM/CSF tissues hence we cannot quantify how well these tissues are segmented, visual assessment shows our method compares favorably to compared counterparts.

\begin{figure}[!htb]
    %\vskip 0.2in
    \centering
    \subfigure[]{\includegraphics[width=0.48\linewidth]{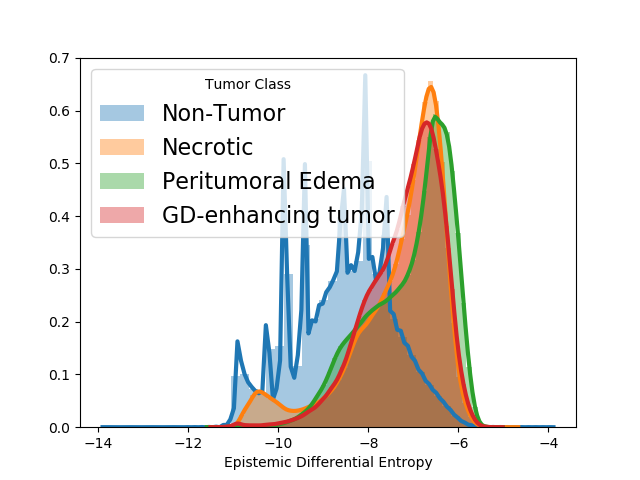}}
    \quad
    \subfigure[]{\includegraphics[width=0.48\linewidth]{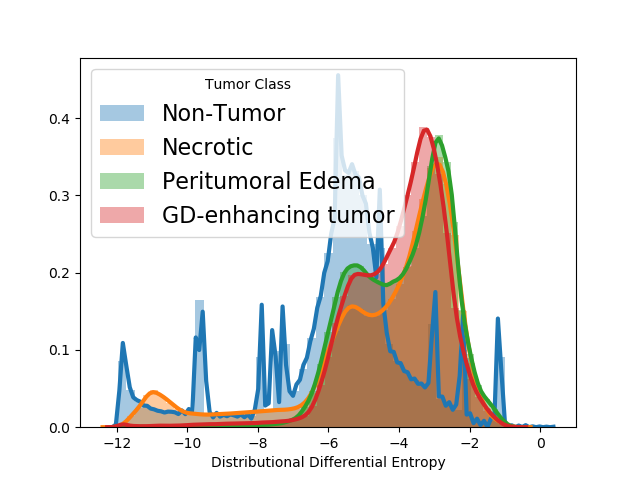}}   
    \quad
    \caption{Comparison in terms of voxel-level epistemic and distributional differential entropy between non-tumor tissues and different tumor gradations from subjects in the BRATS dataset.}
    \label{fig:tumor_class_comparison_brats_tissue}  
    %\vskip -0.2in
\end{figure}

In Figure \ref{fig:tumor_class_comparison_brats_tissue} we plotted the different differential entropy measures based on BRATS scans by overlying their tumor labels on the obtained uncertainties from our model. We can notice that tumor tissue is highlighted with higher inside and outside of the data manifold uncertainty compared to healthy tissue. More detailed plots are available in Appendix \ref{apd:tumors_ood}.

\section{Discussion \& Conclusion}

% 1. Summary of findings and novel contribution and 2. Relevance to previous literature
We have introduced a novel Bayesian convolutional layer with Lipschitz continuity that is capable of reliably propagating uncertainty. We have shown on a wide array of general OOD detection tasks that it surpasses other OOD models from literature, while also offering an increase in accuracy compared to counterpart architectures based solely on Euclidean space SVGPs \citep{blomqvist2018deep}. General criticism surrounding deep and convolutional GP involves the issue of under-performance compared to other Bayesian deep learning techniques, and especially compared to deterministic networks. Our experiments demonstrate that our 3-layers model, size limited due to computational cost, is capable of approaching the performance of a U-Net, an architecture with a much larger receptive field. Further advances in computational efficient GP-based models, an active area of research, will enable our model to scale further and unlock its full potential. Importantly, we showed that our DistGP-Seg network offers better uncertainty estimates for OOD detection than state-of-the-art OOD detection models, and also surpasses some recent unsupervised reconstruction-based deep learning models for identifying outliers corresponding to pathology on brain scans. 

% 3. Wider relevance and importance (technical, applied, clinical)

This framework can also be used for regression and classification tasks within a medical imaging context, facilitating the adoption of deep learning in clinical settings thanks to enhanced accountability in predictions. For example, parts of scans flagged with high distributional uncertainty can be sent back for inspection and quality control. To support our claim, we have included additional results on flagging white matter hyperintensities as outliers (see Appendix \ref{apd:wmh_ood}), respectively retina pathologies (see Appendix \ref{apd:retina_ood}).

% 4. Strengths

Our results indicate that OOD methods that do not take into account distances in latent space, such as OVNNI, tend to fail in detecting outliers, whereas OVA-DM and DUQ that make predictions based on distances in the last layer perform better. Our model utilises distances at every hidden layer, thus allowing the notion of outlier to evolve gradually through the depth of our network. This difference can be noticed in the smoothness of OOD measure for our model in comparison to other methods in Figure \ref{fig:tumor_class_comparison_brats}. Furthermore, the issue of \emph{feature collapse} \citep{van2020simple} in deep networks can be precisely controlled due to the mathematical underpinnings of our proposed network, enabling us to assess the scenarios when this happens by simple equations. Additionally, we have shown that despite the possibility of achieving over-correlation in the latent space via the loss function, that this does not happen in practice.

% 5. Weaknesses and Future research

A drawback of our study resides in the small architecture used on medical imaging scans. Extending our ``measure preserving DistGP'' module to larger architectures such as U-Net for segmentation or modern CNNs for whole-image prediction tasks remains a prospective research avenue fuelled by advances in scalability of SGP. Moreover, our experiments involving more complicated architectures, such as ResNet or DenseNet for standard multi-class classification, have not managed to surpass in accuracy a far less complex model with only 3 hidden layers. A plausible reason behind this under-fitting resides in the factorized approximate posterior formulation, which was shown to negatively affect predictive performance compared to MCMC inference schemes \citep{havasi2018inference}. We posit that using alternative inference frameworks \citep{ustyuzhaninov2019compositional} whereby we impose correlations between layers might alleviate this issue. Moreover, the lack of added representational capacity upon adding new layers raises some further questions regarding what are optimal architectures for hierarchical GPs, what inductive biases do they need or how to properly initialize them to facilitate adequate training. Additionally, our comparison with respect to reconstruction based approaches towards OOD detection was not complete as it did not include a comprehensive list of recent models \citep{10.1007/978-3-030-87240-3_23, pmlr-v143-pinaya21a, schlegl2019f, baur2018deep}. However, comparing our proposed model with reconstruction based approaches was not our intended goal for this paper, the main aim being to compare with models which can provide accurate predictive results alongside OOD detection capabilities at the same time. Another limitation of our work is the training speed for our proposed module, with matrix inversion operations and log determinants being required at each layer. Future work should consider matrix inversion free inference techniques for GPs \citep{van2020variational}.

% 6. Concluding statement
In conclusion, our work shows that incorporating DistGP in convolutional architectures provides both competitive performance and reliable uncertainty quantification in medical image analysis alongside general OOD tasks, opening up a new direction of research.

%%%%%%%%%%%%%%%%%%%%%%%%%%%%%%%%%%%%%%%%%%%%%%%%%%%%%%%%%%%%%%%%%%%%%%%
% Mandatory Sections. Please complete, especially for final publication
%%%%%%%%%%%%%%%%%%%%%%%%%%%%%%%%%%%%%%%%%%%%%%%%%%%%%%%%%%%%%%%%%%%%%%%

% Acknowledgements.
% Please include any funding, intellectual contributions not included in the authorship, and any other acknowledgements.
\acks{SGP is funded by an EPSRC Centre for Doctoral Training studentship award to Imperial College London. KK is funded by the UKRI London Medical Imaging \& Artificial Intelligence Centre for Value Based Healthcare. BG received funding from the European Research Council (ERC) under the European Union's Horizon 2020 research and innovation programme (grant agreement No 757173, project MIRA, ERC-2017-STG). DJS is supported by the NIHR Biomedical Research Centre at Imperial College Healthcare NHS Trust and the UK Dementia Research Institute (DRI) Care Research and Technology Centre. JHC acknowledges funding from UKRI/MRC Innovation Fellowship (MR/R024790/2).}

% Ethical Standards.
% Please edit with the appropriate ethics considerations for your work. Include any pertinent IRB information, etc.
%
% Please note that the submission requirements included:
% The work presented must follow appropriate ethical standards in conducting research and writing the manuscript, following all applicable laws and regulations regarding treatment of animals or human subjects.
\ethics{The work follows appropriate ethical standards in conducting research and writing the manuscript, following all applicable laws and regulations regarding treatment of human subjects.}

% Conflict of Interest
% Declaration of possible conflicts of interest: Authors must disclose any financial, organisational, commercial or personal conflicts of interest that might bias their work.
% If no conflicts, please say "We declare we don't have conflicts of interest."
\coi{BG has received grants from European Commission and UK Research and Innovation Engineering and Physical Sciences Research Council, during the conduct of this study; and is Scientific Advisor for Kheiron Medical Technologies and Advisor and Scientific Lead of the HeartFlow-Imperial Research Team. JHC is a shareholder in and Scientific Advisor to BrainKey and Claritas Healthcare, both medical image analysis software companies.}

\bibliography{library}

% needed in general before appendices.
\newpage
\appendix % optional
\section{Proving Lipschitz bounds in a DistGP layer}\label{apd:lipschitz_proofs}

We here prove Propositions \ref{thm:lipschitz_dist_gp}  and \ref{thm:lipschitz_affine_layer} of Sec.~\ref{sec:lipschitz}. 

\paragraph{\textbf{Lemmas on p-norms/}} We have the following relations between norms :
$\Vert x \Vert_{2} \leq \Vert x \Vert_{1} $ and $\Vert x \Vert_{1} \leq \sqrt{D} \Vert x \Vert_{2}$. Will be used for the proof of Proposition 2.

\iffalse
\begin{gather}
    \Vert x \Vert_{2} \leq \Vert x \Vert_{1}  \label{eqn:p_norm_inequality1} \\
    \Vert x \Vert_{1} \leq \sqrt{D} \Vert x \Vert_{2} \label{eqn:p_norm_inequality2}
\end{gather}
\fi

\paragraph{\textbf{Proof of Proposition  \ref{thm:lipschitz_dist_gp}.}}

Throughout this proof we shall refer to the first two moments of a Gaussian distribution by $m(\cdot)$, $v(\cdot)$. Explicitly writing the Wasserstein-2 distances of the inequality we get: 
\begin{equation}
 |m(F(\mu)) - m(F(\nu))|^{2} + |v(F(\mu)) - v(F(\nu)|^{2} \leq L |m_{1} - m_{2}|^{2} +  |\Sigma_{1} - \Sigma_{2}|^{2} 
\end{equation}
 We focus on the mean part and applying Cauchy–Schwarz we get the following inequality:
 \begin{equation}
 | \left[ K_{\mu u} - K_{\nu u} \right] K_{uu}^{-1}m |^{2} \leq \Vert K_{\mu u} - K_{\nu u} \Vert_{2}^{2} \Vert K_{uu}^{-1}m \Vert_{2}^{2}
 \end{equation}
To simplify the problem and without loss of generality we consider $U_{z}$ to be a sufficient statistic for the set of inducing points $Z$. Expanding the first term of the r.h.s. we get:
\begin{equation}
    \left[ \sigma^{2} \exp{\frac{-W_{2}(\mu,U_{z})}{l^{2}}} - \sigma^{2} \exp{\frac{-W_{2}(\nu, U_{z}) } {l^{2}}} \right]^{2}     
\end{equation}
We assume $\nu = \mu + h$, where $h \sim \mathcal{N}(|m_{1}-m_{2}|,|\Sigma_{1} - \Sigma_{2}|)$ and  $\mu$ is a high density point in the data manifold, hence $W_{2} (\mu - U_{z}) = 0$. We denote $m(h)^{2}+var(h)^{2}=\lambda$. Considering the general equality $\log(x-y) = \log(x) + \log(y) + \log(\frac{1}{y} - \frac{1}{x})$ and applying it to our case we get that:
\begin{align}
    \log \mid m(F(\mu)) - m(F(\nu)) \mid^{2} &\leq \log\left[ \sigma^{2} -\sigma^{2} \exp\frac{{-\lambda}}{l^{2}} \right]^{2} \\
    & \leq 2\log \sigma^{2} - 2 \frac{\lambda}{l^{2}} + 2\log \left[ \exp{\frac{\lambda}{l^{2}}} -1 \right] \\ 
    & \leq 2 \log \left[ \sigma^{2} \exp{\frac{\lambda}{l^{2}}}  \right]
\end{align}
We have the general inequality $\exp{x} \leq 1+x+x^{2}$ for $x\leq 1.79$, which for $0\leq x \leq 1$ can be modified as $\exp{x} \leq 1+2x$. Applying this new inequality and taking the exponential we now obtain:
\begin{align}
    \mid m(F(\mu)) - m(F(\nu)) \mid^{2} &\leq \left[ \sigma^{2} + 2\sigma^{2}\frac{\lambda}{l^{2}}\right]^{2} \\
    &\leq \sigma^{4} + \sigma^{4} \frac{\lambda}{l^{2}} + 4 \sigma^{4} \frac{(\lambda)^{2}}{l^{4}} \\ 
    &\leq 16 \sigma^{4}\frac{\lambda}{l^{2}}    
\end{align}
where the last inequality follows from the ball constraints made in the definition. We now move to the variance components of the Lipschitz bound, we notice that 
\begin{align}
    \mid v(F(\mu))^{\frac{1}{2}} - v(F(\nu))^{\frac{1}{2}} \mid^{2} &\leq \mid v(F(\mu))^{\frac{1}{2}} - v(F(\nu))^{\frac{1}{2}} \mid\mid v(F(\mu))^{\frac{1}{2}} + v(F(\nu))^{\frac{1}{2}} \mid \\ 
    & \leq \mid v(F(\mu)) - v(F(\nu)) \mid 
\end{align}
which after applying Cauchy–Schwarz results in an upper bound of the form:
\begin{equation}
    \Vert K_{\mu,U_{z}} - K_{\nu,U_{z}}\Vert_{2}^{2} \Vert K_{U_{z}}^{-1}(K_{U_{z}-S})K_{U_{z}}^{-1}\Vert_{2}    
\end{equation}
 Using that $\Vert K_{\mu,U_{z}} - K_{\nu,U_{z}} \Vert_{2}^{2} \leq  \frac{16\sigma^{4} \lambda}{l^{2}}$ we obtain that:
 \begin{equation}
     |v(F(\mu)) - v(F(\nu))| \leq \frac{16\sigma^{4}\lambda}{l^{2}} \Vert K_{U_{z}}^{-1}(K_{U_{z}}-S)K_{U_{z}}^{-1}\Vert_{2}     
 \end{equation}

 Now taking into consideration both the upper bounds on the mean and variance components we arrive at the desired Lipschitz constant.

\paragraph{\textbf{Proof of Proposition \ref{thm:lipschitz_affine_layer}.}}

Using the definition for Wasserstein-2 distances for the l.h.s of the inequality, we can re-express as follows:
\begin{equation}
    W_{2}\left(f(\mu), f(\nu)\right) \leq \Vert m_{1}A-m_{2}A\Vert_2^2 +\Vert(\sigma_{1}^{2}A^{2})^{1/2}-(\sigma_{2}^{2}A^{2})^{1/2}\Vert_{F}^2
\end{equation}
which after rearranging terms and noticing that inside the Frobenius norm we have scalars, becomes:
\begin{equation}
    W_{2}\left(f(\mu), f(\nu)\right) \leq \Vert (m_{1}-m_{2})A\Vert_2^2+ [\sigma_{1}^{2}A^{2})^{1/2}-(\sigma_{2}^{2}A^{2})^{1/2}]^2
\end{equation}
We can now apply the Cauchy–Schwarz inequality for the part involving means and multiplying the right hand side with $\sqrt{C}$, which represents the number of channels, we get: 
\begin{equation}
    \Vert (m_{1}-m_{2})A\Vert_2^2 + [\sigma_{1}^{2}A^{2})^{1/2}-(\sigma_{2}^{2}A^{2})^{1/2}]^2 \leq \Vert m_{1}-m_{2} \Vert_2^{2} \sqrt{C}\Vert A \Vert_2^{2} + \sqrt{C} [\sigma_{1}^{2}A^{2})^{1/2}-(\sigma_{2}^{2}A^{2})^{1/2}]^2    
\end{equation}
We can notice that the Lipschitz constant for the component involving mean terms is $\sqrt{C}\Vert A \Vert_2^{2}$. Hence, we try to prove that the same L is also available for the variance terms component. Hence, we can affirm that: 
\begin{equation}
    L=\sqrt{C}\Vert A \Vert_2^{2}  \leftrightarrow \sqrt{C} [\sigma_{1}^{2}A^{2})^{1/2}-(\sigma_{2}^{2}A^{2})^{1/2}]^2 \leq [\sigma_{1}-\sigma_{2}]^{2}  \sqrt{C}\Vert A \Vert_2^{2}     
\end{equation}
By virtue of Cauchy–Schwarz we have the following inequality
$\sqrt{C}[\sigma_{1}A-\sigma_{2}A]^{2} \leq [\sigma_{1}-\sigma_{2}]^{2}  \sqrt{C}\Vert A \Vert_2^{2}$.
Hence the aforementioned if and only if statement will hold if we prove that 
\begin{equation}
    \sqrt{C}\left[(\sigma_{1}^{2}A^{2})^{\frac{1}{2}} - (\sigma_{2}^{2}A^{2})^{\frac{1}{2}}\right]^{2} \leq \sqrt{C}\left[\sigma_{1}A-\sigma_{2}A\right]^{2}     
\end{equation}
which after expressing in terms of norms becomes:
\begin{align}
    \sqrt{C}\left[ \Vert\sigma_{1}A\Vert_{2} - \Vert\sigma_{2}A\Vert_{2}\right]^{2} \leq \sqrt{C}\left[ \Vert\sigma_{1}A\Vert_{1} - \Vert\sigma_{2}A\Vert_{1} \right]^{2}   
\end{align}

\iffalse
\begin{gather}
    \sqrt{C}\left[(\sigma_{1}^{2}W^{2})^{\frac{1}{2}} - (\sigma_{2}^{2}W^{2})^{\frac{1}{2}}\right]^{2} \leq \sqrt{C}\left[\sigma_{1}W-\sigma_{2}W\right]^{2} \\
    \sqrt{C}\left[\Vert\sigma_{1}W\Vert_{2} - \Vert\sigma_{2}W\Vert_{2}\right]^{2} \leq \sqrt{C}\left[\Vert\sigma_{1}W\Vert_{1} - \Vert\sigma_{2}W\Vert_{1}\right]^{2}
\end{gather}
\fi
Expanding the square brackets gives:
\begin{align}
    \sqrt{C}\left[\Vert\sigma_{1}A\Vert_{2}^{2}+ \Vert\sigma_{2}A\Vert_{2}^{2} -  2\Vert\sigma_{1}A\Vert_{2}\Vert\sigma_{2}A\Vert_{2}\right] \leq \sqrt{C}\left[\Vert\sigma_{1}A\Vert_{1}^{2}+\Vert\sigma_{2}A\Vert_{1}^{2}-2\Vert\sigma_{1}A\Vert_{1}\Vert\sigma_{2}A\Vert_{1}\right]    
\end{align}

This inequality holds by applying the p-norm lemma, thereby the if and only if statement is satisfied. Consequently, the Lipschitz constant is $\sqrt{C}\| A \|_2^{2}$.

\section{Deriving function contraction requirements in DistGP Layers} \label{apd:feature_collapse}

We here prove Proposition \ref{thm:feature_collapse} of Sec.~\ref{sec:function_space_dist_gp_layers}. 

\paragraph{Proof of Proposition \ref{thm:feature_collapse}.} 

 We are interested in determining the specific scenarios in which the function space collapses to constant values. Hence we explicitly write $\mathbb{E}\left[ \| u_{l}(x) - u_{l}(x^{*}) \|_{2}^{2} \mid u_{l-1} \right]$ as:
\begin{align}
    &= \sum\limits_{j=1}^{D_{l}} \mathbb{E}\left[ \| u_{l}^{j}(x) - u_{l}^{j}(x^{*}) \|_{2}^{2} \mid u_{l-1} \right] \\
    &= \sum\limits_{j=1}^{D_{l}}  \mathbb{E} \left[ \left(u_{l}^{j}(x)\right)^{2} \mid u_{l-1} \right] - 2 \mathbb{E} \left[ u_{l}^{j}(x)  u_{l}^{j}(x^{*}) \mid u_{l-1} \right]+ \mathbb{E}\left[ \left(u_{l}^{j}(x^{*})\right)^{2} \mid u_{l-1} \right] \\
    &= \sum\limits_{j=1}^{D_{l}} \sigma^{2}_{l} + m_{l}^{2}(x) -2m_{l}(x)m_{l}(x^{*}) -2k^{W_{2}}\left[\mu_{l}(x), \mu_{l}(x^{*}) \right] + \sigma_{l}^{2} + m_{l}^{2}(x^{*}) \\
    &= \sum\limits_{j=1}^{D_{l}} \left[ m_{l}^{j}(x) - m_{l}^{j}(x^{*})\right]^{2} + 2\sigma_{l}^{2} - 2\sigma_{l}^{2} \exp{-\left[ \frac{\mid m_{l-1}^{j}(x) - m_{l-1}^{j}(x^{*}) \mid^{2}}{2l_{l}^{2}}\right]}
\end{align}
, where in the last equation we have ignored the variance part of the Wasserstein-2 kernel since the two variance terms are equal. We make use of the following inequality $1-\exp{-x} \leq x$ for $x\geq 0$ and equality only in the case that $x=0$, resulting in the following upper bound:
\begin{equation}
    \mathbb{E}\left[ \| u_{l}(x) - u_{l}(x^{*}) \|_{2}^{2} | u_{l-1} \right] \leq  \sum\limits_{j=1}^{D_{l}} \left[ m_{l}^{j}(x) - m_{l}^{j}(x^{*})\right]^{2}  + \sigma_{l}^{2} \frac{\mid m_{l-1}^{j}(x) - m_{l-1}^{j}(x^{*}) \mid^{2}}{l_{l}^{2}} \label{eqn:ref_this_1}
\end{equation}
We can now view the previously defined operator $\overline{m_{l}(x)W_{l}}$ as an inner product in vector space between a tiled version of $m_{l}(x)$ and a normalised version of $W_{l}$, more specifically:
\begin{equation}
    \langle \left[m_{l}(x),\cdots,m_{l}(x)\right],\left[\frac{W_{l,1}}{m_{l}},\cdots,\frac{W_{l,m_{l}D_{l-1}}}{m_{l}}\right] \rangle = \overline{m_{l}(x)W_{l}}
\end{equation}
where $m_{l}$ is the number of dimensions caused by the affine embedding function $\Psi_{l}$ in the l-th layer of the hierarchy. 

Our current goal is to relate $m_{l}^{j}(\cdot)$ to $ m_{l-1}^{j}(\cdot)$. We can now apply Cauchy-Schwarz to:
\begin{gather}
    \mid m_{l}^{j}(x) - m_{l}^{j}(x^{*}) \mid^{2} = \mid \overline{m_{l-1}(x)W_{l}} - \overline{m_{l-1}(x^{*})W_{l}} \mid^{2} \\
    = \mid  \langle \left[m_{l-1}(x) - m_{l-1}(x^{*}) ,\cdots, m_{l-1}(x) - m_{l-1}(x^{*})\right],\left[\frac{W_{l,1}}{m_{l}},\cdots,\frac{W_{l,m_{l}D_{l-1}}}{m_{l}}\right] \rangle  \mid^{2} \\
    \leq D_{l-1}m_{l} \left[m_{l-1}(x) - m_{l-1}(x^{*})\right]^{2} * \langle \tilde{W_{l}}, \tilde{W_{l}}\rangle
\end{gather}
where in the last line we denoted $\tilde{W_{l}} = [\frac{W_{l,1}}{D_{l}},\cdots,\frac{W_{l,D_{l-1}m_{l}}}{m_{l}}]$ to avoid cluttering.

We can now apply the previous result to equation \eqref{eqn:ref_this_1}:
\begin{align}
    \mathbb{E}\left[ \| u_{l}(x) - u_{l}(x^{*}) \|_{2}^{2} |\mid u_{l-1} \right] &\leq \sum\limits_{j=1}^{D_{l}}  m_{l}D_{l-1} \left(m_{l-1}(x) - m_{l-1}(x^{*})\right)^{2} * \langle \tilde{W_{l}}, \tilde{W_{l}}\rangle \\ & \nonumber \hspace{1cm} + \frac{ \sigma_{l}^{2}}{l_{l}^{2}} \mid m_{l-1}^{j}(x) - m_{l-1}^{j}(x^{*}) \mid^{2} \\
    &\leq \sum\limits_{j=1}^{D_{l}}  \left[ m_{l}D_{l-1} * \langle \tilde{W_{l}}, \tilde{W_{l}}\rangle + \frac{ \sigma_{l}^{2}}{l_{l}^{2}} \right]  \mid m_{l-1}^{j}(x) - m_{l-1}^{j}(x^{*}) \mid^{2}
\end{align}
We can now recursively apply the previously derived Cauchy-Schwarz based inequality to obtain:
\begin{align}
    \mathbb{E}\left[ \mid\mid u_{l}(x) - u_{l}(x^{*}) \mid\mid_{2}^{2} | \{u_{l}\}_{l=1}^{l-1} \right]  & \leq \left[ m_{l}D_{l-1} * \langle \tilde{W_{l}}, \tilde{W_{l}}\rangle + \frac{ \sigma_{l}^{2}}{2l_{l}^{2}} \right] \prod\limits_{l=1}^{l-1}D_{l}m_{l}D_{l-1}\langle \tilde{W_{l}}, \tilde{W_{l}}\rangle \\ & \nonumber \hspace{1cm} \left[m_{1}(x) - m_{1}(x^{*}) \right]^{2}
\end{align}
By Markov's inequality, for any $\epsilon > 0$ we have that:
\begin{align}
    P\left( \mid\mid u_{l+1}(x) - u_{l+1}(x^{*})  \mid\mid_{2} \geq \epsilon \right) &\leq \frac{1}{\epsilon^{2}} \left[ m_{l} D_{l-1} * \langle \tilde{W_{l}}, \tilde{W_{l}}\rangle + \frac{ \sigma_{l}^{2}}{2l_{l}^{2}} \right] \prod\limits_{l=1}^{l-1} D_{l}m_{l}D_{l-1}\langle \tilde{W_{l}}, \tilde{W_{l}}\rangle \\ & \nonumber \hspace{1cm} \left[m_{1}(x) - m_{1}(x^{*}) \right]^{2}
\end{align}

Then, only in the case that $\left[ m_{l}D_{l-1} * \langle \tilde{W_{l}}, \tilde{W_{l}}\rangle + \frac{ \sigma_{l}^{2}}{2l_{l}^{2}} \right] \leq 1$ and $D_{l}m_{l}D_{l-1}\langle \tilde{W_{l}}, \tilde{W_{l}}\rangle \leq$ is satisfied for intermediate layers, we can apply the first Borel-Cantelli lemma to obtain:
\begin{equation}
    P\left( \cap_{l=1}^{\infty} \cup_{m=l}^{\infty} \mid\mid  u_{m}(x) - u_{m}(x^{*}) \mid\mid_{2} \geq \epsilon \right) = 0
\end{equation}
Lastly, we can express the following:
\begin{align}
    P\left( \mid\mid  u_{n}(x) - u_{n}(x^{*}) \mid\mid_{2} \to 0 \right) &= P\left( \cap_{k=1}^{\infty} \cup_{l=1}^{\infty} \cap_{m=l}^{\infty} \mid\mid  u_{m}(x) - u_{m}(x^{*}) \mid\mid_{2} \leq \frac{1}{k} \right) \\
    &= 1 - P\left( \cup_{k=1}^{\infty} \cap_{l=1}^{\infty} \cup_{m=l}^{\infty} \mid\mid  u_{m}(x) - u_{m}(x^{*}) \mid\mid_{2} \geq \frac{1}{k} \right) \\
    &\geq 1 - \sum\limits_{k=1}^{\infty} P\left( \cap_{l=1}^{\infty} \cup_{m=l}^{\infty} \mid\mid  u_{m}(x) - u_{m}(x^{*}) \mid\mid_{2} \geq \frac{1}{k} \right) = 1
\end{align}
From which we obtain the proof of our proposition, respectively $ P\left( \mid\mid  u_{n}(x) - u_{n}(x^{*}) \mid\mid_{2} \to 0 \right) = 1$

\section{Outlier detection in MRI scans with Tumors.} \label{apd:tumors_ood}

\begin{figure}[!htb]
    %\vskip 0.2in
    \centering
    \includegraphics[width=0.95\linewidth]{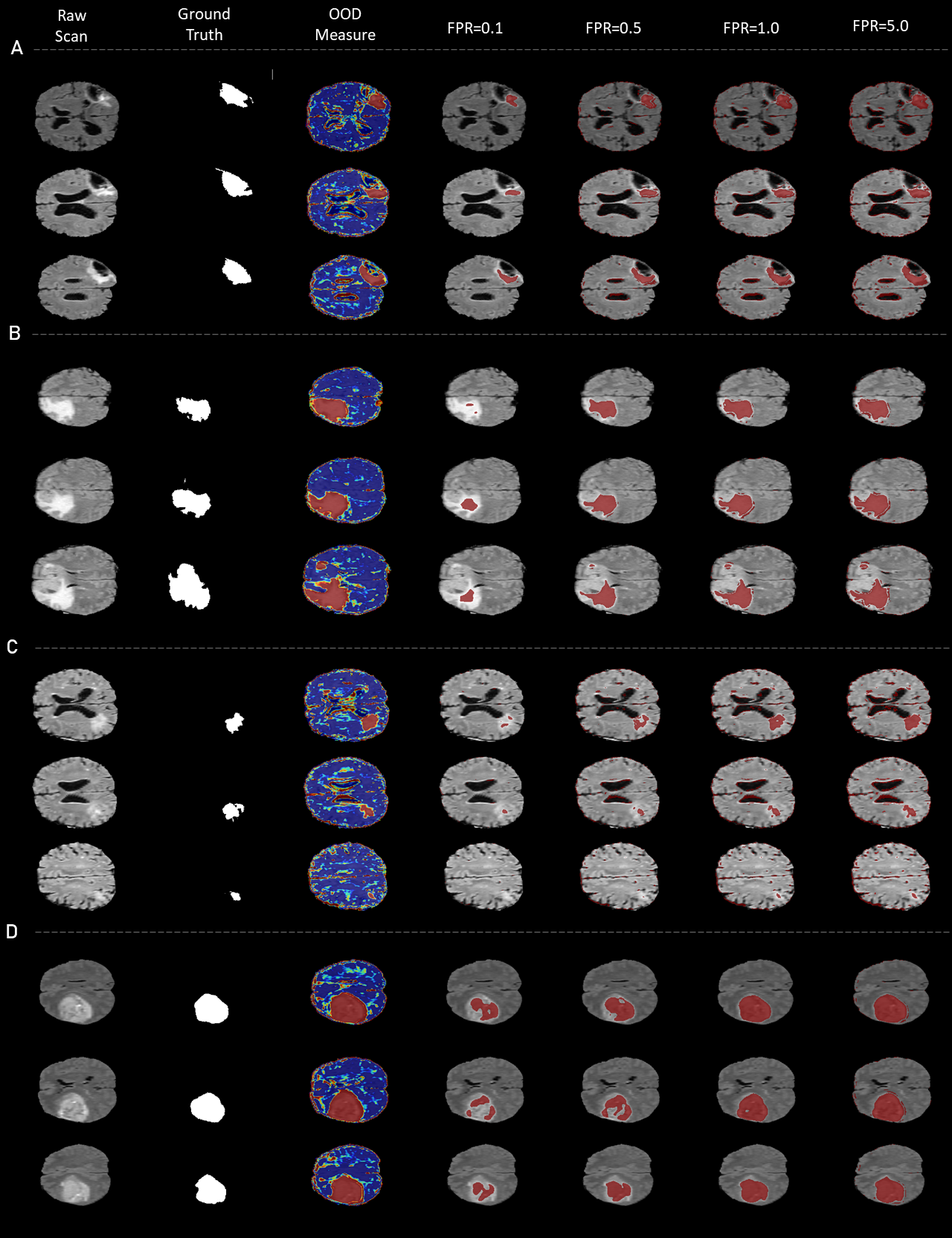}
    \caption{Detailed segmentation output for DistGP-Seg on BRATS. Mean segmentation represents the hard segmentation of brain tissues. OOD measure is the quantification of uncertainty for each model, using their own procedure. Higher values translate to appartenance to outlier status.}
    \label{fig:case_studies_dist_gp_layers_tumor}  
    %\vskip -0.2in
\end{figure}

\begin{figure}[!htb]
    %\vskip 0.2in
    \centering
    \includegraphics[width=0.95\linewidth]{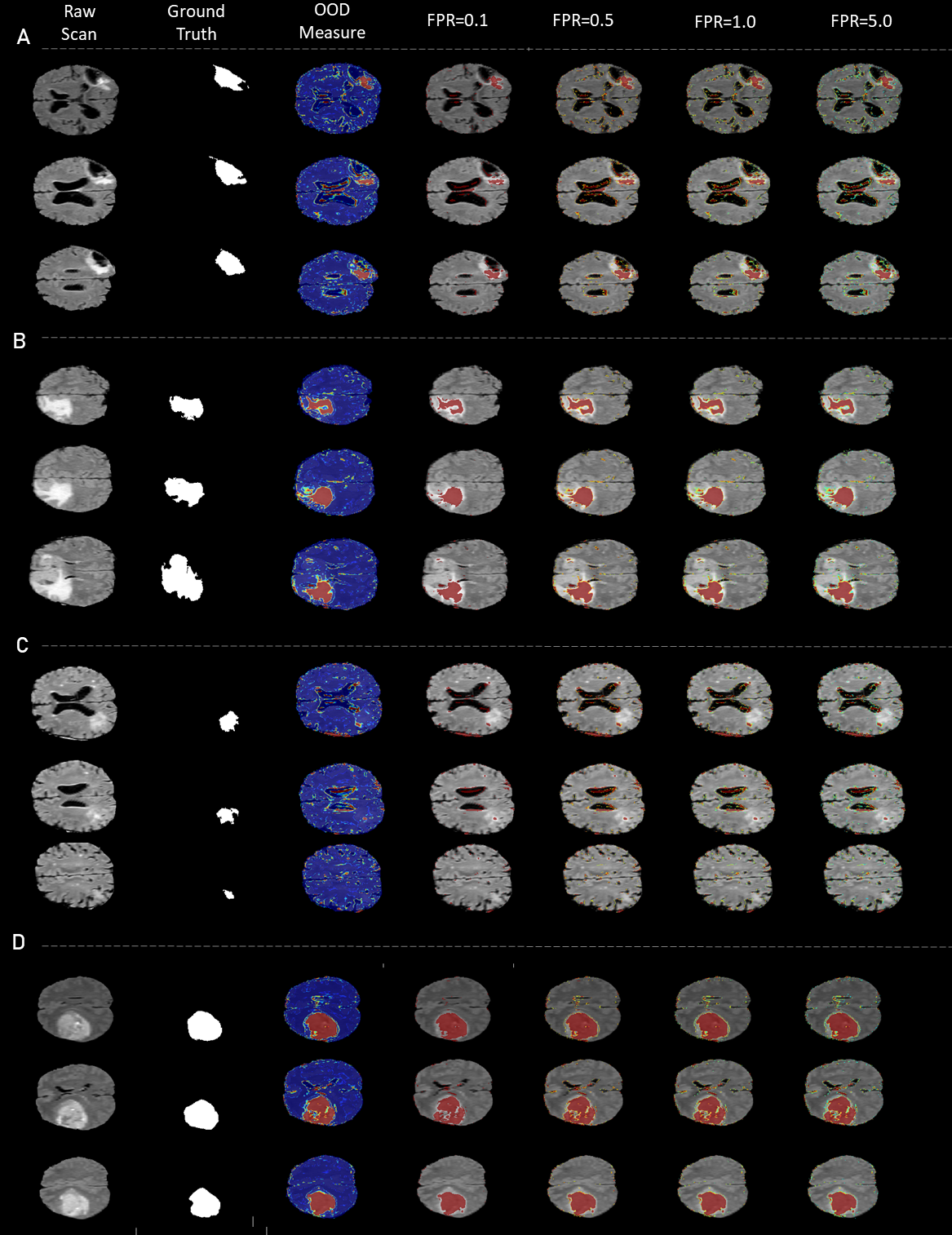}
    \caption{Detailed segmentation output for OVA-DM on BRATS. Mean segmentation represents the hard segmentation of brain tissues. OOD measure is the quantification of uncertainty for each model, using their own procedure. Higher values translate to appartenance to outlier status.}
    \label{fig:case_studies_ova_dm_tumor}  
    %\vskip -0.2in
\end{figure}

\paragraph{Remarks.} We provide additional plots for the task investigated in sec.~\ref{sec:outlier_tumors_brats} for DistGP-Seg and OVA-DM as they were the only models to provide decent outlier detection capabilities. We refer the reader to Figures \ref{fig:case_studies_dist_gp_layers_tumor} and \ref{fig:case_studies_ova_dm_tumor}. From case study A, we can see that OVA-DM is over-segmenting across all FPR levels almost randomly from outside the tumor area, whereas DistGP-Seg is over-segmenting at $FPR=\{1.0,5.0\}$ areas around the margins of the ventricles. For case study B, at $FPR=\{0.5,1.0,5.0\}$ OVA-DM seems to be under-segmenting the tumor in comparison to DistGP-Seg. The same observation can be made again for case study C. Lastly, for case study D DistGP-Seg seems to be under-segmenting for $FPR=\{0.1,0.5\}$.

\section{Outlier detection in MRI scans with WMH.} \label{apd:wmh_ood}

\paragraph{Data and pre-processing.} Brain MRI scans from the 2015 Longitudinal Multiple Sclerosis Lesion Segmentation Challenge \cite{sweeney2013automatic} which comprises of FLAIR, PD, T2-weighted, and T1-
weighted volumes from a total of 110 MR imaging studies (11 longitudinal studies each of 10 subjects). All participants gave written consent and were scanned as part of an institutional review board approved natural history protocol. For the purposes of the task at hand, we only use the baseline FLAIR scans. All FLAIR images are pre-processed with skull-stripping, N4 bias correction, rigid registration to MNI152 space and histogram matching between UKBB and BraTS. Finally, we normalize intensities of each scan via linear scaling of its minimum and maximum intensities to the [-1,1] range.

\begin{figure*}[!htb]
    %\vskip 0.2in
    \centering
    \includegraphics[width=0.85\linewidth]{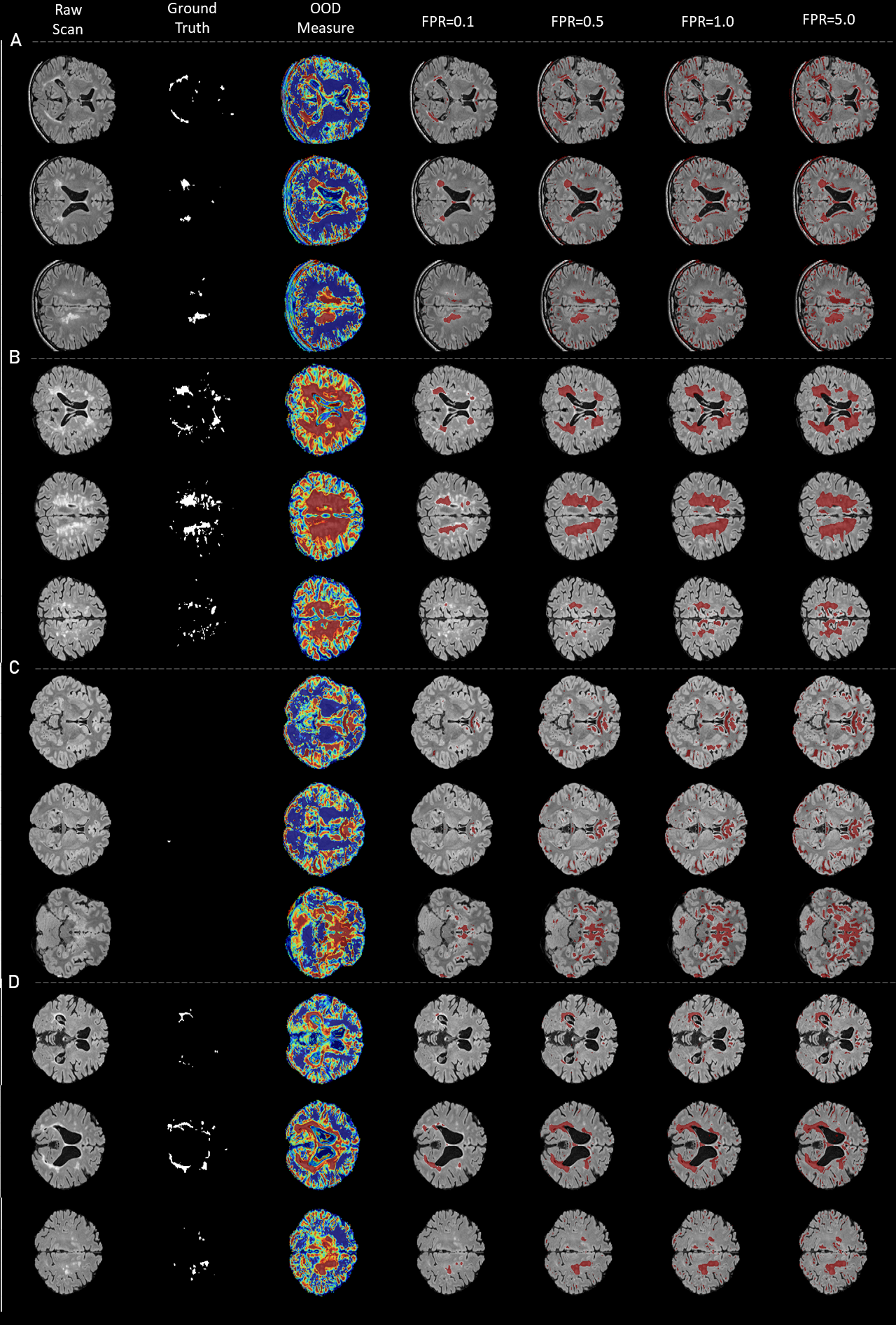}
    \caption{Detailed segmentation output for DistGP-Seg on WMH dataset. Mean segmentation represents the hard segmentation of brain tissues. OOD measure is the quantification of uncertainty for each model, using their own procedure. Higher values translate to appartenance to outlier status.}
    \label{fig:case_studies_dist_gp_layers_wmh}  
    %\vskip -0.2in
\end{figure*}

\paragraph{Remarks.} The task of detecting white matter hyperintensities (WMH) is considerably more difficult than detecting tumors, the latter usually presenting itself as a large blob, whereas the former constitutes of multiple non-contiguous areas of varying shapes. From Figure \ref{fig:case_studies_dist_gp_layers_wmh} we can notice that the large connected WMH regions are reliably detected as outliers, with smaller disconnected WMH regions being only in some cases outlined. Another issue is over-segmentation, as seen in case study C.

\iffalse
axial slice - 90
0.0665223761521192
0.08985292271874176
0.07421984394280176
0.06367589501217948
\fi

\section{Evaluation on Retina scans.} \label{apd:retina_ood}

\paragraph{Data and pre-processing.}

\textbf{DRIVE:} The Digital Retinal Images for Vessel Extraction (DRIVE) dataset \cite{staal2004ridge} is a dataset for retinal vessel segmentation. It consists of a total of JPEG 40 color fundus images; including 7 abnormal pathology cases. The images were obtained from a diabetic retinopathy screening program in the Netherlands. The images were acquired using Canon CR5 non-mydriatic 3CCD camera with FOV equals to 45 degrees. Each image resolution is 584*565 pixels with eight bits per color channel (3 channels).

The set of 40 images was equally divided into 20 images for the training set and 20 images for the testing set. Inside both sets, for each image, there is circular field of view (FOV) mask of diameter that is approximately 540 pixels. Inside training set, for each image, one manual segmentation by an ophthalmological expert has been applied. Inside testing set, for each image, two manual segmentations have been applied by two different observers, where the first observer segmentation is accepted as the ground-truth for performance evaluation.

\textbf{STARE:} STructured Analysis of the Retina (STARE) database \cite{hoover2000locating} was created by scanning and digitizing the retinal image photographs. Hence, the image quality of this database is less than the other public databases. The STARE dataset comprises 97 images (59 AMD and 38 normal) taken using a fundus camera (TOPCON TRV-50; Topcon Corp., Tokyo, Japan) at a $35\degree$ field and with a resolution of $605 \times 700$ pixels. Its retina scans are from subjects suffering from the following retina pathologies: Hollenhorst Emboli
Branch Retinal Artery Occlusion, Cilio-Retinal Artery Occlusion, Branch Retinal Vein Occlusion, Central Retinal Vein Occlusion, Hemi-Central Retinal Vein Occlusion, Background Diabetic Retinopathy, Proliferative Diabetic Retinopathy, Arteriosclerotic Retinopathy, Hypertensive Retinopathy, Coat's, Macroaneurism, Choroidal  Neovascularization.	

\textbf{IDRID:}  The Indian Diabetic Retinopathy Image Dataset (IDRID) dataset \cite{porwal2018indian}, is a publicly available retinal fundus image database consisting of 516 images categorised into two parts: retina images with the signs of Diabetic Retinopathy and/or Diabetic Macular Edema; normal retinal images. Images were acquired using a Kowa VX-10a digital fundus camera
with $50\degree$ field of view (FOV). The images have resolution of $4288 \times 2848$ pixels and are stored in jpg file format. We have pre-processed these images to match the FOV and resolution of DRIVE.

\paragraph{Task.} We train a similar DistGP-Seg architecture (see sec. \ref{sec:medical_imaging}) adapted for 2D data on DRIVE (normative data) to segment blood vessels, subsequently at testing time we use STARE and IDRID (OOD data) to segment blood vessels in the presence of previouslt unseen pathologies on Retina scans.

\paragraph{Blood vessel segmentation on normal Retina scans.}

\begin{figure*}[!htb]
    %\vskip 0.2in
    \centering
    \includegraphics[width=0.85\linewidth]{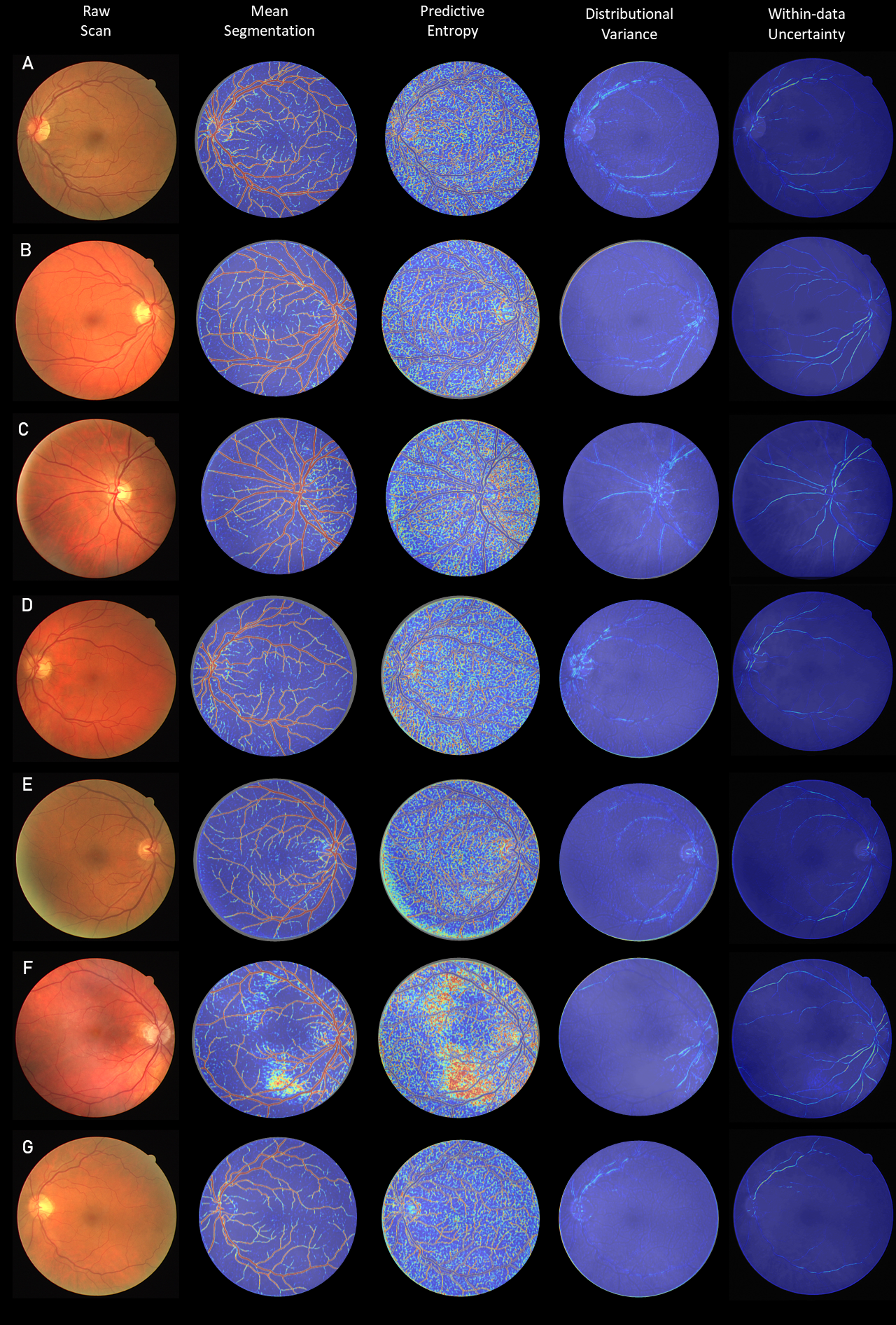}
    \caption{Detailed segmentation output for DistGP-Seg DRIVE dataset. Mean segmentation represents the hard segmentation of brain tissues. OOD measure is the quantification of uncertainty for each model, using their own procedure. Higher values translate to appartenance to outlier status.}
    \label{fig:case_studies_dist_gp_layers_retina_normal}  
    %\vskip -0.2in
\end{figure*}

From Figure \ref{fig:case_studies_dist_gp_layers_retina_normal} we can observe that DistGP-Seg manages to correctly segment the blood vessels, while both distributional and within-data uncertainty are relatively low, which is to be expected as these testing examples represent in-distribution data.

\paragraph{Outlier detection of Retina pathologies.}

\begin{figure*}[!htb]
    %\vskip 0.2in
    \centering
    \includegraphics[width=0.85\linewidth]{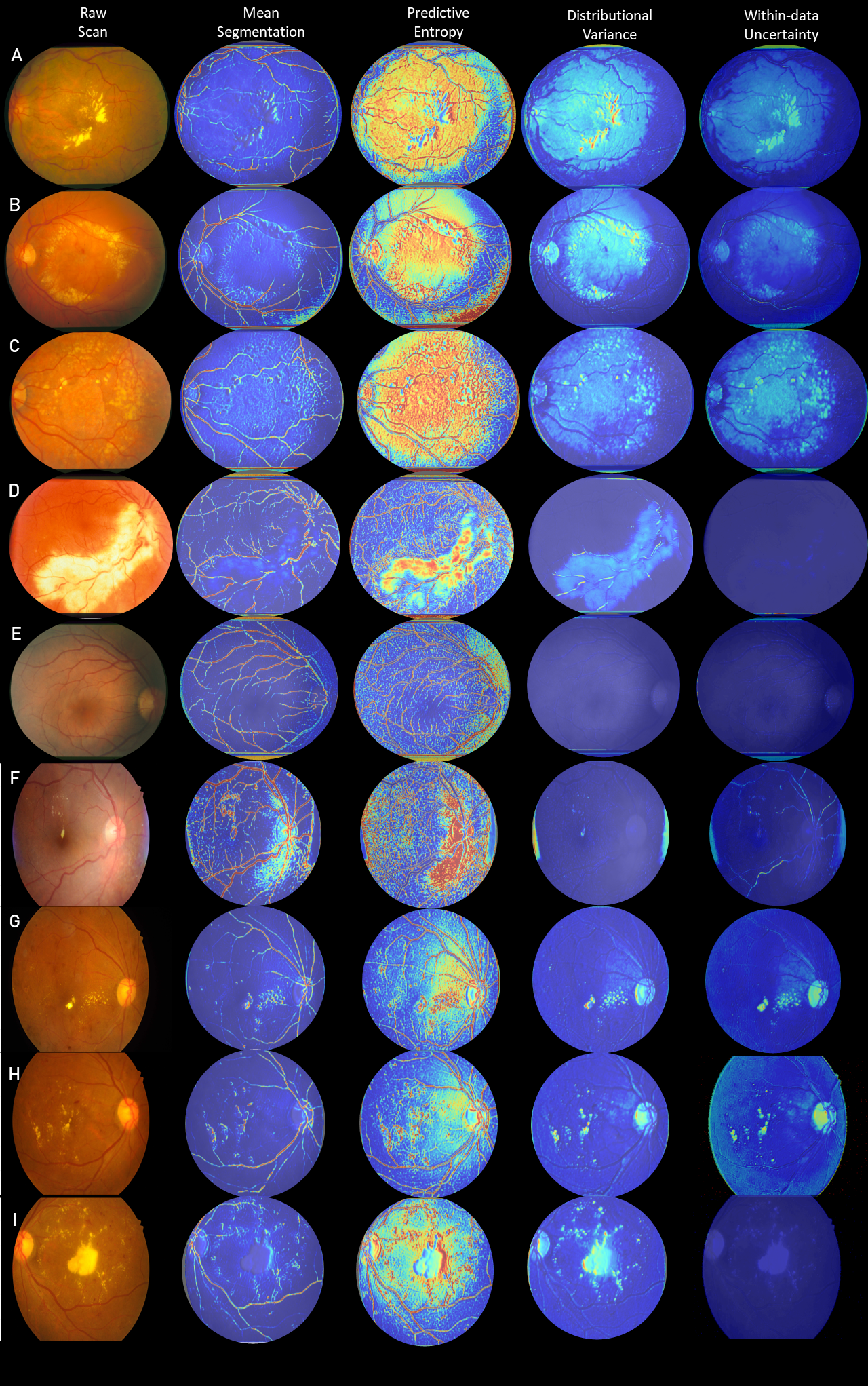}
    \caption{Detailed segmentation output for DistGP-Seg on STARE/IDRID datasets. Mean segmentation represents the hard segmentation of brain tissues. OOD measure is the quantification of uncertainty for each model, using their own procedure. Higher values translate to appartenance to outlier status. Case studies A-E originate from STARE, whereas the remainder from IDRID.}
    \label{fig:case_studies_dist_gp_layers_retina_pathology}  
    %\vskip -0.2in
\end{figure*}

From Figure \ref{fig:case_studies_dist_gp_layers_retina_pathology} we can observe that DistGP-Seg manages to correctly identity the vast majority of soft and hard exudates as outliers.

\end{document}